\documentclass[sigconf]{acmart}

\usepackage{nicefrac}
\usepackage{siunitx}
\usepackage{array,framed}
\usepackage{
  color,
  float,
  epsfig,
  wrapfig,
  graphics,
  graphicx,
  subcaption
}
\usepackage{import}

\usepackage{latexsym,fancyhdr,url}
\usepackage{dsfont}
\usepackage{enumerate}
\usepackage{enumitem}
\usepackage{algorithmic}
\usepackage{algorithm}
\usepackage{graphics}
\usepackage{xspace}
\usepackage{multirow}

\usepackage{csvsimple}

\usepackage{pgf}
\usepackage{
  tikz,
  pgfplots,
  pgfplotstable
}
\usepackage{makecell}

\usepackage{tabularx}
\usepackage{adjustbox}

\usetikzlibrary{
  shapes.geometric,
  arrows,
  external,
  pgfplots.groupplots,
  matrix
}

\pgfplotsset{compat=1.9}

\newcommand{\algcomment}[1]{\hfill {\color{black}
$\triangleright$ \emph{\small{#1}}}} 
\newcommand{\mypara}[1]{\smallskip\noindent\textbf{#1}}

\usepackage{mathtools}
\usepackage{physics}

\DeclareMathAlphabet{\mathcal}{OMS}{cmsy}{m}{n}

\usepackage{wasysym}
\usepackage{tcolorbox}

\DeclareGraphicsExtensions{%
    .png,.PNG,%
    .pdf,.PDF,%
    .jpg,.mps,.jpeg,.jbig2,.jb2,.JPG,.JPEG,.JBIG2,.JB2}

\usepackage{xparse}
\newcommand{\bnm}{\begin{newmath}}
\newcommand{\enm}{\end{newmath}}

\newcommand{\bea}{\begin{eqnarray*}}%
\newcommand{\eea}{\end{eqnarray*}}%

\newcommand{\bne}{\begin{newequation}}
\newcommand{\ene}{\end{newequation}}

\newcommand{\bal}{\begin{newalign}}
\newcommand{\eal}{\end{newalign}}

\newenvironment{newalign}{\begin{align}%
\setlength{\abovedisplayskip}{4pt}%
\setlength{\belowdisplayskip}{4pt}%
\setlength{\abovedisplayshortskip}{6pt}%
\setlength{\belowdisplayshortskip}{6pt} }{\end{align}}

\newenvironment{newmath}{\begin{displaymath}%
\setlength{\abovedisplayskip}{4pt}%
\setlength{\belowdisplayskip}{4pt}%
\setlength{\abovedisplayshortskip}{6pt}%
\setlength{\belowdisplayshortskip}{6pt} }{\end{displaymath}}

\newenvironment{newequation}{\begin{equation}%
\setlength{\abovedisplayskip}{4pt}%
\setlength{\belowdisplayskip}{4pt}%
\setlength{\abovedisplayshortskip}{6pt}%
\setlength{\belowdisplayshortskip}{6pt} }{\end{equation}}

\newcounter{ctr}

\newcounter{mytable}
\def\mytable{\begin{centering}\refstepcounter{mytable}}
\def\endmytable{\end{centering}}

\newcounter{myfig}
\def\myfig{\begin{centering}\refstepcounter{myfig}}
\def\endmyfig{\end{centering}}

\newlength{\saveparindent}
\newlength{\saveparskip}
\newcommand{\E}{{\rm I\kern-.3em E}}

\renewcommand{\eqref}[1]{\mbox{Equation~(\ref{#1})}}

\def \part {part}

\def \blackslug{\hbox{\hskip 1pt \vrule width 4pt height 8pt
    depth 1.5pt \hskip 1pt}}
\def \qed{\quad\blackslug\lower 8.5pt\null\par}

\newcounter{mynote}[section]

\newcommand\ignore[1]{}

\newcounter{rcnote}[section]

\newcounter{mrnote}[section]

\newcounter{fknote}[section]

\newcounter{anote}[section]

\DeclareMathSymbol{\mlq}{\mathord}{operators}{``}
\DeclareMathSymbol{\mrq}{\mathord}{operators}{`'}

\newcommand{\rhf}[2]{R_{f, \gamma}}

\DeclareDocumentCommand{\edist}{o o}{
  \ensuremath{
    \IfNoValueTF{#1}{{d}}{{\sf d}(#1,#2)}
  }
}

\newcommand{\olrk}[1]{\ifx\nursymbol#1\else\!\!\mskip4.5mu plus 0.5mu\left(\mskip0.5mu plus0.5mu #1\mskip1.5mu plus0.5mu \right)\fi}

\NewDocumentCommand{\indseq}{ O{1} O{r} }{{#1}\ldots {#2}}

\newcommand{\dataset}{\mathcal{D}}
\newcommand{\loss}{\ell}
\newcommand{\targetDataset}{\dataset_T}
\newcommand{\targetModel}{F_{T}}
\newcommand{\auxDataset}{\dataset_A}
\newcommand{\attack}{\mathcal{A}}

\newcommand{\newwriting}[1]{\textcolor{blue}{}}

\AtBeginDocument{%
  }

\copyrightyear{2025}
\acmYear{2025}
\setcopyright{acmlicensed}\acmConference[CCS '25]{Proceedings of the 2025 ACM SIGSAC Conference on Computer and Communications Security}{October 13--17, 2025}{Taipei, Taiwan}
\acmBooktitle{Proceedings of the 2025 ACM SIGSAC Conference on Computer and Communications Security (CCS '25), October 13--17, 2025, Taipei, Taiwan}
\acmDOI{10.1145/3719027.3744818}
\acmISBN{979-8-4007-1525-9/2025/10}

\begin{document}
\title{Membership Inference Attacks as Privacy Tools: Reliability, Disparity and Ensemble}


%

\author{Zhiqi Wang}
\affiliation{%
 \institution{Rensselaer Polytechnic Institute}
 \city{Troy}
 \state{New York}
 \country{USA}
}
\email{wangz56@rpi.edu}

\author{Chengyu Zhang}
\affiliation{%
\institution{Rensselaer Polytechnic Institute}
\city{Troy}
\state{New York}
\country{USA}}
\email{zhangc26@rpi.edu}

\author{Yuetian Chen}
\affiliation{%
 \institution{Rensselaer Polytechnic Institute}
\city{Troy}
\state{New York}
\country{USA}}
\email{cheny63@rpi.edu}

\author{Nathalie Baracaldo}
\affiliation{%
 \institution{IBM Research - Almaden}
 \city{San Jose}
\state{CA}
 \country{USA}}
\email{baracald@ibm.com}

\author{Swanand Kadhe}
\affiliation{%
 \institution{IBM Research - Almaden}
 \city{San Jose}
\state{CA}
 \country{USA}}
\email{swanand.kadhe@ibm.com}

\author{Lei Yu}
\affiliation{%
\institution{Rensselaer Polytechnic Institute}
\city{Troy}
\state{New York}
\country{USA}}
\email{yul9@rpi.edu}

\renewcommand{\shortauthors}{}

\begin{abstract}
Membership inference attacks (MIAs) pose a significant threat to the privacy of machine learning models and are widely used as tools for privacy assessment, auditing, and machine unlearning. While prior MIA research has primarily focused on performance metrics such as AUC, accuracy, and TPR@low FPR—either by developing new methods to enhance these metrics or using them to evaluate privacy solutions—we found that it overlooks the disparities among different attacks. These disparities, both between distinct attack methods and between multiple instantiations of the same method, have crucial implications for the reliability and completeness of MIAs as privacy evaluation tools. In this paper, we systematically investigate these disparities through a novel framework based on coverage and stability analysis. Extensive experiments reveal significant disparities in MIAs, their potential causes, and their broader implications for privacy evaluation.
To address these challenges, we propose an ensemble framework with three distinct strategies to harness the strengths of state-of-the-art MIAs while accounting for their disparities. This framework not only enables the construction of more powerful attacks but also provides a more robust and comprehensive methodology for privacy evaluation.
\end{abstract}





\maketitle

\section{Introduction}
\label{sec:intro}

With the burgeoning development of machine learning (ML) applications, there is an increasing use of sensitive data, including financial transactions, medical records, and personal digital footprints, for training purposes. Numerous studies~\cite{papernot2018sok,rigaki2023survey,yu2024survey} have highlighted serious privacy risks associated with ML models, such as data extraction~\cite{fredrikson2015modelinversion}, membership inference~\cite{hu2022membership}, and property inference~\cite{karan2018propertyinference} attacks, primarily due to their capacity to memorize training datasets.

Membership inference attacks (MIAs) on ML models aim to determine if a specific data sample was used to train a target model or not. These attacks have received significant attention and are widely studied in ML privacy research. Beyond highlighting membership inference as a critical privacy threat, they are also frequently employed as evaluation tools across a broad range of privacy-related tasks and research efforts. These include: 
\begin{itemize}[noitemsep,topsep=0pt,parsep=0pt,partopsep=0pt,leftmargin=*]
    \item Privacy Risk Assessment: MIAs have been increasingly utilized to examine privacy risks in various machine learning contexts, such as generative adversarial networks~\cite{chen2020gan},  explainable ML~\cite{liu2024please}, diffusion models~\cite{matsumoto2023membership}, federated learning~\cite{nasr2019comprehensive}, large language models~\cite{mattern2023membership,carlini2021extracting}, and multi-modal models~\cite{duan2024membership}. MIAs are also applied across diverse applications such as social media networks~\cite{liu2019socinf}, recommendation systems~\cite{zhang2021membership}, and clinical models~\cite{jagannatha2021membership}.  
    \item Privacy Auditing: MIAs are often used as empirical tools for privacy auditing to quantify privacy leakage~\cite{mireshghallah2022quantifying,kazmi2024panoramia}. With their underlying privacy notion closely tied to differential privacy (DP), MIAs have been used to validate the bounds of DP algorithms~\cite{nasr2021adversary} and debug their implementations~\cite{tramer2022debugging}.
    \item Machine Unlearning Verification: Machine unlearning~\cite{bourtoule2021machine} involves removing the influence of a data item from a model to ensure privacy and compliance. MIAs are often used to assess whether a sample has been unlearned or not ~\cite{kurmanji2023unboundedmachineunlearning}.    
    \item Benchmarking performance of privacy-enhancing methods: Because of the effectiveness of MIAs in the above tasks, they are widely used to evaluate and benchmark the effectiveness of various privacy-preserving solutions~\cite{wang2023lds,rezaei2023accuracy}, DP algorithms~\cite{dupuy2022efficient}, and unlearning methods~\cite{nguyen2022survey,foster2024fast}.
\end{itemize}

Due to the critical nature of these tasks, extensive research efforts are being made to develop more effective and powerful MIAs~\cite{hu2022membership}. These advances are important to ensure a more accurate and comprehensive assessment of privacy risks, auditing, and unlearning verification. While \textit{balanced accuracy} and \textit{AUC} are commonly used to measure the performance of MIAs, Carlini et al.~\cite{lira} argue that these aggregate metrics often do not correlate with success rates at low false positive rates (FPRs), which are crucial for a practically meaningful evaluation of MIA effectiveness. Therefore, the true positive rate at low FPR (TPR@low FPR) has become the standard metric for evaluating the ``practical effectiveness'' of MIAs. In recent works on MIAs~\cite{lira, losstraj,diff_calibration,zarifzadeh2024lowcosthighpowermembershipinference}, both aggregate metrics and TPR@low FPR are used to evaluate and demonstrate the superiority of their proposed methods over prior attacks.

\begin{figure}[t]
\centering
\begin{subfigure}[b]{0.45\linewidth}
    \centering
    \includegraphics[width=\linewidth]{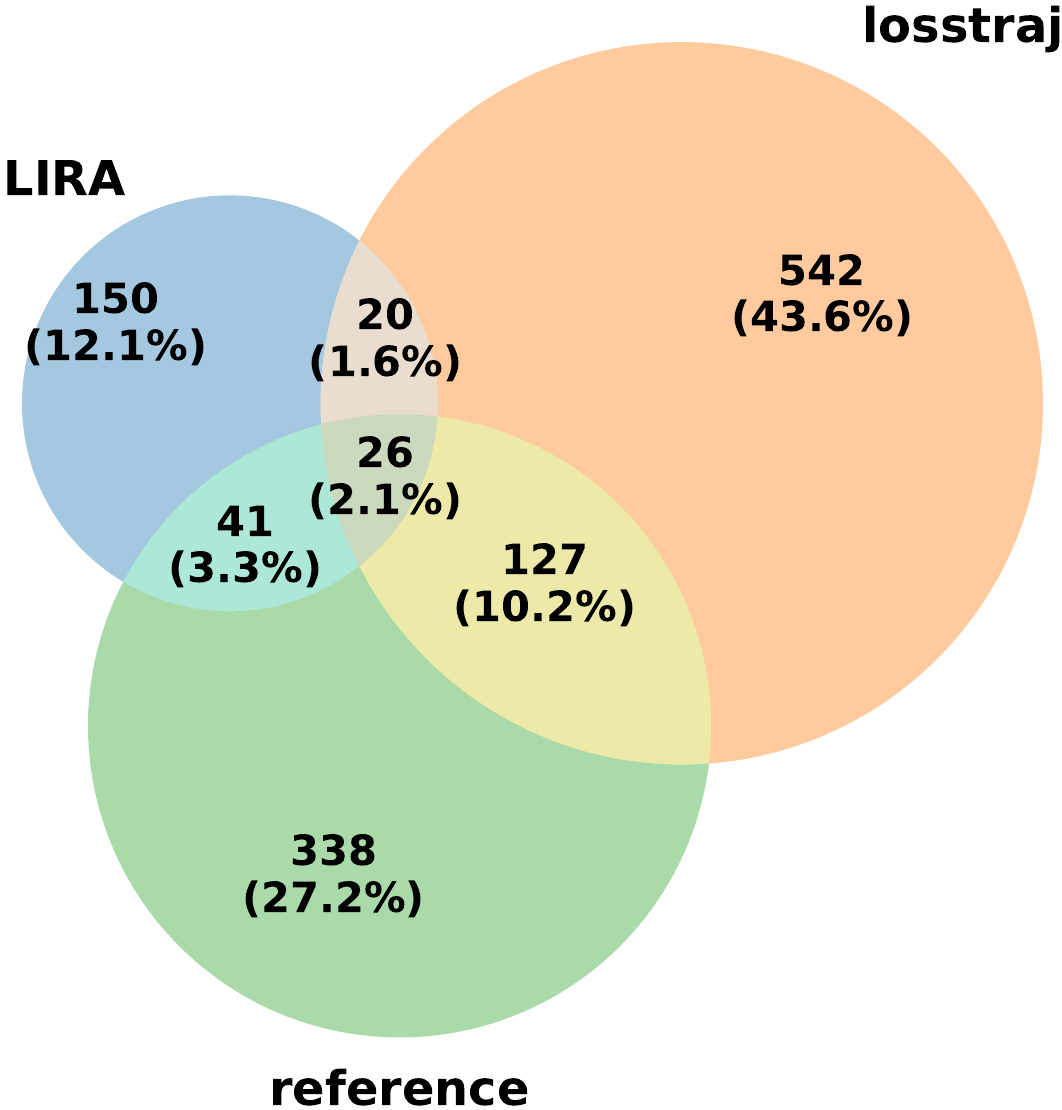}
    \caption{Members Identified by Different Attacks}
    \label{fig:intro-fig-diff-attack-venn}
\end{subfigure}%
\hfill
\begin{subfigure}[b]{0.45\linewidth}
    \centering
    \includegraphics[width=\linewidth]{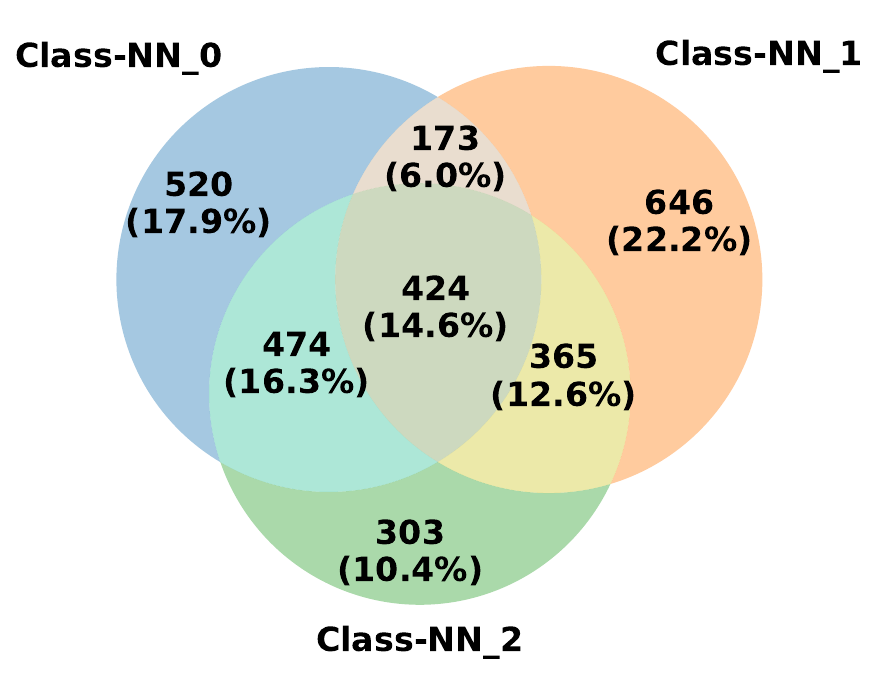}
    \caption{Members Identified by the Same Attack's Different Instances}
    \label{fig:intro-fig-same-attack-venn}
\end{subfigure}

\vspace{1em}

\caption{Venn diagram of member sets detected by (a) different attacks at a low FPR (0.1), (b) different instances of the Class-NN attack (with the same auxiliary dataset) at a low FPR (0.1). All Attacks are done on CIFAR-10. \color{black}}
\label{fig:intro-fig}
\vspace{-5pt}
\end{figure}

In this paper, however, we argue that the evaluation, even with all these metrics, may still not capture a complete picture of MIA performance. To elaborate on this, consider a target model $\targetModel$ trained on dataset $\dataset$ and two attack instances $\attack_a$ and $\attack_b$ having the same FPR. Suppose $\dataset_a$ and $\dataset_b$ represent the member subsets that can be detected by $\attack_a$ and $\attack_b$, respectively. Even if $\attack_a$ performs better than $\attack_b$ in both aggregate metrics and TPR@low FPR, relying only on $\attack_a$ may not reliably assess privacy risks and verify unlearning outcomes associated with $\dataset_b \setminus \dataset_a$. For illustration, Figure \ref{fig:intro-fig-diff-attack-venn} shows a Venn diagram of member subsets detected by three different attacks LiRA~\cite{lira}, Loss Trajectory~\cite{losstraj}, and Reference Attack~\cite{ye}, with the same FPR in our MIA experiment on CIFAR-10. The minimal overlap among them indicates that different attacks may implicitly target different subsets of members.
This observation highlights a potential limitation in the common practice of favoring one attack over another in privacy-related tasks based solely on performance metrics. Better metrics do not necessarily imply a greater overall capability of an MIA, as a sample undetected by a ``stronger'' attack may still be exposed by another. It raises two important questions relevant to ongoing MIA privacy research and practice:
\begin{itemize}[leftmargin=*,noitemsep, topsep=0pt]
    \item \textbf{Q1}: Should the effectiveness of an MIA be judged solely based on those traditional metrics~\cite{hu2022membership,duan2024membership}? More broadly, should research on developing new MIAs primarily focus on improving performance metrics while overlooking the member detection disparities between different methods?
    \item \textbf{Q2}: Is it sufficient for privacy evaluations to rely on a single ``top-performing'' MIA based on performance metrics without accounting for the disparities between different MIAs?
\end{itemize}
In this paper, we argue that the significant disparities in member detection at the sample level across different MIAs should not be overlooked when evaluating their effectiveness and employing them as tools for privacy assessment.



In addition, \textit{reliability} and \textit{consistency} are essential attributes that MIAs must possess to function effectively as privacy evaluation tools. Most existing works~\cite{duan2024membership, song2021systematic, sula2024silverliningsshadowsharnessing} that utilize MIA for privacy assessment and machine unlearning verification employ a single instance of MIA in their experiments. However, the construction of these MIAs involves inherent randomness, associated with data shuffling/sampling and training shadow/attack models, where randomness stems from factors such as optimization, weight initialization, and data batching. It has been shown that the training of randomly initialized neural networks explores different modes in the function space~\cite{fort2019deep}. Therefore, factors involving randomness inevitably lead to different decision boundaries for membership detection, often resulting in significant variance in attack outcomes among different instances of the same attack with the same auxiliary knowledge. Figure \ref{fig:intro-fig-same-attack-venn} shows that for the same attack, Class-NN attack~\cite{shokri}, three different instances that are trained on the same shadow dataset with different random seeds have large non-overlapping member sets. This indicates that the attack outcome can be highly sensitive to the randomness of attack construction. This raises another common issue in current research that uses MIAs for privacy assessment and performance evaluation: 
\begin{itemize}[leftmargin=*,noitemsep, topsep=0pt]
\item \textbf{Q3}: Is it sufficient to evaluate and report results based solely on a single instance of an MIA—as is common in existing works—while disregarding the disparities among instances that naturally arise from randomness in attack construction?
\end{itemize}
In this paper, we argue that using MIAs in their current form for evaluation, without accounting for these disparities, may lead to incomplete or potentially unreliable results.

To address these concerns, this paper first systematically investigates the disparities among different MIA methods and their instances. We propose a novel framework that introduces \textit{coverage} and \textit{stability} analysis to evaluate and quantify the disparities of MIA methods through multiple attack instances constructed with different random seeds.
Our extensive experiments highlight significant issues of instability and disparity inherent in MIAs. To better understand these disparities, we analyze the signals and features used by different MIAs to determine membership and the influence of randomness in their constructions. Our analysis reveals that different attacks may focus on samples with distinct characteristics, resulting in divergent member detection outcomes.

Furthermore, we propose an ensemble framework with three different strategies to address disparity issues in MIAs. It integrates different MIA methods from distinct perspectives: coverage-oriented, stability-oriented, and majority-oriented. These strategies combine multiple random instances of each MIA and further integrate different MIA methods to account for detection disparities. This framework not only enables the construction of more powerful attacks by leveraging the diverse strengths of existing MIAs and incorporating future advancements, but also provides an evaluation protocol to enhance the comprehensiveness of privacy evaluation.
Our extensive experiments demonstrate that these ensemble strategies achieve higher performance in traditional metrics. For example, compared to the top-performing MIA, our ensemble improves the ROC AUC and balanced accuracy by 36\% and 24\%, respectively, and increases the TPR at 0.1\% FPR by a factor of five on CIFAR-10. In addition, we discuss and evaluate practical strategies to reduce the computational cost of the ensemble.

Beyond the metrics, our ensemble strategies and their pronounced increase in attack performance serve as \emph{constructive proof} of the issues raised in Q1, Q2, and Q3. Specifically, a ``less powerful'' but high-disparity MIA remains valuable for uncovering privacy risks that are overlooked by other attacks and can further improve overall effectiveness through the ensemble (Q1). Relying on a single attack or instance, even one considered state-of-the-art, may underestimate true membership privacy risks, as members undetected by one attack may still be exposed by another (Q2, Q3). This has concrete implications for privacy practitioners and researchers applying MIAs in machine unlearning, privacy auditing, and defense evaluation: current evaluation practices that rely on a single attack instance may be unreliable, since they fail to capture the full spectrum of vulnerabilities posed by inherent disparities in MIAs. We conclude this paper with a discussion of these implications and actionable directions for future MIA research, advocating for holistic consideration of disparities and applying ensemble strategies as an evaluation protocol to enable more reliable and comprehensive privacy assessments. The source code is accessible at \href{https://github.com/RPI-DSPlab/mia-disparity}{https://github.com/RPI-DSPlab/mia-disparity}.
\section{Background and Related Work}
\label{sec:relwork}
 Membership Inference Attacks (MIAs) aim to identify whether or not a specific sample was used as training data for a target model. This paper focuses on black-box attacks in which attackers can only query the target model to obtain a prediction for a data point and use it to infer membership. In addition, attackers are able to leverage an auxiliary dataset that comes from a similar distribution as the training set of the target model. Formally, given a target sample $x$, a target model $\targetModel$ trained on the dataset $\targetDataset$, and an auxiliary dataset $\auxDataset$, membership inference attack $\attack$ can be defined as:
 \begin{equation}
     \attack(\targetModel,\auxDataset, x, \phi) \rightarrow \{ 0 , 1\}
 \end{equation}

 Here $\phi$ represents a feature extraction function applied to samples, and $\attack$ uses $\phi(x)$ as a signal to determine the membership of $x$, where 1 indicates $x$ is a member, i.e., $x\in\targetDataset$, and 0 indicates otherwise. For simplicity, $x$ represents a sample and its ground truth class as a pair $(x, y)$. Current MIAs utilize various feature extraction functions $\phi$, such as loss~\cite{yeom, choquette2021label}, full confidence vector output~\cite{shokri} of $\targetModel$, or the loss trajectory~\cite{losstraj}. As a typical intermediate step, $\attack$ assigns a membership score $\text{Score}_\attack(x)$ to every sample $x$, and compares it with a threshold to decide the membership.
 
\subsection{Representative MIAs}
\label{representative-mias}
MIA has been developed widely for different applications, language models, etc.
In this paper, we focus on a number of representative MIAs that have been widely used for privacy evaluation and assessment, unlearning, or as the comparing target for developing better MIAs against them.

\paragraph{LOSS Attack} (\citet{yeom}) This method considers an instance $x$ a member of the training set if the loss of $\targetModel$ on $x$ is less than a global threshold set as the average loss across the training set. Formally, let $\loss(x, \targetModel)$ be the loss of $\targetModel$ on instance $x$. The LOSS attack predicts $x \in \dataset_T$ if:

\begin{equation}
\label{eq:yeom-def}
\small
    \loss(x, \targetModel) < \frac{1}{|\dataset_T|} \sum_{x' \in \dataset_T} \loss(x', \targetModel)
\end{equation}

The right-hand side of (\ref{eq:yeom-def}) serves as the threshold for membership prediction. For each sample \(x\), its MIA score is computed as \(1\) minus its normalized loss on \(\targetModel\).

\paragraph{Class-NN} (\citet{shokri}) This attack involves training class-specific neural networks as membership classifiers for each class using data from shadow models.
The adversary divides $\auxDataset$ into subsets $\dataset_1, \dataset_2, \ldots, \dataset_k$, then further split each subset to $\dataset_k^{in}$ and $\dataset_k^{out}$. Subsequently, $k$ shadow models $f_1, f_2, \ldots, f_k$ are trained on $\dataset_1^{in}, \dataset_2^{in}, \ldots, \dataset_k^{in}$. An attack dataset can be constructed as
\begin{equation}
\label{eq: attack_training_set_shokri}
    \{(f_i(x), y, \text{in}) | x \in D_i^{in}, \forall i \in k\} \cup \{(f_i(x), y, \text{out}) | x \in D_i^{out}, \forall i \in k\}
\end{equation}
To train the attack classifier $C_j$ for class $j$, it finds entries in the attack dataset where $(f_i(x), y, \text{in/out}), y=j$. Then it uses those entries to train $C_j$ for each class $j$. To determine if a sample $x$ belongs to the training set of $\targetModel$, the adversary queries $C_{j=y}$ with $\targetModel(x)$, where $y$ is the label of $x$. The MIA score of this attack is the logit of the attack classifier $C_j$ in sample $x_j$ being a member.

\paragraph{Augmentation Attack} (\citet{choquette2021label})  This label-only attack uses data augmentation techniques to generate translated versions of data points, querying a shadow model trained on $\auxDataset$ to gather predictions which train an attack classifier $C$. To infer membership for a data point \( x \), it generates translated versions \(\{\hat{x}_{1}, \ldots, \hat{x}_{n}\}\), queries them on the target model \( \targetModel \) to obtain predictions \(\{\targetModel(\hat{x}_{1}), \ldots, \targetModel(\hat{x}_{n})\}\), and use these predictions to make inferences using \( C \). The MIA score is the logit of the attack classifier's prediction $C(x)$ being a member.

\paragraph{Difficulty Calibration Loss Attack} (\citet{diff_calibration}) This attack improves traditional loss-based attacks by calibrating membership scores using losses from both the target and shadow models, accounting for difficulty. It queries both the target model \( \targetModel \) and shadow model (trained on \(\auxDataset\)) \( f_s \) on all \( x \in \targetDataset \), producing two sets of predictions: \(\hat{y}^T\) and \(\hat{y}^s\). The losses for each prediction, \(\loss^T\) and \(\loss^s\), are computed using cross-entropy loss. The uncalibrated membership scores \(\loss^T\) are adjusted by computing \( s^{cal} = \loss^T - \loss^s \). The threshold \(\tau\) for determining membership is done by optimizing the prediction accuracy by splitting the \(\auxDataset\) to members (trainset for shadow model) and non-members. This time, the target model becomes the model we use to calibrate. And $\tau$ is selected by optimizing the accuracy of losses of $\auxDataset$ on the shadow model $f_s$ calibrated by the target model $\targetModel$. Similar to the Loss Attack, the MIA scores are the $1-\Vec{\loss}$ where $\Vec{\loss}$ are normalized calibrated losses of $\targetDataset$.

\paragraph{LiRA} (\citet{lira}) This Likelihood Ratio-based Instance-specific attack computes the likelihood ratio of losses for models trained with and without a particular instance, determining membership based on a threshold that optimizes attack effectiveness.
For each instance $x$, let $\dataset_{A,x}$ and $\dataset_{A,\bar{x}}$ be the subsets of $\auxDataset$ with and without $x$, respectively. The adversary trains shadow models $\{f_{x,1}, f_{x,2}, \ldots, f_{x,m}\}$ on random subsets of $\dataset_{A,x}$, and \\ $\{f_{\bar{x},1}, f_{\bar{x},2}, \ldots, f_{\bar{x},n}\}$ on random subsets of $\dataset_{A,\bar{x}}$. The likelihood ratio for $x$ is then computed as:

\begin{equation}
\label{eq:lira}
    LR(x) = \frac{\prod_{i=1}^m p(\loss(x, f_{x,i}) \mid x \in \dataset_T)}{\prod_{i=1}^n p(\loss(x, f_{\bar{x},i}) \mid x \notin \dataset_T)}
\end{equation}
where $p(\cdot \mid x \in \dataset_T)$ and $p(\cdot \mid x \notin \dataset_T)$ are the probability density functions of the losses conditioned on $x$ being a member or non-member of $\dataset_T$, respectively. The adversary then chooses a threshold $\tau$ for the likelihood ratio that optimizes the effectiveness of the attack, especially aiming for a low false-positive rate. The MIA score of LiRA reflects the likelihood ratio of $x$ being a member.

\paragraph{Reference Attack} (\citet{ye}) This attack (Attack R) uses a similar approach to LiRA by \citet{lira}. It prepares $m$ shadow models $\{f_{x,1}, f_{x,2}, \ldots, f_{x,m}\}$ on $\dataset_{\text{A}}$ with different train-test partition. It calculates the membership score as:

\begin{equation}
\label{eq:reference-attack}
    \text{Pr}_{\theta'}\left(\frac{\text{Pr}(x | \theta)}{\text{Pr}(x | \theta')} \geq 1\right)
\end{equation}
where $\text{Pr}(x | \theta')$ is the likelihood (confidence) of sample $x$ evaluated on all shadow models $\theta' \in \{f_{x,1}, f_{x,2}, \ldots, f_{x,m}\}$, and $\theta$ is the target model $\targetModel$. Similar to LiRA, the MIA score is the likelihood of $x$ being a member.

\paragraph{Loss Trajectory Attack} (\citet{losstraj}) This attack monitors the change in the loss of each sample over multiple epochs, using knowledge distillation and cross-entropy loss to track and compare loss trajectories for membership inference. 
It involves training a shadow model \(f_s\) on $\auxDataset$  and applying knowledge distillation \cite{hinton2015distilling} on both \( f_s \) and \( \targetModel \) with saving the checkpoints \( f^I \) at each epoch \( I \) over \( n \) training epochs, for capturing the loss trajectory for each sample.
For each sample \( x \in \auxDataset \), its loss trajectory \(\loss(x, f_s)\) can be obtained using each distillation checkpoint of the shadow model. We collect all loss trajectories to construct an attack training set similar to (\ref{eq: attack_training_set_shokri}) to train an attack classifier \( C \).
For a target sample \(x\), the loss trajectory \(\loss(x, \targetModel)\) is obtained using the distillation checkpoints of the target model \( \targetModel \). The classifier \( C \) is then queried with \(\loss(x, \targetModel)\) to determine membership. The MIA score is $C$'s output logit on the loss trajectory of $x$ for predicting it as a member.

\subsection{MIA Performance Metrics.} The following metrics are commonly used to evaluate the performance of MIAs \cite{choquette2021label, lira, diff_calibration, losstraj}.

\begin{itemize}[leftmargin=*,leftmargin=*,noitemsep,topsep=0pt,parsep=0pt,partopsep=0pt]
    \item \textbf{Balanced Accuracy} measures the accuracy of membership predictions on a test set with balanced priors (equal numbers of members and non-members). 

    \item \textbf{ROC} (Receiver Operating Characteristic) curve plots the true positive rate (TPR) against the false positive rate (FPR) at various threshold levels, providing a comprehensive view of the trade-offs between MIA's TPR and FPR.

    \item \textbf{AUC} (Area Under the ROC Curve) provides a single scalar value summarizing the overall performance of an attack. A higher AUC reflects that the attack achieves better overall separability between members and non-members, independent of any particular threshold.
    
    \item \textbf{TPR@Low FPR} focuses on the practical effectiveness of MIAs. A low false positive rate imposes a constraint on membership predictions, requiring the model to be more "cautious" when predicting members to minimize false alarms. 
\end{itemize}

\section{Disparity Evaluation Methodology}
\label{sec:methodology}
In this section, we present the metrics and methodology to evaluate disparities of MIAs at both the instance level and the method level. 


\subsection{Instance Level Disparity Over Randomness}
\label{sec:mia-prediction-over-randomness}
Most MIAs~\cite{shokri,losstraj, lira, ye} for deep learning rely on the shadow training technique, which trains multiple shadow models on an auxiliary dataset to replicate the behavior of the target model. This process inherently involves randomness from several sources, including the partitioning of the auxiliary dataset into member and non-member sets, weight initialization in the training algorithm, and data shuffling and batching. These factors introduce variability in the outcomes of both shadow models and attack classifiers, ultimately affecting the detection outcomes of MIAs. In our study, we abstract this randomness using a single random seed, representing a random MIA instance that an attacker might create using the same algorithm but under different randomness sources in real-world scenarios.

\noindent\textbf{Disparity Metric}: To evaluate an MIA’s instance-level disparity in member detection, we introduce \emph{consistency} score, which quantifies the similarity of membership predictions between attack instances using pairwise \emph{Jaccard Index}.
The Jaccard Index (or Jaccard Similarity) measures the similarity between finite sample sets and is defined as the size of the intersection divided by the size of the union of the sample sets. 

Given a set of random seeds $S$, we create $|S|$ number of instances for an MIA \(\attack\), its consistency on target dataset \(\dataset\) is defined as the average of Jaccard Index between every pair of attack instances $\attack^i$ and $\attack^j$ on their detected member sets $\mathbb{M}_\dataset(\attack^i)$ and $\mathbb{M}_\dataset(\attack^j)$, i.e.,

\begin{equation}
\label{eq:Consistency}
    \text{Consistency}_S^{\dataset}(\attack) = \frac{1}{\binom{|S|}{2}} \sum{\substack{i, j \in S \\ i < j}} J(\mathbb{M}_\dataset(\attack^i), \mathbb{M}_\dataset(\attack^j))
\end{equation}
where the Jaccard Index
$J(\mathbb{M}(\attack^i), \mathbb{M}(\attack^j)) = \frac{|\mathbb{M}(\attack^i) \cap \mathbb{M}(\attack^j)|}{|\mathbb{M}(\attack^i) \cup \mathbb{M}(\attack^j)|}$.

A lower consistency score indicates greater discrepancy in the detected member sets across different instances of the same attack, even when provided with identical auxiliary information and target model. We use this metric to measure the variance in an MIA method’s membership detection outcomes due to randomness in its construction, indicating that evaluations based on a single instance may not reliably capture an MIA method’s true privacy risks. For the LOSS attack, which uses a fixed global threshold and does not involve randomness in its construction, its consistency score is 1.

\noindent\textbf{Coverage and Stability}: Due to the discrepancy in member detection across random instances of an MIA, evaluations based on a single random instance—as is common in most existing works—cannot fully capture the true privacy risk posed by an MIA method at the method level under the same auxiliary knowledge, as opposed to the leakage revealed by a specific instantiation. Consequently, such evaluations may provide an incomplete picture of the effectiveness of a privacy solution. To address this limitation, we introduce the evaluation measures \emph{coverage} and \emph{stability}.

The union of true positive attack results from multiple instances of an MIA constructed with different random seeds, referred to as the \textbf{coverage} of an attack.
Formally, given an attack $\attack$ and a set of possible seeds $S$, for each random seed $s \in S$, we can construct an instance of $\attack$ with randomness generated from random seed $s$, denoted by $\attack^s$. The membership prediction for the data point $x$ is $\attack^s(x)$. The coverage of attack $\attack$ is represented as
\begin{equation}
\small
\label{eq:coverage}
    \text{Coverage}_S(\attack) = \left\{ x \in \targetDataset :  \bigcup_{s \in S} \attack^s(x) = 1 \right\}
\end{equation}

Similarly, we define \textbf{stability} as the intersection of true positives across all instances, reflecting how consistently an attack identifies members despite randomness. This excludes members whose status is inconsistently predicted across runs:
\begin{equation}
\small
\label{eq:stability}
    \text{Stability}_S(\attack) = \left\{ x \in \targetDataset :  \bigcap_{s \in S} \attack^s(x) = 1 \right\}
\end{equation}


\begin{figure*}[h]
    \centering
    \begin{subfigure}[b]{0.15\textwidth}
        \includegraphics[width=\textwidth]{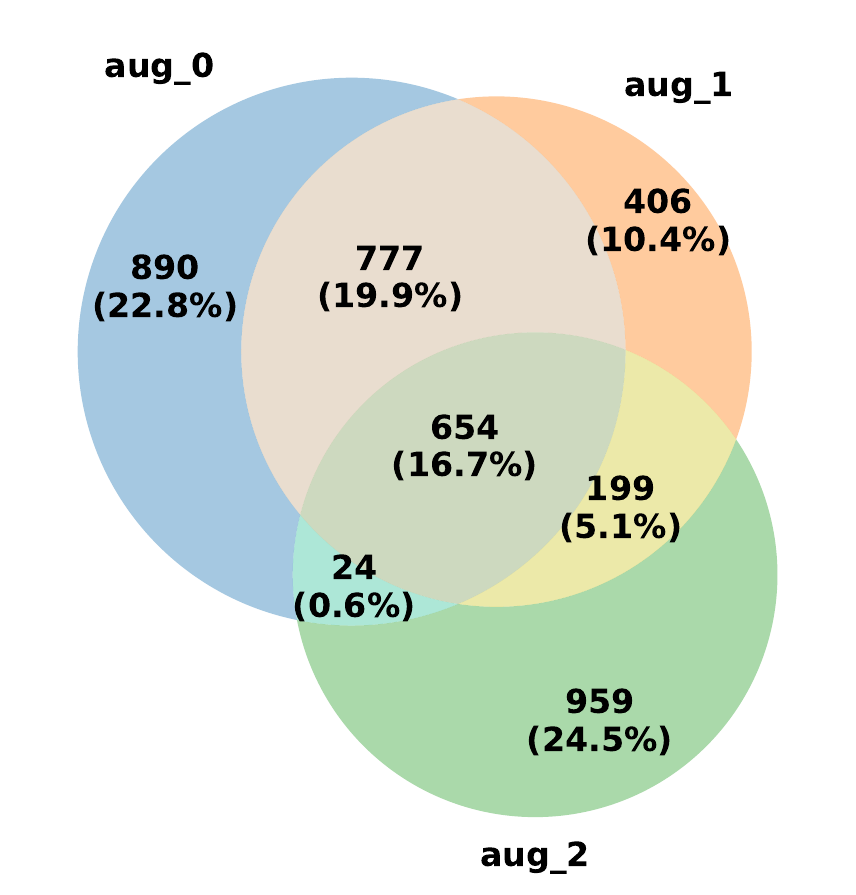}
        \caption{Augmentation Attack}
        \label{fig:aug_venn}
    \end{subfigure}
    \begin{subfigure}[b]{0.16\textwidth}
        \includegraphics[width=\textwidth]{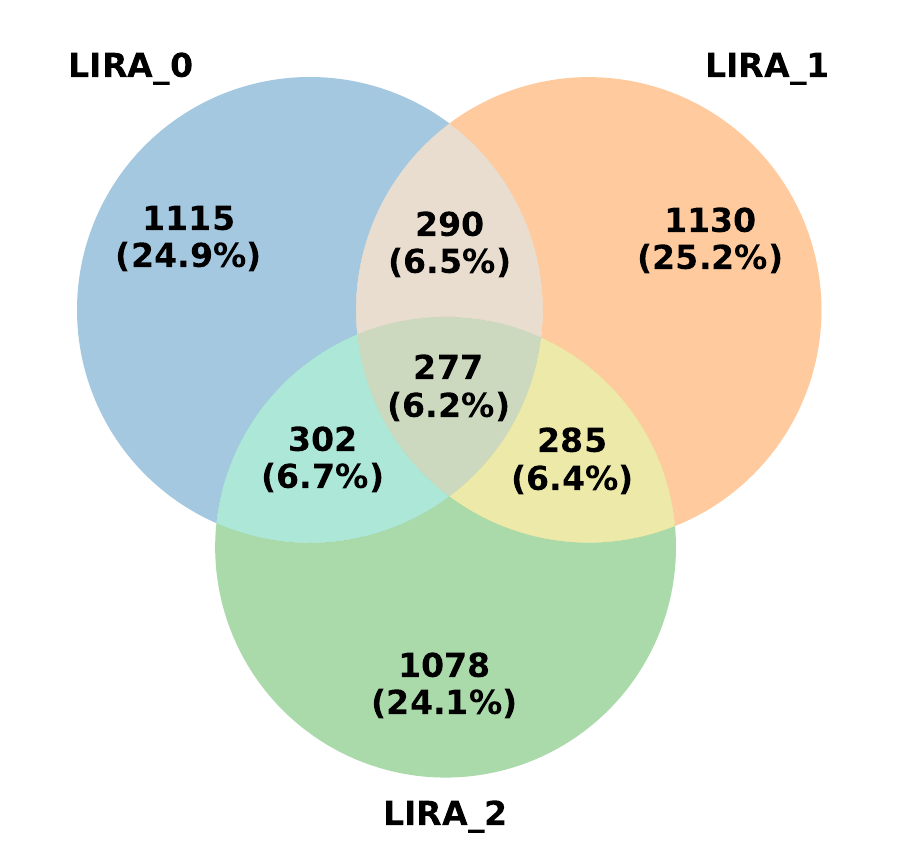}
        \caption{LIRA \\ ~}
        \label{fig:lira_venn}
    \end{subfigure}    
    \begin{subfigure}[b]{0.16\textwidth}
        \includegraphics[width=\textwidth]{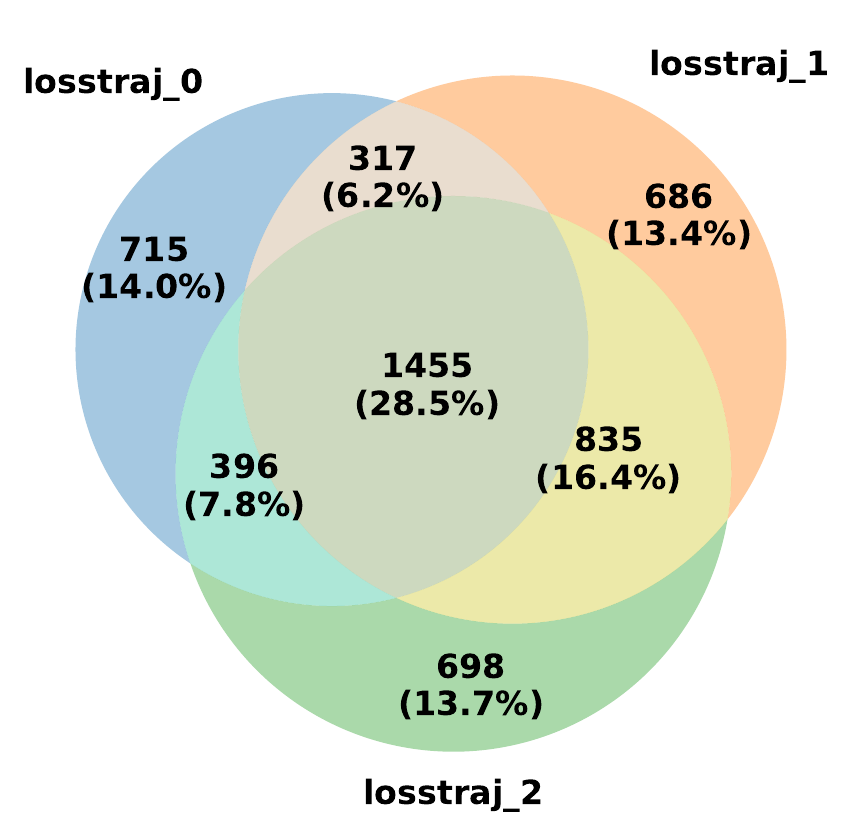}
        \caption{Loss Trajectory Attack}
        \label{fig:losstraj_venn}
    \end{subfigure}
    \begin{subfigure}[b]{0.16\textwidth}
        \includegraphics[width=\textwidth]{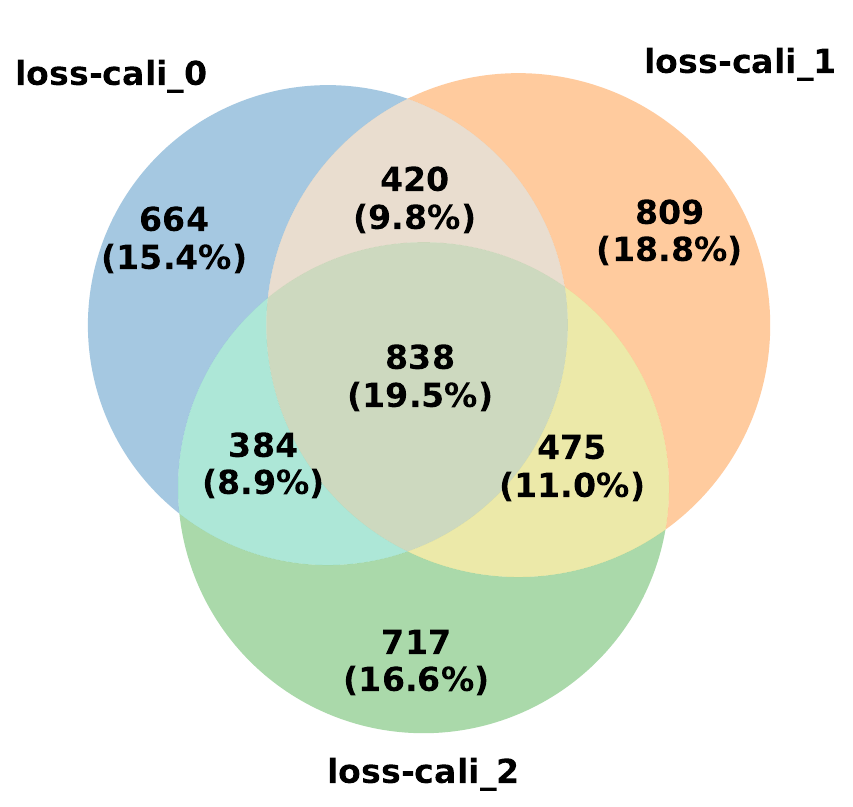}
        \caption{Diff-Calibration Loss Attack}
        \label{fig:cali_venn}
    \end{subfigure}
    \begin{subfigure}[b]{0.17\textwidth}
        \includegraphics[width=\textwidth]{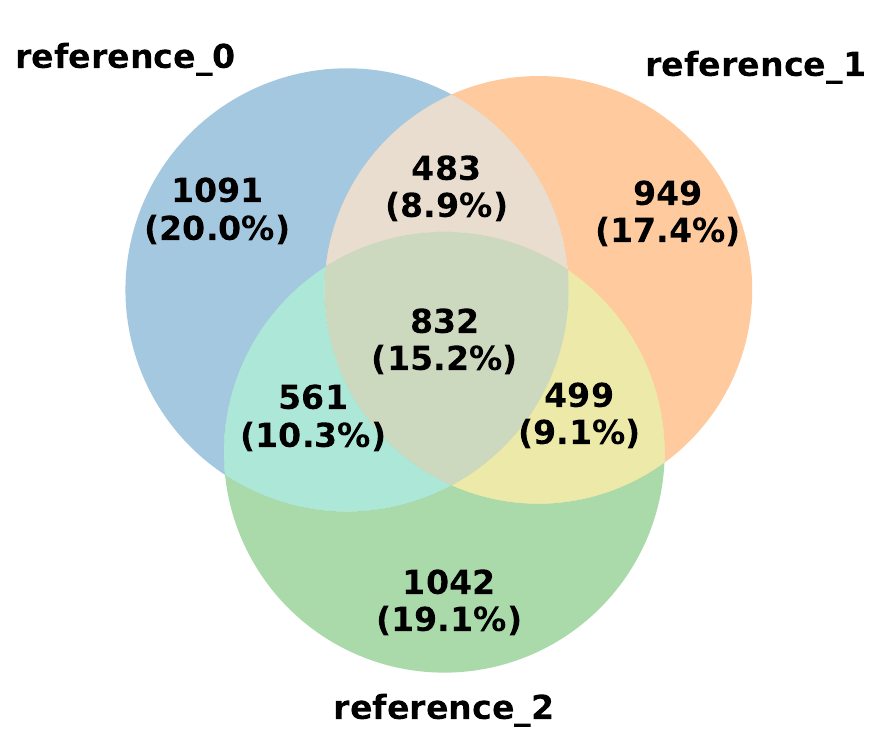}
        \caption{Reference \\ Attack}
        \label{fig:reference_venn}
    \end{subfigure}
    \begin{subfigure}[b]{0.15\textwidth}
        \includegraphics[width=\textwidth]{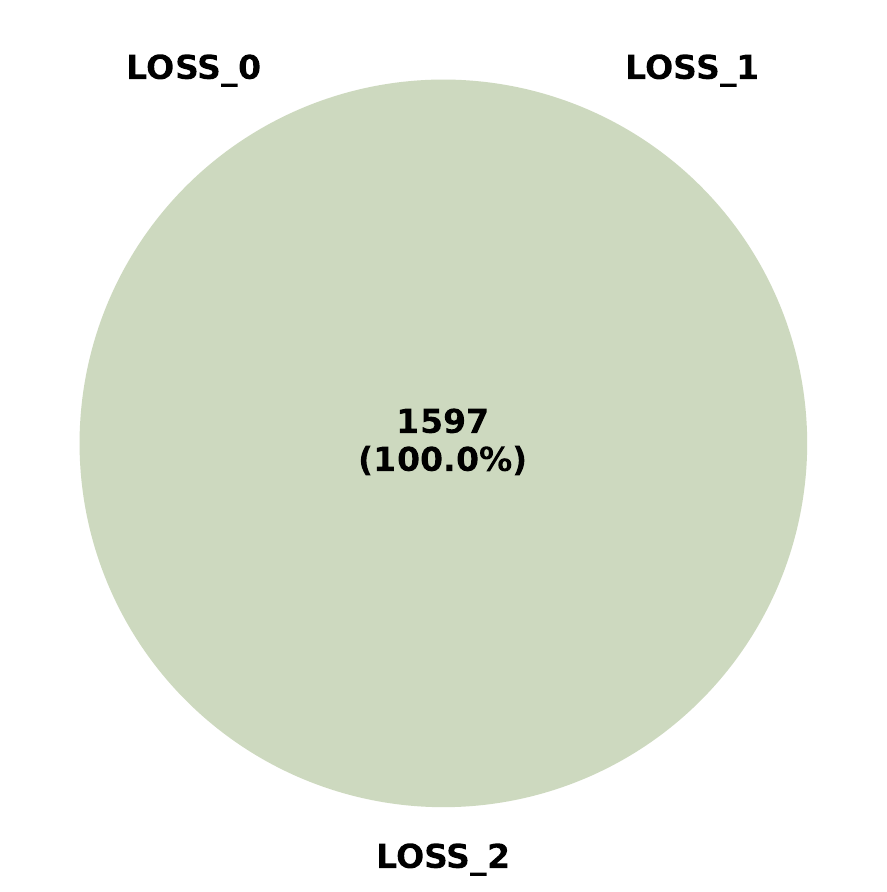}
        \caption{LOSS \\ Attack}
        \label{fig:yeom_venn}
    \end{subfigure}

    \vspace{1em}
    \caption{\textbf{Venn Diagram of three MIA instances at FPR = 0.1 for different attack methods}. Each set represents the true positive samples from one instance. The Venn diagram of the Class-NN attack is shown in Figure \ref{fig:intro-fig-same-attack-venn}.}
    \label{fig:venn_diagram_intra_attack}
\end{figure*}
Coverage reflects the extent of potential privacy leakage, while stability captures the consistency of privacy vulnerability under an MIA method. Because the privacy risks they reveal are independent of any specific instance, given their convergence observed in Section \ref{sec:coverage-and-stability-over-randomness}, we compute the Jaccard similarity of coverage and stability to characterize the method-level disparities across different MIAs, i.e., the differences in the subsets of the training data targeted by different MIA methods (regardless of any specific instance).

\noindent\textbf{Illustrative Example:} Figure \ref{fig:venn_diagram_intra_attack} shows the union (coverage) and intersection (stability) of three instances for each attack method. As we can see, all attacks that involve randomness from shadow model training, except the LOSS attack~\cite{yeom}, exhibit significant variations in member detection, with their coverage and stability changing considerably from one instance to three instances. Therefore,
it is evident that single-instance-based MIA evaluations or assessments may be unreliable. 

\subsection{Multi-instance Attack Analysis}
\label{sec:multi-instances-attack-analysis}
\begin{figure}
\centering
\includegraphics[width=1\columnwidth]{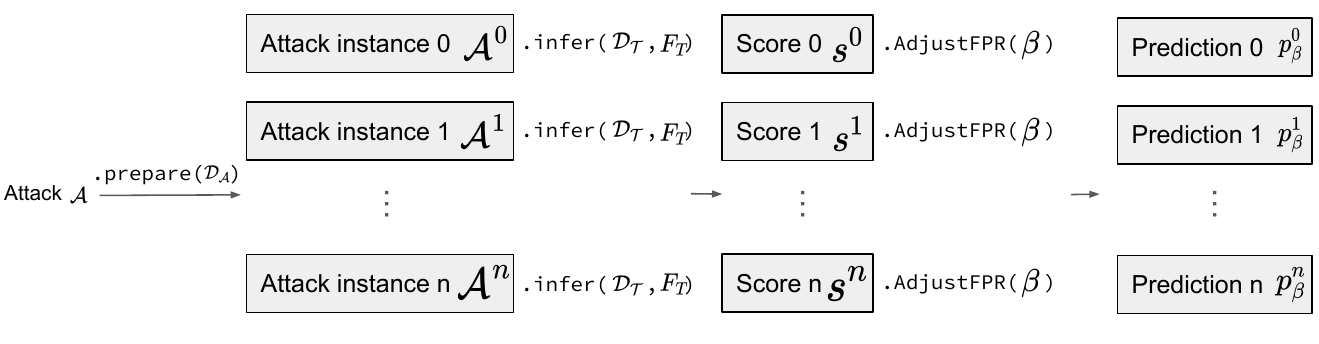}
\caption{\textbf{MIA Multi-Instance Analysis Pipeline}. The process includes preparing attack instances, inferring membership, and adjusting predictions based on a given FPR.}
\label{fig:pipline}
\end{figure}

To analyze the instance-level disparity from the lens of coverage and stability, we introduce a multi-instance analysis framework.
As demonstrated in Figure \ref{fig:pipline}, we first prepare \( n \) instances of an attack \(\attack\) using the same auxiliary dataset \(\auxDataset\) but different random seeds. To attack a target model \(\targetModel\), each MIA instance performs inferences on the target dataset \(\targetDataset\) with access to \(\targetModel\). The MIA scores obtained are converted to binary membership predictions using Algorithm \ref{alg:adjust_fpr} (\textbf{AdjustFPR}) to obtain predictions at a specific FPR level \(\beta\). It is crucial because we need to maintain a consistent level of FPR to ensure a fair comparison between different MIA instances on their coverage and stability derived from true positive detections. 
With the predictions of multiple instances of \(\attack\), we can compute the coverage and stability of \(\attack\) over \( n \) instances. 
\begin{algorithm}
\small
\begin{algorithmic}[1]
    \REQUIRE Membership ground truth array $gt$, MIA score array $scores$, target false positive rate $\beta$
    \STATE $\text{FPRs}, \text{TPRs}, \text{thres} \gets \texttt{roc\_curve}(gt, scores)$ \algcomment{function $\texttt{roc\_curve}$ from \text{scikit-learn}\cite{scikit-learn}.}
    \STATE $idx \gets \text{argmin}(|\text{FPRs} - \beta|)$
    \STATE $\tau \gets \text{thres}[idx]$
    \STATE $pred \gets \mathds{1}[scores \geq \tau]$ \algcomment{membership indicator function.}
    \RETURN $pred$
\end{algorithmic}
    \caption{\textbf{FPR-Based Thresholding (\textit{AdjustFPR})} This algorithm predicts membership by determining a MIA score threshold \( \tau \) that achieves a specified target FPR level \( \beta \).}
\label{alg:adjust_fpr}
\end{algorithm}
\begin{algorithm}
\small
 \begin{algorithmic}[1]
    \REQUIRE Attacks $\mathbb{A}$, untrained model $f$, Dataset $\dataset$, seeds $\mathcal{S}$,  stability or coverage analysis $analyze$, desired false positive rate $\beta$

    \STATE $\targetDataset, \auxDataset \gets \texttt{partition}(\dataset)$
    \STATE $ \dataset_{\text{target\_train}}, \dataset_{\text{target\_{test}}} \gets \texttt{partition}(\targetDataset)$
    \STATE $gt = \vec{1}_{\text{\texttt{len}}(\dataset_{\text{target\_train}})} \oplus \vec{0}_{\text{\texttt{len}}(\dataset_{\text{target\_test}})}$ \algcomment{ground truth array}
    \STATE $\targetModel = \texttt{train\_model}(f, \dataset_{\text{target\_train}})$
    \STATE $f_{\text{access}} = \texttt{blackbox\_access}(\targetModel)$
    \FOR{each attack $\attack \in \mathbb{A}$ }
        \FOR{each seed $s$ in $\mathcal{S}$}
            \STATE $\texttt{set\_seeds}(s)$
            \STATE $\attack\texttt{.prepare}(f_{\text{access}}, \auxDataset)$ \algcomment{shadow model training}
            \STATE $scores = \attack\texttt{.infer}(\mathcal{\targetDataset})$
            \STATE $preds_s \gets \texttt{AdjustFPR}(gt, scores, \beta)$
        \ENDFOR
        \STATE $analyze(\{preds_s: s\in \mathcal{S}\}, gt)$
    \ENDFOR
 \end{algorithmic}
    \caption{\textbf{Multi-instance Attack Analysis Framework}  This procedure handles the training, preparation, and execution of attacks, and computes aggregated results to assess stability and coverage across different attack configurations.}
 \label{alg:miae}   
\end{algorithm}

Given multi-instance membership inference predictions, coverage helps capture all possible risks, i.e., members that are vulnerable to any MIA instance at a given FPR level \(\beta\), while stability focuses on vulnerable members that are consistently detected by an MIA across different random instantiations. Our complete evaluation framework follows Algorithm \ref{alg:miae} to compare different attack instances under the same conditions. It splits the dataset \(\dataset\) into two non-overlapping datasets, the auxiliary dataset \(\auxDataset\) and the target dataset \(\targetDataset\). \(\targetDataset\) is further divided into two equal-size, non-overlapping subsets in line 2, one for training the target model (constituting the members) and one for testing (comprising the non-members). This division ensures that both subsets are of equal size (\(|\dataset_{\text{target\_train}}| = |\dataset_{\text{target\_test}}|\)), ensuring a balanced prior of memberships. Lines 6 to 12 apply the pipeline in Figure \ref{fig:pipline} to each attack method. Line 13 calculates the stability and coverage of each attack over \(|S|\) instances.

\subsection{MIA Method Level Disparity}
\label{sec:evaluating-multiple-attacks}
As discussed in Section~\ref{sec:mia-prediction-over-randomness}, we use coverage and stability to evaluate disparities between different MIAs. These measures enable us to analyze how existing attacks differ at the method level in terms of both the extent of vulnerable members they expose (i.e., coverage) and the consistency of vulnerable samples across instances (i.e., stability), despite the presence of instance-level variance.

To assess the method-level disparity, we compute the Jaccard index between the coverage/stability sets of different MIA methods. For each MIA, both coverage and stability are evaluated based on the predictions from the same number of instances at an identical FPR level to ensure fairness, as described in Section~\ref{sec:multi-instances-attack-analysis}. In addition, we conducted a preliminary empirical analysis of the following two aspects to explore potential factors that contribute to method-level disparities.

\begin{itemize}[leftmargin=*,noitemsep, topsep=0pt]
\item \textbf{Stability Difference in Model Output Space}:
    We define \emph{\(\attack\)-unique samples} as the set of members correctly identified by the stability of MIA \(\attack\), but not by the stability of any other MIA methods. Formally, the set of \(\attack\)-unique samples, denoted as \(S_\attack^{\text{unique}}\), is expressed as:
\begin{equation}
    S_\attack^{\text{unique}} = \{x \mid x \in \text{Stability}(\attack) \land x \notin \bigcup_{B \neq \attack} \text{Stability}(B) \}
\end{equation}

In a black-box attack setting, logits encapsulate the maximum information returned by a query.
Given the model's logits output on those samples uniquely ``consistently captured'', we look into the difference in their distributions to understand if an MIA may target or is more sensitive to distinct output distributions of members, which may help explain the disparities among MIAs.


\item \textbf{Attack Signal Difference}: Different MIA methods use different feature extraction method $\phi$, resulting in different signals for MIA. 
To understand the impact of signals while isolating the effect of randomness and MIA methodology difference, we focused on the Class-NN MIA method and \emph{\(\attack\)-covered samples} that are the members identified by MIA \(\attack\)'s coverage,
\begin{equation}
    S_\attack^{\text{covered}} = \{x \mid x \in \text{Coverage}(\attack)\}
\end{equation}
Class-NN uses logits as attack signals, so we can easily manipulate the signal by restricting it to only the top-$x$ logits while masking out the rest, referred to as ``$x$-top'' Class-NN. This modification allows us to observe how variations in the signals received by the same MIA influence the resulting detected member sets.


\end{itemize}

\section{Evaluation}
\label{sec:eval}
In this section, we evaluate the disparities between the seven widely used MIAs described in Section~\ref{sec:relwork}, using the methodology introduced earlier to assess both the instance-level and the method-level disparities, and investigate their potential causes.

\subsection{Experiment Setup}
\label{sec:experiement-setup}

To make sure our empirical analysis is comprehensive, our experiment uses five datasets and four neural network architectures, listed below. A more detailed set-up including the hyperparameter choices of MIAs is provided in Appendix Section~\ref{sec:app_experiment_setup}. 

\mypara{Datasets.} We use five datasets commonly adopted in MIA research: CIFAR-10, CIFAR-100, CINIC-10, Purchase100, and Texas100. CIFAR-10 and CIFAR-100 consist of 60,000 32x32 color images divided into 10 and 100 classes, respectively. CINIC-10, an extension of CIFAR-10, includes 270,000 images derived from CIFAR-10 and ImageNet. Purchase100 and Texas100 are structured datasets representing consumer purchase behaviors and hospital discharge records, respectively. Detailed dataset descriptions are provided in Appendix Section~\ref{sec:app_dataset_details}. 
Unless otherwise specified, we present experimental results based on CIFAR-10.

\mypara{Models.} We employed ResNet-56~\cite{resnet}, MobileNetV2~\cite{mobilenet}, VGG-16~\cite{vgg16}, and WideResNet-32~\cite{wrn} as our primary model architectures for image datasets, with ResNet-56 being the main model used for reporting experimental results. All models are optimized with SGD and a cosine learning rate scheduler~\cite{sgdr}.  We choose MLP for tabular datasets Purchase100 and Texas100. Training and evaluation configurations, including dataset partitions and training epochs, are detailed in Appendix Section~\ref{sec:app_model_setup}.

\mypara{MIA setup.} For most MIAs examined in this paper, we adhered to the standard settings used to produce the main results in their respective papers, except LiRA. Our experiment uses LiRA's online version in its paper. A detailed discussion of the setup for these MIAs and LiRA is provided in Appendix~\ref{sec:appe_setup-for-mias}, and the consistency result of offline LiRA is discussed in Appendix Section~\ref{sec:lira-online-offline-comp-appendix}. For disparity evaluation, we utilize six instances to compute coverage, stability, and consistency scores, as these metrics generally start to converge in most cases at this number of instances, as shown in Section~\ref{sec:coverage-and-stability-over-randomness}. Additionally, we focus on presenting results at FPR=0.1, with results for other FPR settings available in the Appendix.

Additionally, we examine the impact of outliers and auxiliary-target dataset distribution gap, with relevant results presented in Appendix Section~\ref{sec:mia-on-ood} and Section~\ref{ssec:distributionshift}.

\subsection{MIA Instance Level Disparity}
\label{sec:inconsistency-of-membership-inference-attacks}

    

    


\begin{figure*}[t] 
    \centering
    \begin{subfigure}{0.33\textwidth} 
        \centering
        \includegraphics[width=\textwidth]{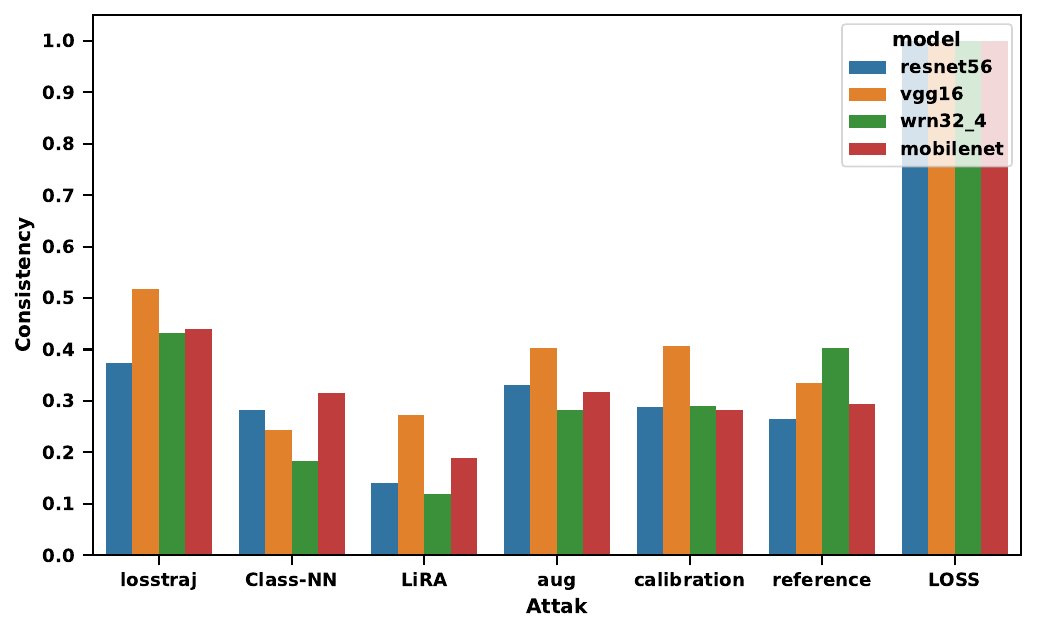}
        \caption{CIFAR-10}
        \label{fig:consistency-all-model-cifar10-appendix}
    \end{subfigure}
    \hfill 
    \begin{subfigure}{0.33\textwidth} 
        \centering
        \includegraphics[width=\textwidth]{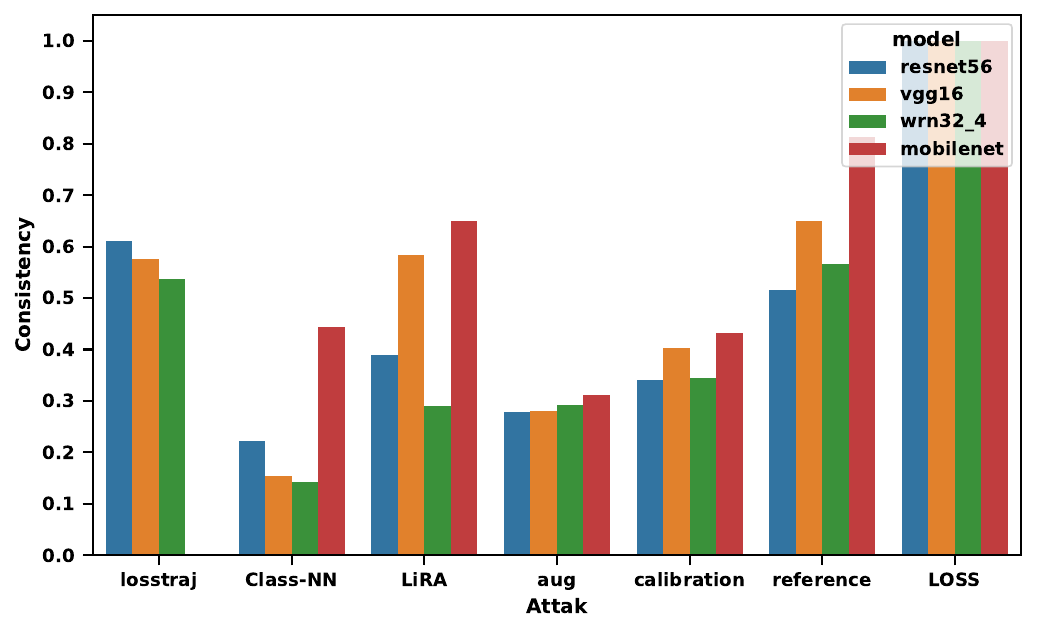}
        \caption{CIFAR-100}
        \label{fig:consistency-all-model-cifar100-appendix}
    \end{subfigure}
    \hfill 
    \begin{subfigure}{0.33\textwidth} 
        \centering
        \includegraphics[width=\textwidth]{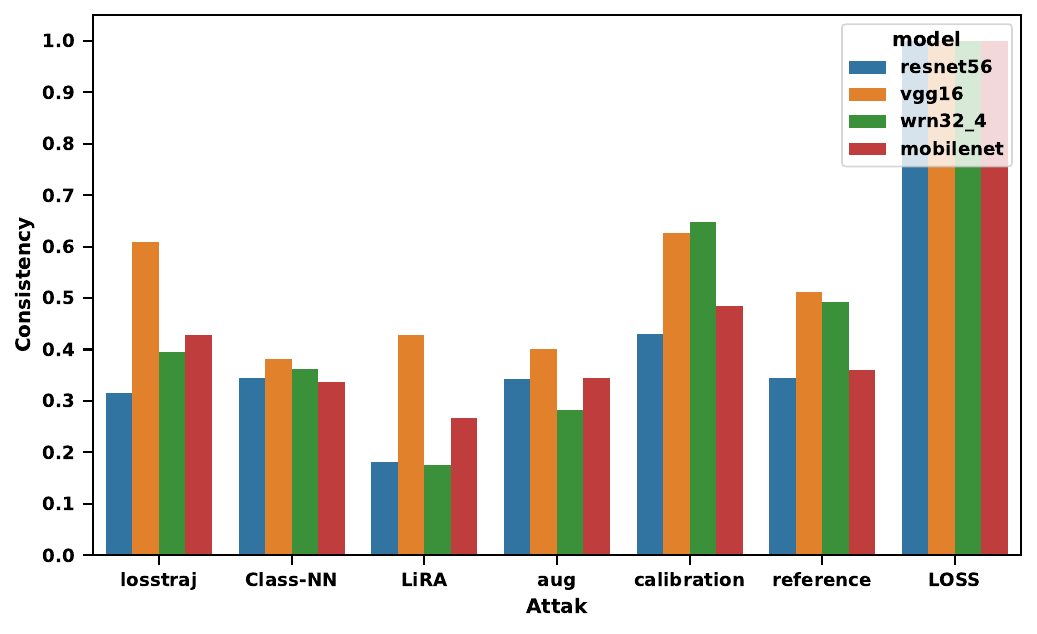}
        \caption{CINIC-10}
        \label{fig:consistency-all-model-cinic10-appendix}
    \end{subfigure}
    \vspace{1em}
    \caption{Consistency score shows inherent disparities among pairs of instances of MIAs (except LOSS attack) across datasets and models (ResNet-56, VGG-16, WideResNet-32, MobileNetV2). Consistency evaluated at FPR=0.1.}
    \label{fig:consistency-all-model-all-ds}
\end{figure*}

\subsubsection{Inherit Low Consistency of MIA}
\label{sec:inherit-low-consistency-of-mia}
Following the methodology introduced in Section~\ref{sec:mia-prediction-over-randomness}, we evaluate the consistency scores of different MIAs, each using six instances under the standard setting.
Figure~\ref{fig:consistency-all-model-all-ds} shows the consistency scores for each MIA. Except for the LOSS attack, all other attacks exhibit low consistency across datasets and model architectures, with an average consistency score below 0.4, highlighting the inherent instance-level inconsistency.

The instance-level MIA consistency appears to be influenced by the dataset. Attacks on CIFAR-100 demonstrate the highest overall consistency, likely due to increased over-fitting, as indicated by the larger generalization gap between the training and testing performance of target models (Appendix Table~\ref{table:train_test_acc}). Greater over-fitting results in a larger set of common members that are easier for MIAs to infer, thereby leading to higher consistency. Additionally, certain attacks exhibit higher consistency on specific datasets and model architectures. For example, LiRA and Reference Attack achieve relatively high consistency on CIFAR-100 with VGG-16 and MobileNetV2. In contrast, Class-NN consistently shows low consistency across all datasets. This inconsistency arises from its non-overlapping shadow training sets, which lead to less aligned shadow models.

\subsubsection{Disparity Factors}
The randomness in MIA construction involves random partitioning of the auxiliary dataset into member and non-member sets, shadow model training, and attack model training. We investigate how these factors contribute to instance-level disparity in MIAs, particularly at low false positive rates (FPR).

\begin{figure}[]
    \centering
    \begin{subfigure}[b]{0.233\textwidth}
        \centering
        \includegraphics[width=\textwidth]{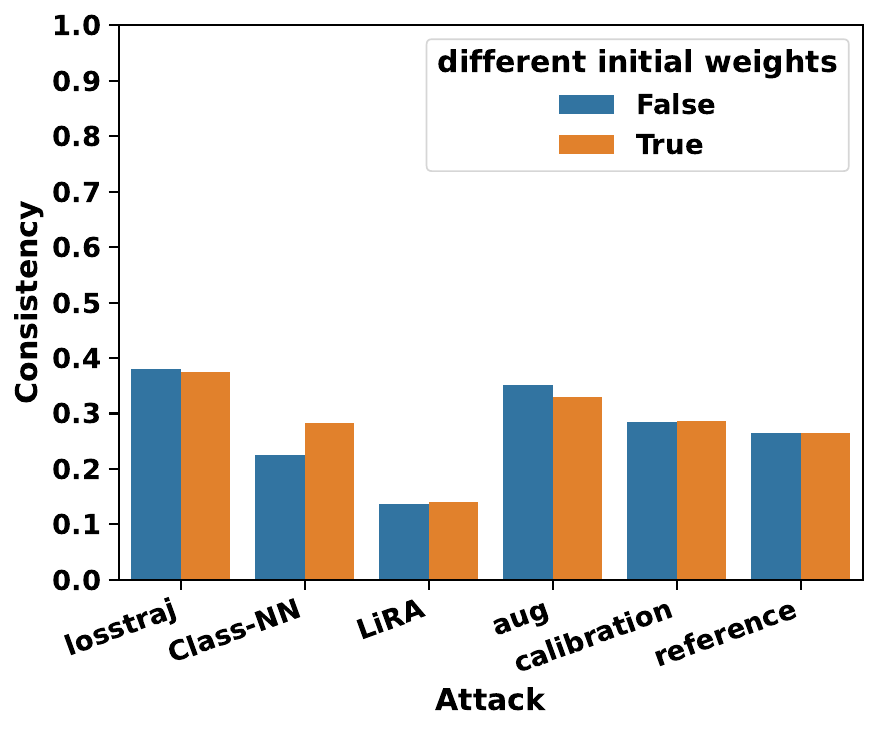}
        \caption{Shadow Model Training}
        \label{fig:shadow-init}
    \end{subfigure}
    \hfill
    \begin{subfigure}[b]{0.233\textwidth}
        \centering
        \includegraphics[width=\textwidth]{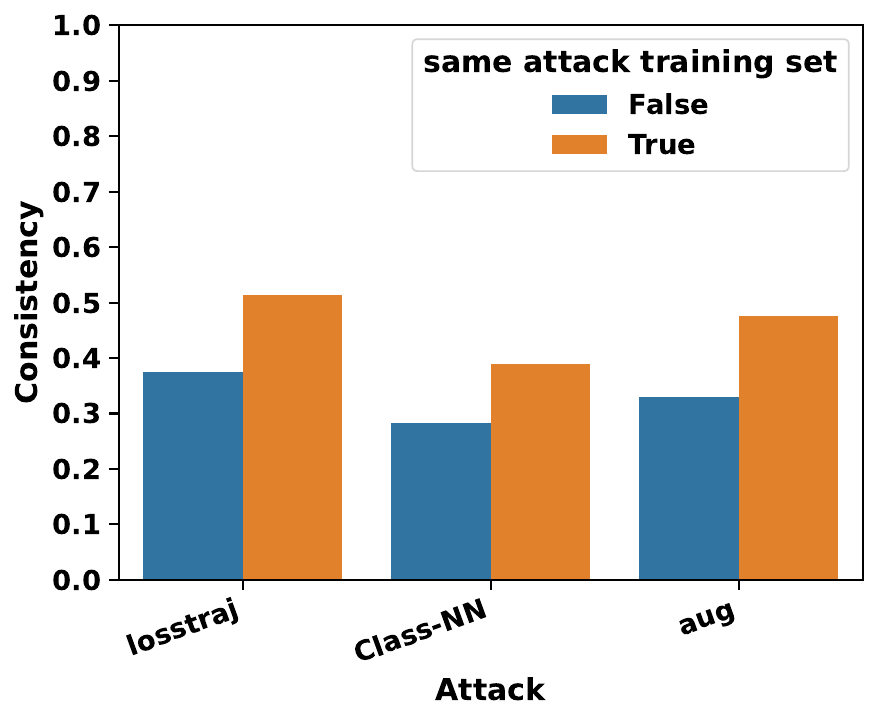}
        \caption{Attack Model Training}
        \label{fig:same-shadow-diff-attack}
    \end{subfigure}
    
    \vspace{1em}

\caption{Shadow model training and attack model training both contribute to the disparity. Consistency is measured with six instances at FPR=0.1 on CIFAR-10.}
\label{fig:same-shadow-diff-attack-shadow-init}

\end{figure}

\noindent\textbf{Shadow Model Training:}
For MIAs that rely on shadow models, shadow model training involves data shuffling and partitioning, weight initialization, and other randomness factors that are specific to an MIA, such as model distillation in Loss Trajectory Attack.

To analyze the effects of shadow model training, we compute the consistency score for instances created with different initial weights and compare it to the score for instances created with the same initial weights. When all instances start with the same set of initial weights for shadow models, data shuffling and partitioning become the primary sources of randomness for shadow model training. Figure~\ref{fig:shadow-init} shows the consistency scores of each MIA under these two conditions. The ``False'' condition represents the same set of initial model weights. Comparing the two, we observe that the effect of varying initial weights on disparity is minimal (i.e., changes in the consistency score are no more than 0.05), indicating that data random shuffling and partitioning are the primary contributors to the disparity.

\noindent\textbf{Attack Model Training:}
For attacks that include attack classification models, the training of these models can also contribute to instance-level disparity. To evaluate this effect, we fix the shadow models across all instances, ensuring that their attack models are trained on the same attack training set (corresponding to the ``True'' case in Figure~\ref{fig:same-shadow-diff-attack}). Comparing this setup with the normal scenario where shadow models and attack training sets vary, we find that the consistency score increases by 8\% to 12\%. This indicates that attack model training also contributes to MIA disparity, though its impact is less pronounced than that of shadow model training.

These findings highlight that both shadow model training and attack model training are critical factors driving high instance-level disparity in MIAs.

\begin{figure}[]
    \centering
    \begin{subfigure}[b]{0.233\textwidth}
        \centering
        \includegraphics[width=\textwidth]{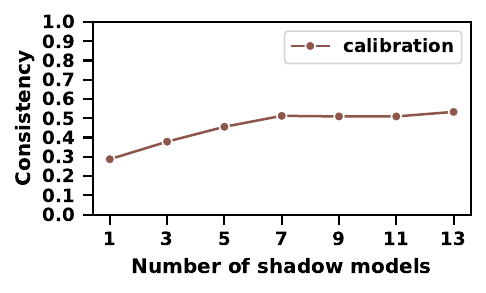}
        \caption{Diff-Calibration}
        \label{fig:diff-num-shadow-calibration}
    \end{subfigure}
    \hfill
    \begin{subfigure}[b]{0.233\textwidth}
        \centering
        \includegraphics[width=\textwidth]{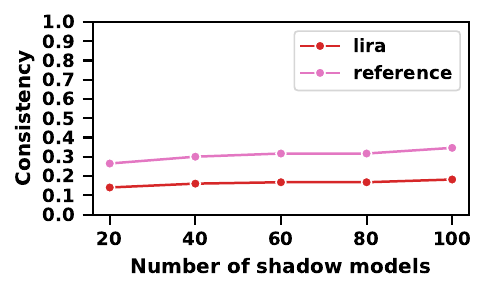}
        \caption{LiRA and Reference Attack}
        \label{fig:diff-num-shadow-lira-reference}
    \end{subfigure}
    
    \vspace{1em}

\caption{\textbf{Number of shadow models' relation with Consistency.} the model architecture is ResNet-56, and the dataset is CIFAR-10. Consistency is measured with six instances at FPR=0.1.}
\label{fig:diff-num-shadow}

\end{figure}

\noindent\textbf{Number of Shadow Models:}
Additionally, we examine the impact of the number of shadow models used in an MIA on disparity. The Class-NN Attack is excluded from this analysis, as it trains shadow models on disjoint datasets, meaning that increasing the number of shadow models reduces the training data size and thus the quality of shadow models. Loss Trajectory and Augmentation attacks are also excluded, as their methodologies do not specify how they operate with multiple shadow models.

As shown in Figure~\ref{fig:diff-num-shadow}, increasing the number of shadow models in an MIA slightly increases the consistency at the instance level.
For the Difficulty Calibration Loss Attack, the consistency increases as the calibration term (computed from shadow-model losses) becomes a more stable empirical estimate of difficulty. This reduces the variance of the calibration loss and enhances consistency. For LiRA and the Reference Attack, the increase in consistency is relatively smaller. Overall, despite the increase in the number of shadow models, instance-level disparity remains significant across MIAs. 

\subsection{Coverage and Stability Over Randomness}
\label{sec:coverage-and-stability-over-randomness}
\subsubsection{Coverage Over Randomness.}
\label{sec:coverage-over-randomness-result}

\begin{figure}
    \centering
    \begin{subfigure}[b]{0.23\textwidth}
        \includegraphics[width=\textwidth]{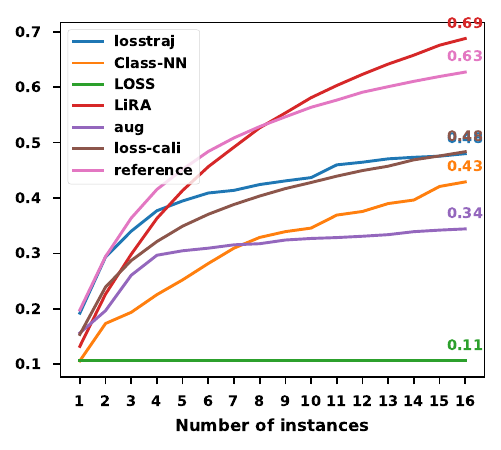}
        \caption{True positive rate}
        \label{fig:union_tpr_conv}
    \end{subfigure}
    \hfill
    \begin{subfigure}[b]{0.23\textwidth}
        \includegraphics[width=\textwidth]{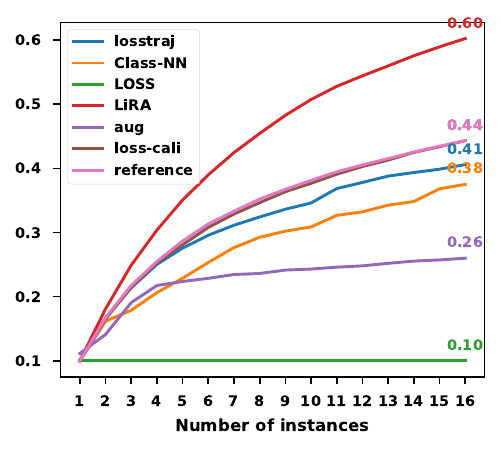}
        \caption{False Positive Rate}
        \label{fig:union_fpr_conv}
    \end{subfigure}

    \vspace{1em}

\caption{Trends in TPR and FPR for coverage under different numbers of instances with $FPR=0.1$. For all attacks, each instance is created using the same auxiliary dataset of 30,000 samples, and they predict membership on a disjoint target dataset of 30,000 samples.}
\label{fig:coverage-convergence-tpr-fpr}

        \vspace{1em}

    \begin{subfigure}[b]{0.23\textwidth}
        \includegraphics[width=\textwidth]{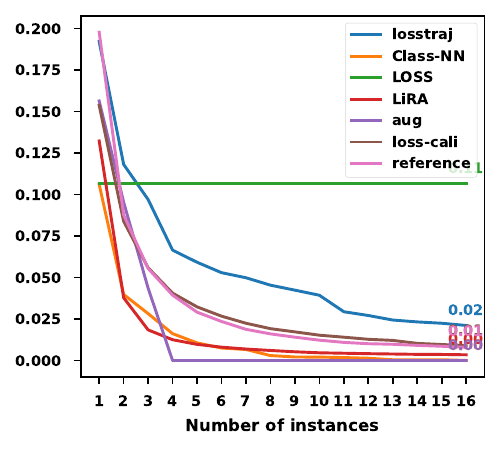}
        \caption{True positive rate}
        \label{fig:intersection_tpr_conv}
    \end{subfigure}
    \hfill
    \begin{subfigure}[b]{0.23\textwidth}
        \includegraphics[width=\textwidth]{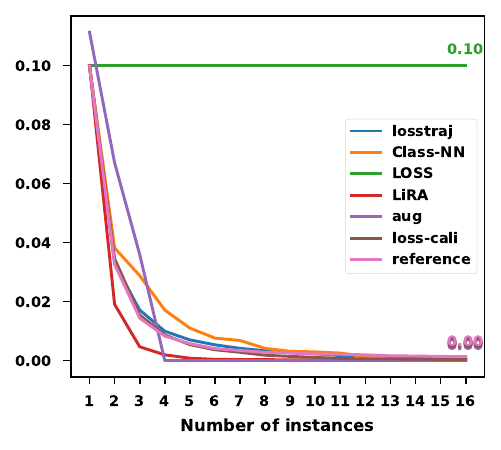}
        \caption{False Positive Rate}
        \label{fig:intersection_fpr_conv}
    \end{subfigure}

    \vspace{1em}
    
    \caption{Trends in TPR and FPR for stability, following the same setup as Figure~\ref{fig:coverage-convergence-tpr-fpr}.}
    \label{fig:stability-convergence-tpr-fpr}
\end{figure}

\begin{figure}
    \begin{subfigure}[b]{0.23\textwidth}
        \includegraphics[width=\textwidth]{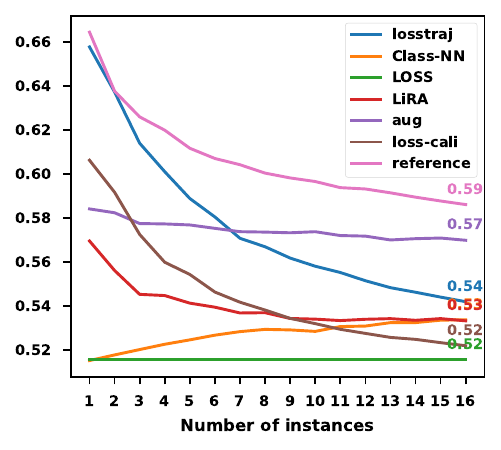}
        \caption{Coverage Precision}
        \label{fig:union_precision_conv}
    \end{subfigure}
        \hfill
        \begin{subfigure}[b]{0.23\textwidth}
        \includegraphics[width=\textwidth]{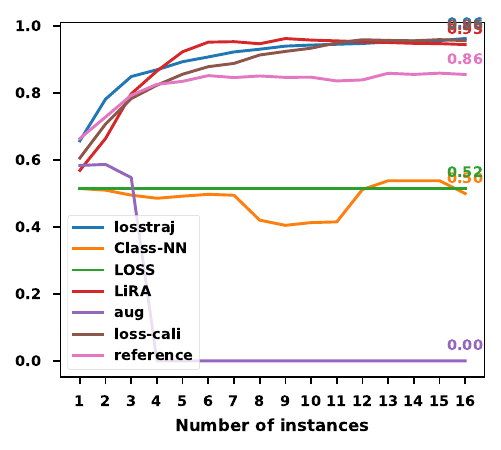}
        \caption{Stability Precision}
        \label{fig:intersection_precision_conv}
    \end{subfigure}
             \vspace{1em}

        \caption{Precision of coverage and stability corresponding to Figure \ref{fig:coverage-convergence-tpr-fpr} and \ref{fig:stability-convergence-tpr-fpr}.}
        \label{fig-precision-convergence}
\end{figure}

To evaluate the coverage of each attack, we compute the union of positive membership predictions across varying numbers of instances and present the results in Figure~\ref{fig:coverage-convergence-tpr-fpr}.
Figure ~\ref{fig:union_tpr_conv} shows that as we increase the number of instances, TPR (i.e., coverage) increases accordingly. 
However, as shown in Figure~\ref{fig:union_fpr_conv}, the FPR also rises with additional instances, indicating that more nonmembers are incorrectly classified as members. Figure~\ref{fig:union_precision_conv} further illustrates the corresponding decrease in precision. Notably, the drop in precision is less pronounced, as the growth in true positives partially offsets the increase in false positives. As the true positive set stabilizes, precision also converges.

We repeat the same experiment by configuring each MIA instance with different FPR thresholds: 0.001, 0.01, and 0.2. The results are provided in Figure~\ref{fig:coverage-tp-diff-fpr}   in Appendix Section \ref{sec:mia-method-disparity-at-different-fpr}. We observe that Loss Trajectory, Reference, and Calibrated Loss attacks consistently achieve the highest number of true positive samples when multiple instances are used. At FPR 0.01, the coverage of these attacks with multiple instances captures approximately five times more members compared to a single instance. 
In contrast, the coverage of the Loss attack remains unchanged across all metrics, as it is not affected by randomness. 
We also observe convergence at the tail end of each TPR curve, indicating that an MIA instance under a fixed FPR can only identify a subset of members within a bounded group, even under different randomization conditions.

\subsubsection{Stability Over Randomness}
\label{stability-over-randomness-result}

Following the same setting as the coverage evaluation, we compute the intersection of positive membership predictions to assess stability. The results are presented in Figure~\ref{fig:stability-convergence-tpr-fpr}.
The figures show that both TPR and FPR decrease with the number of instances, indicating substantial variance in the detection results between different MIA instances, except for the LOSS attack, which is not affected by randomness.

We also conduct the same experiment with MIA instances at different FPR values of 0.001, 0.01, and 0.2. The results are provided in Figure \ref{fig:stability-tp-diff-fpr} in Appendix Section \ref{sec:mia-method-disparity-at-different-fpr}. We observed that at low FPRs (e.g., 0.001, 0.01), the stability of most attacks converge to fewer than 10 and 50 true positive members, respectively. This finding highlights that only a small subset of data points is consistently vulnerable to a given MIA method, regardless of randomization effects.
As the number of consistently identified members decreases, precision (Figure~\ref{fig:intersection_precision_conv}) increases significantly. Specifically, the Loss Trajectory, Calibrated Loss, and LiRA achieve precision values that exceed 95\%. In contrast, the Augmentation attack and Class-NN attack fail to show similar improvements, reflecting their limited capability to consistently predict vulnerable members. Importantly, achieving high precision does not require all 16 instances; most precision gains are realized within the first six instances, while the FPR of stability drops to near zero.

\subsection{MIA Method Level Disparity}
\label{sec:agreement-among-mia-methods}
\begin{figure}[t]
    \centering
    \begin{subfigure}[b]{0.23\textwidth}
        \centering
        \includegraphics[width=\textwidth]{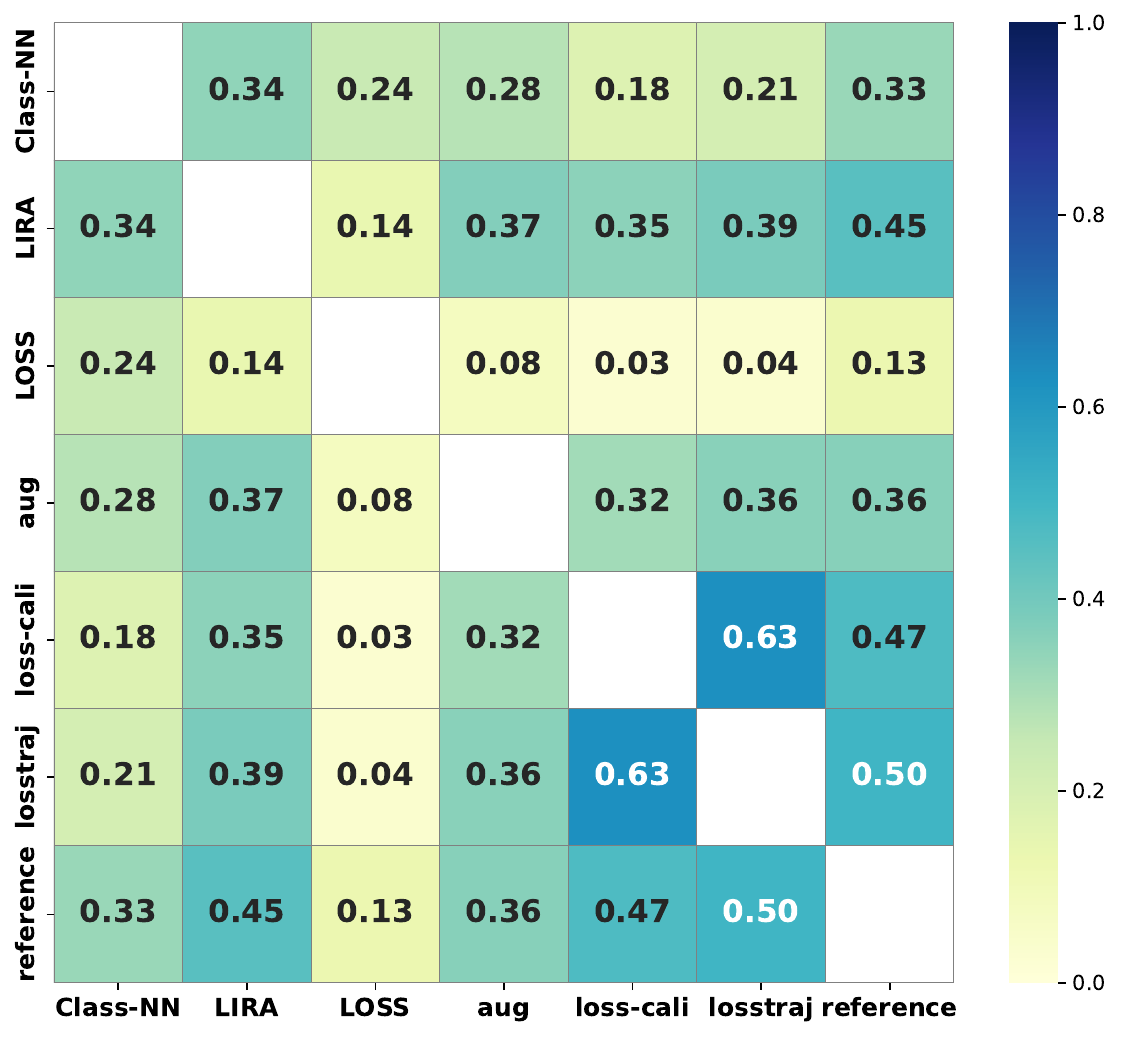}
        \caption{Coverage}
        \label{fig:similarity-matrix_coverage}
    \end{subfigure}
    \hfill
    \begin{subfigure}[b]{0.23\textwidth}
        \centering
        \includegraphics[width=\textwidth]{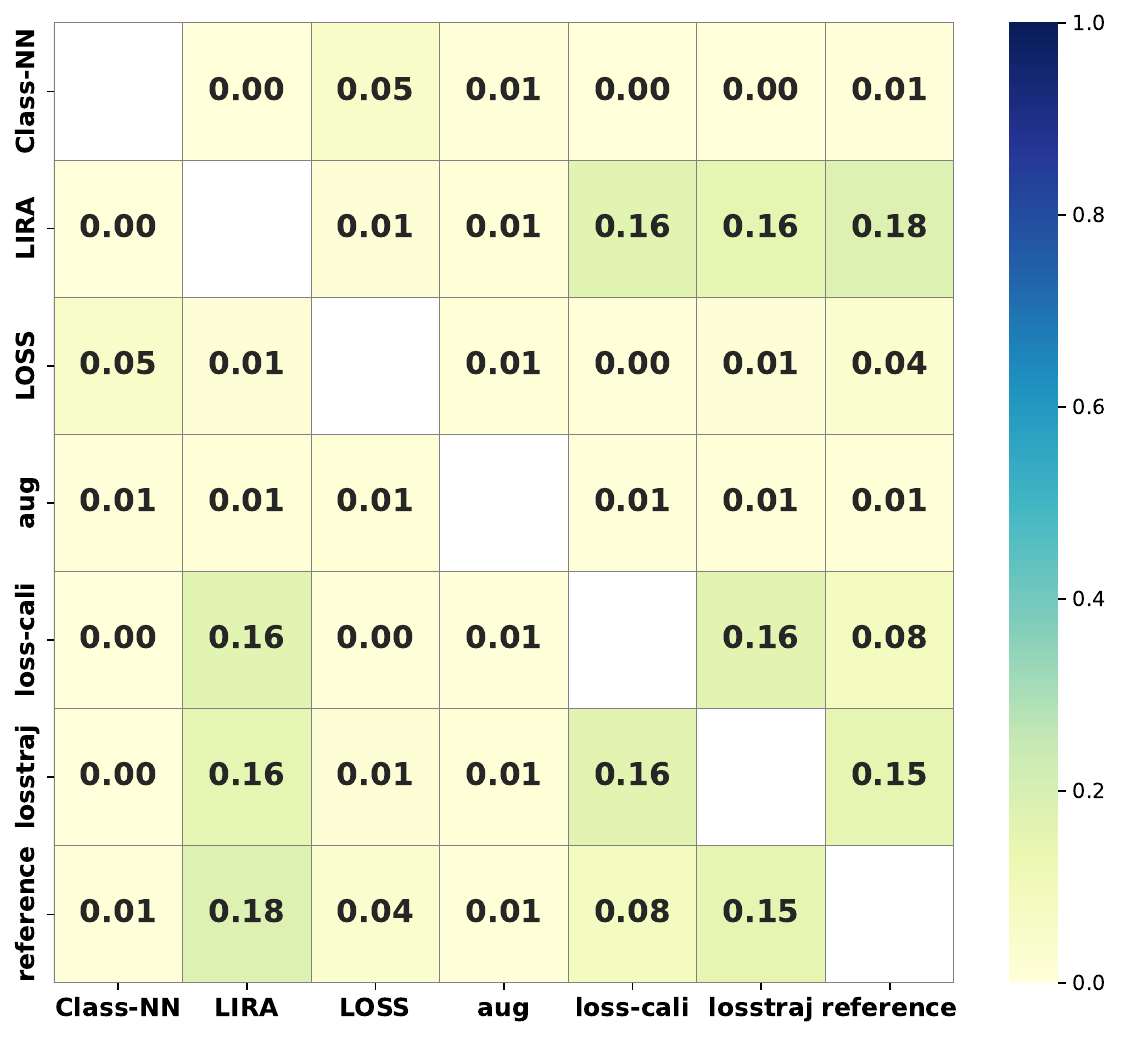}
        \caption{Stability}
        \label{fig:similarity-matrix_stability}
    \end{subfigure}
    
    \vspace{1em}
    
    \caption{MIA Method Disparity. The values represent the average Jaccard similarity of 4 experiment runs. Attacks' coverage and stability are calculated with six instances at FPR=0.1 on CIFAR-10.}
    \label{fig:attack_similarity}
\end{figure}

Following the methodology in Section \ref{sec:evaluating-multiple-attacks}, we compute the Jaccard similarity of coverage/stability between every pair of MIA methods.
Figure~\ref{fig:attack_similarity} presents the results for coverage and stability derived from six instances constructed for each MIA method with FPR=0.1. Both coverage and stability show low similarity between most attack pairs, with Jaccard similarity generally below 0.4 for coverage and 0.1 for stability. The notably lower Jaccard similarity in stability compared to coverage underscores the significant disparities in consistently detected vulnerabilities across different MIAs.

Certain attack pairs exhibit similar trends in both coverage and stability. For example, LiRA and Reference Attack show higher similarity in both measurements, likely due to their shared approach to shadow model training. Similarly, Loss Calibration, Loss Trajectory, and Reference Attacks show mutual similarity, likely because they are all based on loss signals. Conversely, some attack pairs show significant disagreement; for example, the Loss Attack consistently demonstrates low Jaccard similarity with all other attacks. We also observe that attacks that perform well at low FPR (e.g., Loss Trajectory, Reference attack, Loss Calibration attack, and LiRA) tend to be more mutually similar compared to attacks designed for average-case performance (e.g., Class-NN, Loss, Augmentation attack). Overall, while certain attack pairs produce membership predictions with moderately higher similarity than others, the Jaccard similarity remains low across the board when predictions are made at FPR=0.1.

\begin{figure}
    \centering
    \begin{subfigure}[b]{0.23\textwidth}
        \centering
        \includegraphics[width=\textwidth]{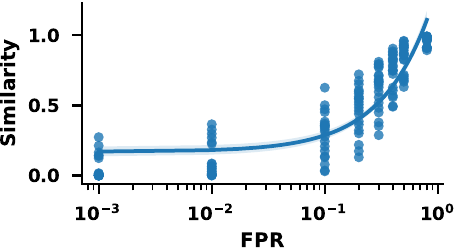}
        \caption{Coverage; CIFAR-10}
        \label{fig:trend-of-agreement-vs-fpr-coverage-cifar-10}
    \end{subfigure}
    \hfill
    \begin{subfigure}[b]{0.23\textwidth}
        \centering
        \includegraphics[width=\textwidth]{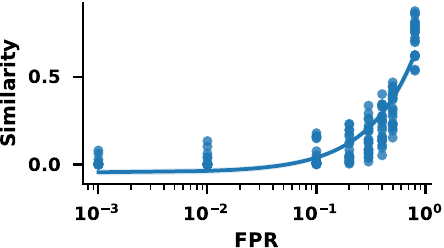}
        \caption{Stability; CIFAR-10}
        \label{fig:trend-of-agreement-vs-fpr-stability-cifar-10}
    \end{subfigure}

    \vspace{1em}

    \begin{subfigure}[b]{0.23\textwidth}
        \centering
        \includegraphics[width=\textwidth]{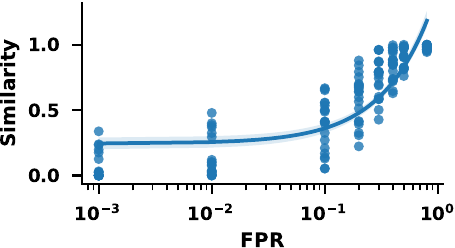}
        \caption{Coverage; CIFAR-100}
        \label{fig:trend-of-agreement-vs-fpr-coverage-cifar-100}
    \end{subfigure}
    \hfill
    \begin{subfigure}[b]{0.23\textwidth}
        \centering
        \includegraphics[width=\textwidth]{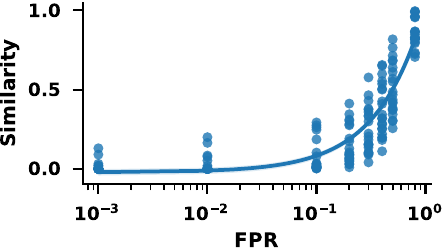}
        \caption{Stability; CIFAR-100}
        \label{fig:trend-of-agreement-vs-fpr-stability-cifar-100}
    \end{subfigure}
    
    \vspace{1em}
    
    \caption{Correlation of Disparity and FPR. The line is a linear regression on all Jaccard similarity scores for different FPR values of instances.}
    \label{fig:trend-of-agreement-vs-fpr}
\end{figure}

We also evaluate the pairwise similarity between different MIA methods in terms of coverage and stability under varying instance-level FPR values (from 0.001 to 0.2). The results are presented in Appendix Section~\ref{sec:mia-method-disparity-at-different-fpr}. Across different FPR settings, the overall similarity trend remains consistent, and we observe a positive correlation between FPR and similarity. In Figure~\ref{fig:trend-of-agreement-vs-fpr}, each point represents the Jaccard Similarity value between a pair of attacks at a given FPR level.
As shown, both coverage and stability exhibit low similarity at low FPRs. As the FPR increases, the similarity also increases, with coverage similarity values approaching 1. The relatively low similarity at low FPR suggests that different attacks may have distinct insights into predicting members, especially at low FPRs. The member predictions in which these attacks are most confident tend to be nearly disjoint across methods, implying that traditional metrics—particularly those evaluated at low FPR—do not fully capture the diverse behaviors and strengths of different attacks.

\subsection{Disparity Empirical Analysis}
\label{sec:empirical-analysis}
In this section, we conduct a preliminary empirical analysis to analyze the potential causes of disparities between MIAs with the methodology presented in Section \ref{sec:evaluating-multiple-attacks}.
This analysis offers insights into the underlying factors that lead different MIA methods to implicitly target different subsets of training data.

\subsubsection{Output Distribution of \(\attack\)-Unique Samples}
\label{sec:pca-result}
To understand the difference in data points that are vulnerable to different attacks, we analyze the output space of the target model \(\targetModel\). We apply Principal Component Analysis (PCA) to extract feature scores from the logits predicted by \(\targetModel\) for each member. Focusing on \(\attack\)-unique members (as defined in Section~\ref{sec:evaluating-multiple-attacks}) allows us to understand the disparities among MIAs through their uniquely identified members.

\begin{figure}
    \centering
    \begin{subfigure}[b]{0.233\textwidth}
        \centering
        \includegraphics[width=\textwidth]{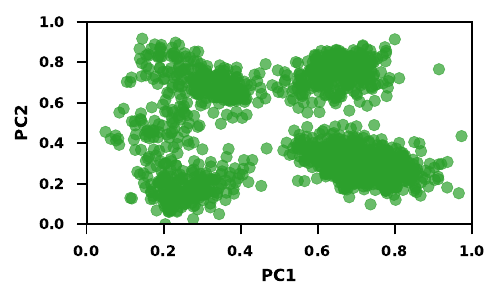}
        \caption{Loss Attack}
        \label{fig:yeom-pca}
    \end{subfigure}
    \hfill
    \begin{subfigure}[b]{0.233\textwidth}
        \centering
        \includegraphics[width=\textwidth]{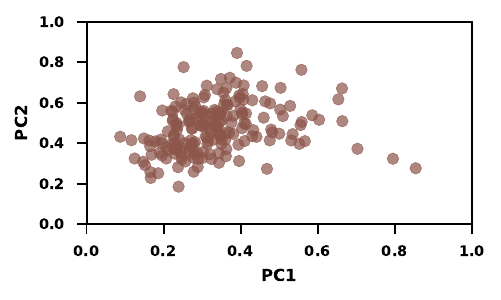}
        \caption{Calibration Attack}
        \label{fig:calibration-pca}
    \end{subfigure}
    
    \vspace{1em}

    \begin{subfigure}[b]{0.7\columnwidth}
    \centering
    \includegraphics[width=\textwidth]{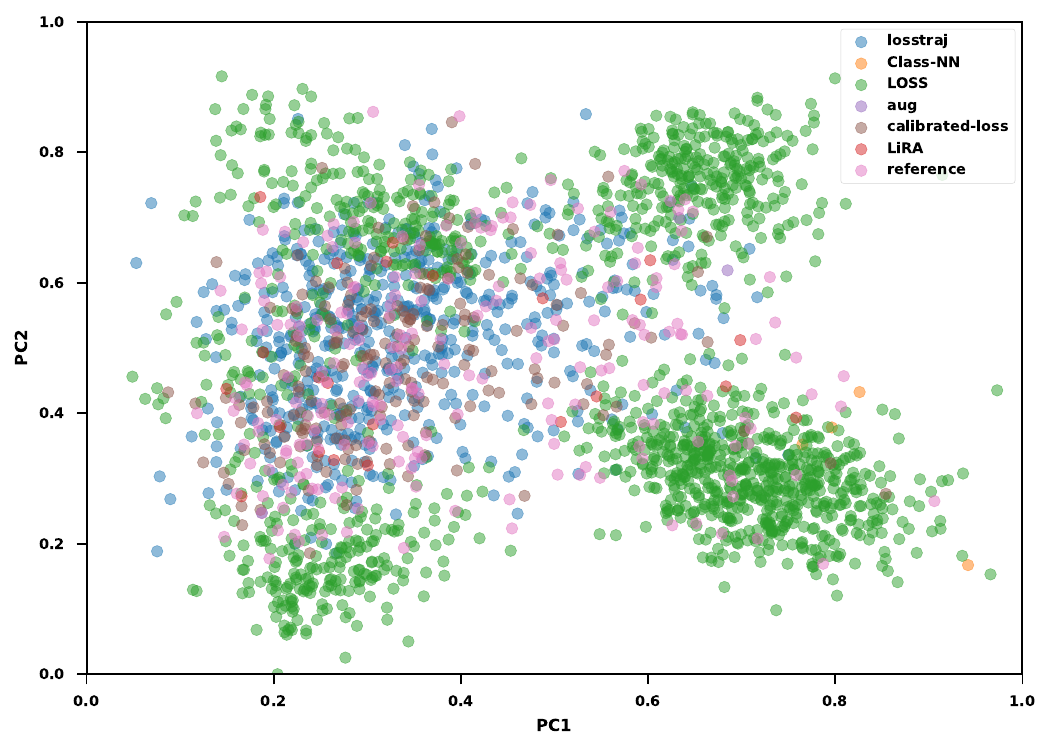}
    \caption{All Attacks}
    \label{fig:all-attack-pca}
    \end{subfigure}
    
    \vspace{1em}
    
    \caption{PCA of $\attack$-unique Members. We conduct PCA on all the target model $\targetModel$'s logits of all samples in target dataset $\targetDataset$, and plot $\attack$-unique members (defined in Section \ref{sec:evaluating-multiple-attacks}) for each attack. It is obtained with six instances at FPR@0.1.
    Figure ~\ref{fig:yeom-pca} and Figure ~\ref{fig:calibration-pca} are picked to demonstrate the distribution difference between two MIAs.
    The explained variance ratio is 48.56\%, and 15.95\% for the PC1 (first component) and PC2 (second component), respectively. 
    }
    \label{fig:pca}
\end{figure}
Figure~\ref{fig:pca} shows that the \textit{\(\attack\)-unique} samples identified by different attacks can exhibit different distributions in the model output space. This suggests that each MIA may implicitly favor certain groups or distributions of members, although PCA may only be able to partially capture the underlying characteristics. In particular,
as shown in Figures~\ref{fig:yeom-pca} and~\ref{fig:calibration-pca}, the samples uniquely identified by the Loss attack and those identified by the Augmentation attack form visibly different clusters.


\subsubsection{Attack Signals of $\attack$-Covered Samples}
\label{sec:top-x-signal-result}
In addition to difference in the distribution of detected samples in the model output space, disparities among MIAs also stems from their distinct methodologies and signals they exploit. Quantifying these methodological differences is challenging, as each attack employs different processes that are difficult to formalize within a unified framework. Therefore, we focus on examining how the signals obtained from the target and shadow models contribute to variation in member detection. Following the methodology in Section \ref{sec:evaluating-multiple-attacks}, we use "Top-x" Class-NN attack with varying signals.

\begin{figure}
    \centering
        \begin{subfigure}[b]{0.45\columnwidth}
        \includegraphics[width=\textwidth]{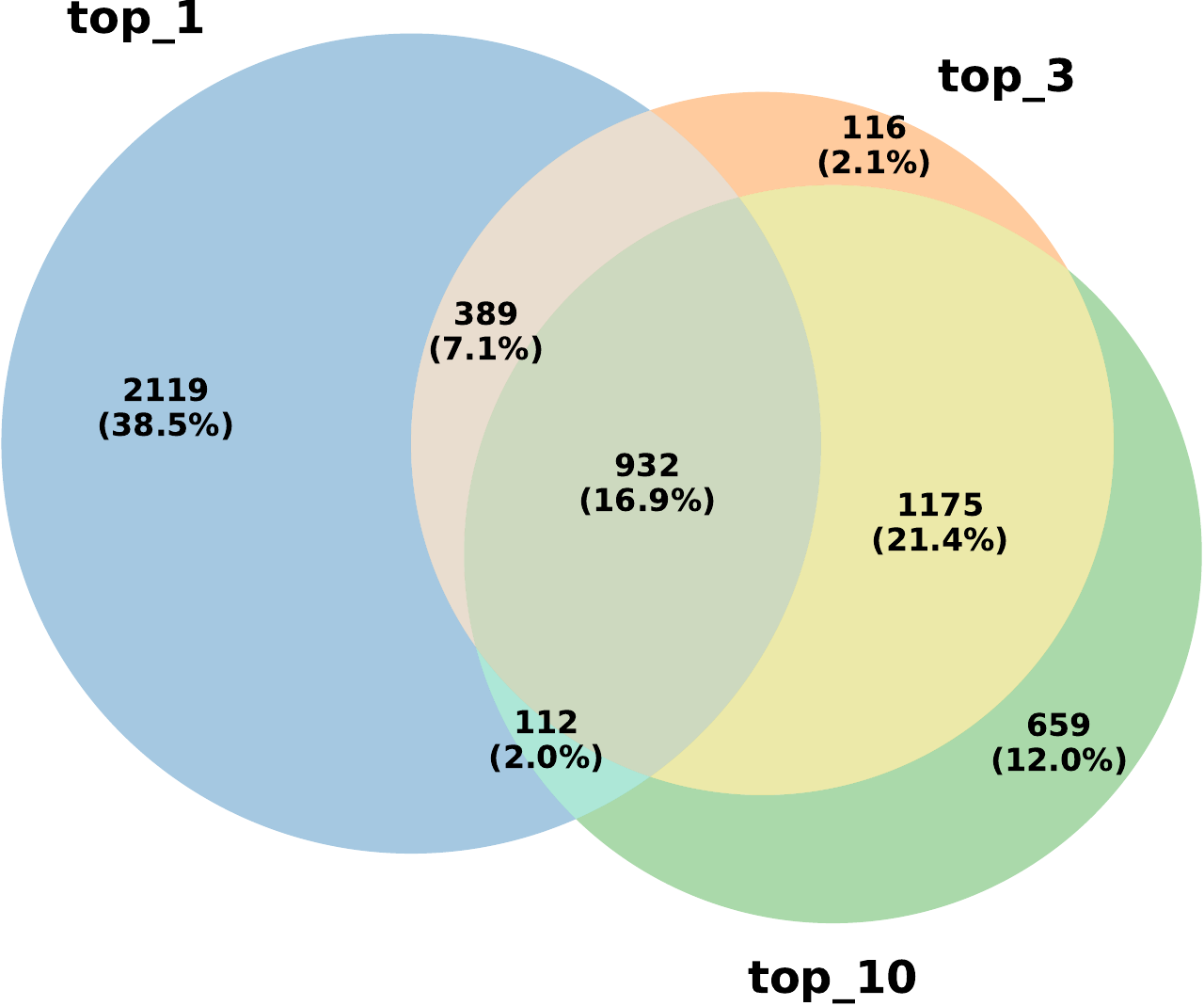}
        \caption{Top-x Class-NN Attack's Coverage}
        \label{fig:top-x-shokri-venn}
    \end{subfigure}
    \hfill
    \begin{subfigure}[b]{0.5\columnwidth}
        \includegraphics[width=\textwidth]{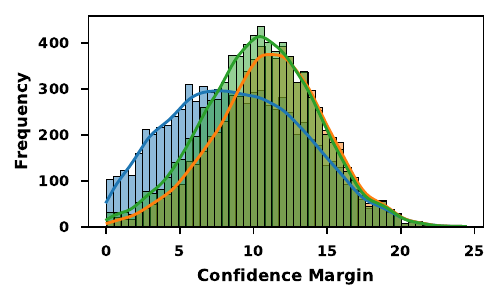}
        \caption{Distribution Shift of \\Confidence Margin}
        \label{fig:distribution-confidence-margin}
    \end{subfigure}
    \vspace{1em}   
    \caption{Top-x Class-NN Attack-Covered Samples with Different Signals. The coverage is calculated using 6 instances at FPR = 0.1.}
    \label{fig:same-attack-diff-signal}
    \vspace{-5pt}
\end{figure}

Figure~\ref{fig:top-x-shokri-venn} shows a Venn diagram illustrating the coverage of three variations: Top-1, Top-3, and Top-10 Class-NN attacks. We observe a greater overlap between Top-3 and Top-10 Class-NN attacks compared to their overlap with Top-1. This is because Top-3 and Top-10 attacks leverage more similar signals, as logits beyond the top 3 positions are typically close to zero and contribute little additional information.

To better understand the distribution difference of detected members under different top-x signals, we measure the \textbf{Confidence Margin} of the target model’s predictions for each member. The confidence margin is defined as the difference between the highest confidence score (representing the most likely class) and the second highest confidence score (representing the next most likely class) in the output of the target model. It represents how much more confident the model is in its top prediction relative to the closest alternative. Figure~\ref{fig:distribution-confidence-margin} presents the kernel density estimate (KDE) of confidence margin values for correctly predicted members, showing that less similar signals lead to greater disparity in the distribution of detected members.
Specifically, the confidence margin distributions for Top-3 and Top-10 signals are similar, while the distribution for Top-1 is more left-skewed and exhibits higher variance. This difference arises because the Class-NN attack with access to Top-3 or Top-10 logits can leverage additional information beyond the single highest logit used by Top-1. As a result, Top-3 and Top-10 attacks tend to identify more common members with larger confidence margins. These observations on confidence margin distributions align with the overlap patterns observed in the Venn diagram in Figure~\ref{fig:top-x-shokri-venn}.

The impact of attack signals extends beyond the Top-x Class-NN model. Existing MIAs rely on various signals including loss, confidence score vector, and loss trajectories. While the loss, computed from the confidence score vector, provides a scalar summary, it may omit finer details present in the vector itself which can be beneficial for membership inference. The loss trajectory offers insights into the training-time patterns that may reveal more nuanced information. These differences suggest that each type of signal carries unique information that may contribute to disparities in member detection across different MIAs.




\subsection{Practical Implications of MIA Disparities}\label{ssec:implication}
Given the significance of instance-level and method-level disparities in MIAs, caution must be exercised when using them for privacy assessment and performance evaluation. Evaluating each MIA separately or relying on a single instance may fail to capture the full spectrum of privacy leakage risks, leading to incomplete assessments. Below, we examine several privacy tasks where MIAs are commonly used and discuss the implications of MIA disparities.\\
\noindent\textbf{Privacy Auditing and Risk Assessment:} Several open-source toolkits have been developed for privacy assessment (e.g., \cite{kumar2020mlprivacy, 10.14778/3681954.3681994}), which often incorporate multiple MIA methods to assess privacy leakage in trained models. However, these tools typically run one instance per method and report risk based on population-level metrics derived from single-instance results. This overlooks both instance-level and method-level disparities, which may result in undetected vulnerable member samples and lead to overly optimistic privacy estimates.\\
\noindent\textbf{Machine Unlearning:} In many unlearning works (e.g., \cite{NEURIPS2023_062d711f, choi2023machineunlearningbenchmarksforgetting}), MIAs are used to evaluate the effectiveness of unlearning by checking how many samples from a forgetting set (a subset of the training set) remain identifiable after unlearning. A common practice is to instantiate an MIA with an auxiliary dataset and report results from that single instance. However, due to instance-level variability, a sample undetected by the reported instance may still be identified by others, potentially leading to overestimation of unlearning effectiveness. Similarly, relying on a single MIA method can overlook member samples that would be detected by other methods due to method-level disparities.\\
\noindent\textbf{Privacy Defense Evaluation:} Many defense mechanisms (e.g., \cite{10.5555/3698900.3699034, jia2019memguard}) are evaluated against a single MIA instance. The reported metric reflects exposure risk only for the members detected by that specific instance. However, due to randomness in shadow model training and method-specific biases, it is unclear whether a defense appears stronger simply because it performs better on the particular subset exposed by that instance. It remains uncertain whether the defense would perform similarly on samples exposed by other instances or MIA methods.

Thus, explicitly addressing MIA disparities is essential for fair and reliable evaluation of privacy risks and defenses.

\section{MIA Ensemble}
\label{sec:ensemble}
In this section, we propose an ensemble framework that employs various strategies to account for MIA disparities. This framework not only enables the construction of more powerful attacks but also provides an evaluation protocol for more comprehensive and reliable privacy assessments.

\subsection{Ensemble Strategies}
\subsubsection{Attack Stability Ensemble}
\label{attack-stability-ensemble}
Section~\ref{sec:coverage-and-stability-over-randomness} shows that MIA stability converges as more instances are aggregated, and this aggregation reduces false positives, resulting in lower FPR and improved precision. These findings suggest that stability captures members consistently identified across instances regardless of randomness, and can be leveraged within each attack to achieve higher precision.
Furthermore, Section \ref{sec:agreement-among-mia-methods} shows that the stability of different attacks has very little overlap, particularly at low FPR (Figure \ref{fig:trend-of-agreement-vs-fpr}).
Accordingly, we propose a 2-step ensemble approach:
\begin{itemize}[leftmargin=*, noitemsep,topsep=0pt,parsep=0pt,partopsep=0pt]
    \item[1)]\textbf{Multi-instance Stability}: This step uses multiple instances of an attack $\attack$ and uses the "logical and" (i.e., conjunction) to determine membership to improve precision. That is, a sample $x$ is regarded as a member only if all of $\attack$'s instances determine $x$ is a member.
    \item[2)] \textbf{Multi-attack Union}: This step capitalizes on the high precision achieved by multi-instance and the complementary nature of different attacks, which tend to identify distinct sets of members. By taking the "logical or" (i.e., disjunction) of prediction from multiple attacks from the multi-instance step, the ensemble can detect members across all MIAs' stable predictions while maintaining high precision.
\end{itemize}

Formally, let \( p_i^\attack(x) \) represent the membership prediction of \( i \)-th instance of attack \( \attack \) on sample $x$. Note that \( p_i^\attack(x) \) is a prediction that's thresholded to either $1$ or $0$. The multi-instance prediction \( P^\attack_{n} \) is the conjunction of \(\{p^\attack_1, \ldots, p^\attack_n\}\) for attack \( \attack \). The multi-attack prediction \( P_n^{\{\attack_1, \ldots, \attack_m\}} \) is the disjunction of multi-instance predictions \(\{P^{\attack_1}_n, \ldots, P^{\attack_m}_n\}\).
\begin{gather}
\small
    P^{\attack_j}_n(x) = \bigwedge_{i=1}^n p^\attack_i(x) \label{eq:multi-instances-stability} \\
    P_n^{\{\attack_1, \ldots, \attack_m\}}(x) = \bigvee_{j=1}^m P^{\attack_j}_n(x) \label{eq:multi-attacks-stability}
\end{gather}

\subsubsection{Attack Coverage Ensemble}
\label{sec:attack-coverage-ensemble}
The previous attack ensemble strategy applies multi-instance intersection to improve reliability and precision, however, at the cost of coverage. In contrast, the attack coverage strategy here applies the multi-instance union to improve the coverage, followed by the same multi-attack union step. Similarly, we can describe this approach as follows:
\begin{align}
\small
    &\text{1) Multi-instance Coverage:}& P^{\attack_j}_n(x) &= \bigvee_{i=1}^n p^\attack_i(x) \label{eq:multi-instances-coverage} \\
    &\text{2) Multi-attack Union:}& P_n^{\{\attack_1, \ldots, \attack_m\}}(x) &= \bigvee_{j=1}^m P^{\attack_j}_n(x) \label{eq:multi-attack-coverage}
\end{align}

\subsubsection{Attack Majority Ensemble}
\label{sec:attack-mv-ensemble}
While coverage and stability represent two extremes—capturing all potential risks and the most consistently vulnerable samples, respectively—the majority-voting ensemble offers a balanced alternative. This strategy captures samples that are identified as members by the majority of the running instances of a given MIA method. Formally, we have:
\begin{align}
\small
&\text{1) Multi-instance Majority Voting:} \nonumber \\
& \qquad \qquad \qquad \qquad\qquad \qquad  P^{\attack_j}_n(x) = \left(\sum_{i=1}^n p^\attack_i(x)\right) > \frac{n}{2}  \label{eq:multi-instances-mv}   &\\
&\text{2) Multi-attack Union:} \qquad\qquad P_n^{\{\attack_1, \ldots, \attack_m\}} = \bigvee_{j=1}^m P^{\attack_j}_n & \label{eq:multi-attack-mv}
\end{align}

\subsection{Evaluation}
\label{sec:ensemble-evaluation}
For the ensemble, we consider four attacks: Difficulty Calibration Loss Attack, Reference Attack, LiRA, and Loss Trajectory Attack, because out of our seven implemented attacks, only these four improve the precision with stability over multiple instances at a low FPR, as shown in Figure~\ref{fig:intersection_precision_conv}. As in our previous setup, we utilize six instances of each MIA for the ensemble.
Our proposed ensemble operates on membership predictions rather than membership scores. Therefore, to measure its performance in the TPR-FPR plane, we vary the FPR of base instances with 100 different FPR values, ranging from $10^{-6}$ to 1, evenly spaced on a logarithmic scale. Under each instance FPR, we compute the predictions by ensemble and derive the corresponding TPR and FPR values for the ensemble. 

\begin{figure*}[h]
    \centering
    \begin{minipage}{0.8\textwidth}
        \centering
        \begin{subfigure}[t]{0.33\textwidth} 
            \includegraphics[width=\textwidth]{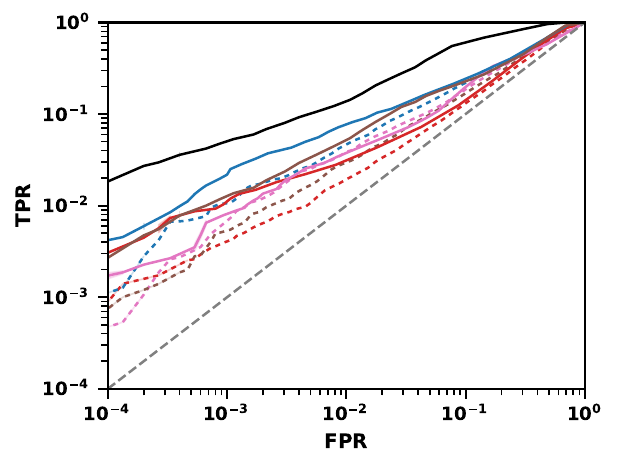}
            \caption{Stability Ensemble}
            \label{fig:ensemble-roc-stability}
        \end{subfigure}
        \begin{subfigure}[t]{0.33\textwidth} 
            \includegraphics[width=\textwidth]{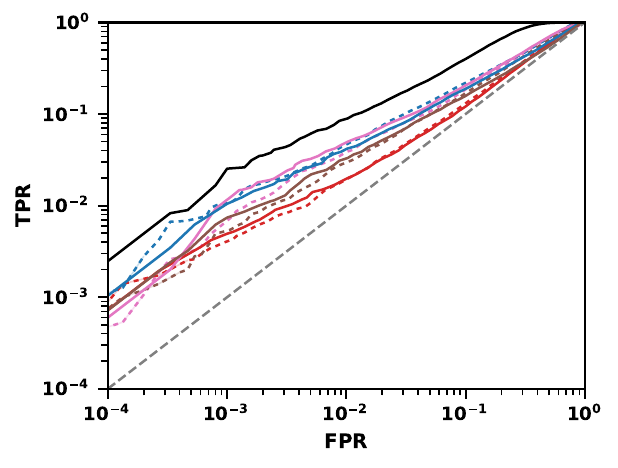}
            \caption{Coverage Ensemble}
            \label{fig:ensemble-roc-coverage}
        \end{subfigure}
            \begin{subfigure}[t]{0.33\textwidth} 
            \includegraphics[width=\textwidth]{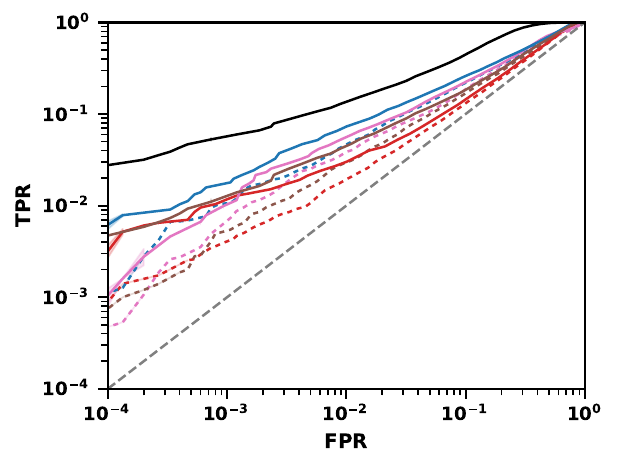}
            \caption{Majority Voting Ensemble}
            \label{fig:ensemble-roc-majority-voting}
        \end{subfigure}
    \end{minipage}
    \begin{minipage}{0.15\textwidth} 
        \centering
        \begin{figure}[H]
            \flushleft
            \includegraphics[width=0.9\textwidth]{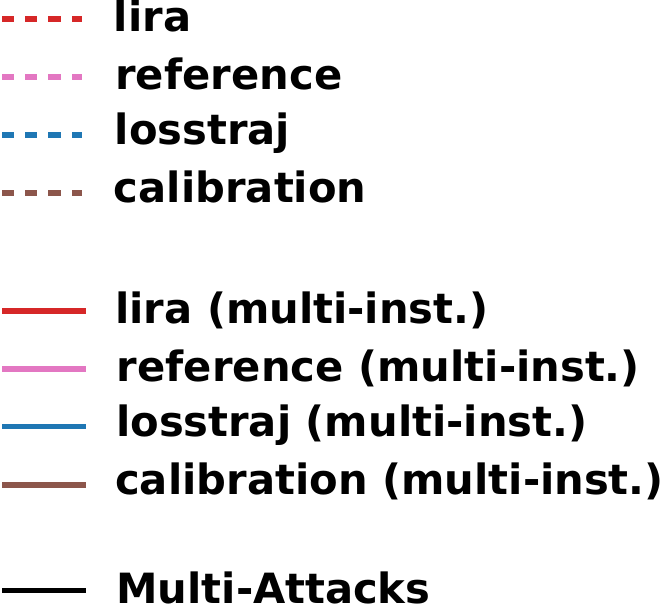}
        \end{figure}
    \hfill
    \end{minipage}
    \vspace{1em}
    \caption{\textbf{ROC Curve for Ensemble.} Dashed lines show single-instance ROC, solid lines show multi-instance ROC and the black line represents the complete four-attack ensemble.}
    \label{fig:ensemble-roc}
\end{figure*}

In Figure~\ref{fig:ensemble-roc}, we observe that the TPR of all three ensembles consistently outperforms single-instance and \emph{multiple-instance} methods in the TPR-FPR plane. Here, the \emph{multi-instance} method refers to the ensemble approach without the multi-attack union step, i.e., only using (\ref{eq:multi-instances-stability}), (\ref{eq:multi-instances-coverage}), or (\ref{eq:multi-instances-mv}). Interestingly, we find that the multi-instance method alone often outperforms its single-instance counterpart, particularly when using stability or majority-voting strategies. This further demonstrates that relying on a single MIA instance for evaluation underestimates the true privacy risks, as, in real-world scenarios, multiple MIA instances could be generated by the same or different attackers, and inherent instance-level disparities in membership inference persist. Additionally, we evaluate all possible combinations of the four attacks and compare their ROC curves. As shown in Appendix Figure~\ref{fig:ensemble_roc_appendix}, the full ensembles leveraging all four attacks consistently achieve higher TPR across all FPR values compared to ensembles using fewer or different combinations of attacks.

Table~\ref{tab:ensemble-perf-table} further compares our full ensembles and multi-instance-only ensembles against each single MIA instance in terms of AUC, accuracy, and TPR at 0.1\% FPR (see Appendix Table~\ref{tab:ensemble-perf-table-appendix} for results on Texas100 and Purchase100). We choose FPR = 0.1\% to showcase the ensemble's capabilities under low FPR conditions, aligning with evaluation metrics used in recent works~\cite{lira, losstraj}. The results are based on the ResNet-56 architecture, and comparisons for other model architectures can be found in Appendix Table~\ref{tab:ensemble-perf-table-all-other-model-appendix}. Across all settings, the final three rows in Table~\ref{tab:ensemble-perf-table} show that all three full ensemble strategies consistently outperform individual instances under three traditional performance metrics. Compared to single-instance attacks, the multi-instance-only ensemble (denoted as ‘Multi-inst’ in the table) shows improved performance under both stability and majority-voting strategies. However, it underperforms under the coverage-based ensemble, where only the multi-instance Reference attack shows a slight performance improvement. By comparing the full ensembles with the multi-instance-only ensembles, we observe that the benefit gained from multi-attack union often exceeds that achieved through multi-instance aggregation alone.

\begin{table}
\centering
\scriptsize
\scalebox{0.82}{
\renewcommand{\arraystretch}{1.2}
\begin{tabular}{ll|ccc|ccc|ccc}
\toprule
& & \multicolumn{3}{c|}{\textbf{CIFAR-10}} 
  & \multicolumn{3}{c|}{\textbf{CIFAR-100}} 
  & \multicolumn{3}{c}{\textbf{CINIC-10}} \\
\cmidrule(lr){3-5}\cmidrule(lr){6-8}\cmidrule(lr){9-11}
\textbf{Ens. Lvl} & \textbf{Attack} 
    & \textbf{AUC} & \textbf{ACC} & \textbf{TPR} 
    & \textbf{AUC} & \textbf{ACC} & \textbf{TPR} 
    & \textbf{AUC} & \textbf{ACC} & \textbf{TPR} \\
\midrule
\multirow{4}{*}{Single-inst.}
& losstraj    
  & 0.635 & 0.588 & 0.002
  & 0.852 & 0.771 & 0.042
  & 0.673 & 0.619 & 0.008 \\
& reference    
  & 0.608 & 0.596 & 0.010
  & 0.813 & 0.774 & 0.034
  & 0.615 & 0.605 & 0.005 \\
& lira         
  & 0.585 & 0.570 & 0.005
  & 0.802 & 0.729 & 0.034
  & 0.601 & 0.578 & 0.003 \\
& calibration  
  & 0.603 & 0.572 & 0.005
  & 0.721 & 0.676 & 0.008
  & 0.623 & 0.593 & 0.002 \\
\midrule
\multirow{4}{*}{\shortstack{Multi-inst.\\Coverage}}
& losstraj    
  & 0.595 & 0.557 & 0.010 
  & 0.788 & 0.695 & 0.032
  & 0.643 & 0.592 & 0.006 \\
& reference
  & 0.638 & 0.600 & 0.012
  & 0.863 & 0.783 & 0.045
  & 0.652 & 0.619 & 0.005\\
& lira         
  & 0.570 & 0.561 & 0.005
  & 0.801 & 0.729 & 0.036
  & 0.590 & 0.576 & 0.003 \\
& calibration  
  & 0.551 & 0.532 & 0.007
  & 0.652 & 0.607 & 0.010
  & 0.617 & 0.589 & 0.003 \\
\midrule
\multirow{4}{*}{\shortstack{Multi-inst.\\Stability}}
& losstraj    
  & 0.659 & 0.610 & 0.022
  & 0.898 & 0.817 & 0.108
  & 0.685 & 0.630 & 0.013 \\
& reference    
  & 0.582 & 0.582 & 0.008
  & 0.765 & 0.772 & 0.054
  & 0.585 & 0.582 & 0.005 \\
& lira         
  & 0.603 & 0.588 & 0.011
  & 0.839 & 0.764 & 0.087
  & 0.623 & 0.593 & 0.009 \\
& calibration  
  & 0.650 & 0.620 & 0.013
  & 0.855 & 0.763 & 0.031
  & 0.643 & 0.616 & 0.003 \\
\midrule
\multirow{4}{*}{\shortstack{Multi-inst.\\Maj-vote}}
& losstraj    
  & 0.668 & 0.602 & 0.018 
  & 0.877 & 0.786 & 0.061 
  & 0.712 & 0.647 & 0.008 \\
& reference    
  & 0.636 & 0.615 & 0.011 
  & 0.860 & 0.806 & 0.070
  & 0.637 & 0.626 & 0.006 \\
& lira         
  & 0.596 & 0.579 & 0.012
  & 0.840 & 0.766 & 0.068
  & 0.621 & 0.595 & 0.008 \\
& calibration  
  & 0.629 & 0.586 & 0.013 
  & 0.741 & 0.672 & 0.018
  & 0.634 & 0.603 & 0.003 \\
\midrule
\multirow{2}{*}{Multi-attack}
& Coverage     
  & 0.841 & 0.781 & 0.025
  & 0.913 & 0.835 & 0.077
  & \textbf{0.898} &\textbf{ 0.834} & \textbf{0.014} \\
& Stability    
  & \textbf{0.863} & 0.770 & 0.050
  & \textbf{0.978} & \textbf{0.935} & \textbf{0.280}
  & 0.807 & 0.740 & 0.009 \\
& Maj-vote    
  & 0.858 & \textbf{0.789} & \textbf{0.056}
  & 0.961 & 0.897 & 0.208
  & 0.865 & 0.800 & \textbf{0.014} \\
\bottomrule
\end{tabular}
} 

\vspace{0.5cm}
\caption{\textbf{Performance of ensembles with four attacks vs. single instance attacks}. TPR is measured at 0.1\% FPR.}
\label{tab:ensemble-perf-table}
\vspace{-5pt}
\end{table}

\begin{figure}[H]
    \centering
        \begin{subfigure}{0.45\columnwidth}
        \includegraphics[width=\textwidth]{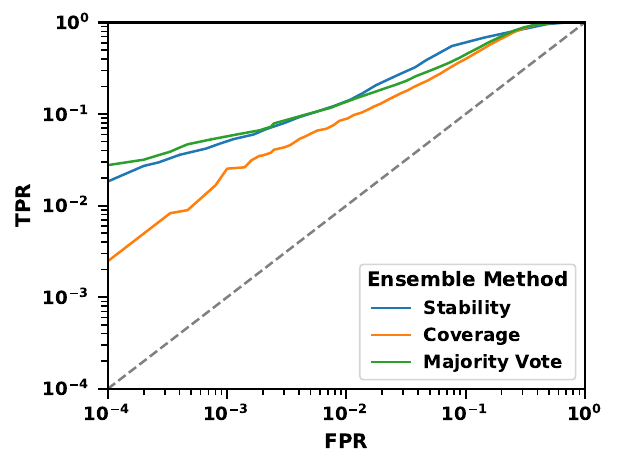}
        \caption{Log Scale}
        \label{fig:compare-ensemble-methods-roc-log-scale}
    \end{subfigure}
    \hfill
    \begin{subfigure}{0.45\columnwidth}
        \includegraphics[width=\textwidth]{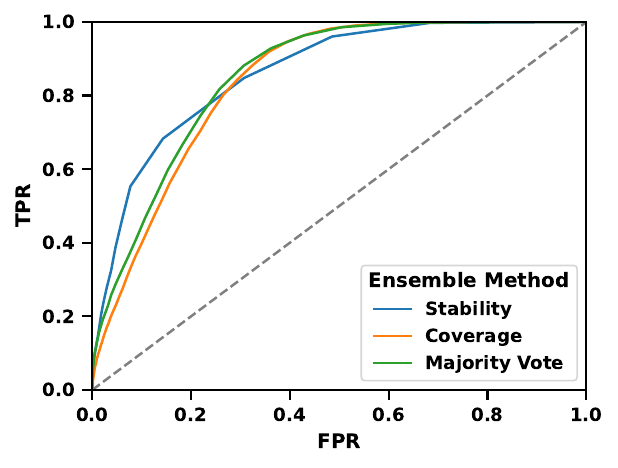}
        \caption{Linear Scale}
        \label{fig:compare-ensemble-methods-roc-linear-scale}
    \end{subfigure}
    \vspace{1em}   
    \caption{ROC Curves of Different Ensemble Strategies.}
    \label{fig:compare-ensemble-methods-roc}
\end{figure}

While all three full ensembles achieve improved performance, each exhibits unique strengths across different FPR ranges. Figure ~\ref{fig:compare-ensemble-methods-roc} shows their ROC curves side by side. From linear-scale ROC in Figure~\ref{fig:compare-ensemble-methods-roc-linear-scale}, we can see that the Stability Ensemble outperforms the Coverage Ensemble in the lower FPR region (FPR < 0.3), while the Coverage Ensemble achieves a higher TPR in the higher FPR region (FPR > 0.3). On log-scale ROC, we can see that the Majority Voting performs comparably to the Stability Ensemble at low FPR (also demonstrated in Table~\ref{tab:ensemble-perf-table}), and also exceeds the Stability's performance in the high FPR region. These trends align with the design of each ensemble method.
Coverage tends to cover more potential risks at the cost of increased FPR, stability focuses on consistently identifying vulnerabilities with high precision, while Majority Voting balances their strengths.

\subsection{Ensemble in Practice}

\subsubsection{Optimization Strategies for Ensemble}
\label{sec:optimization-strategies-for-ensemble}
The ensemble framework leverages both multi-instance and multi-attack approaches, achieving comprehensive coverage of privacy risks at the expense of increased computational cost. Below, we discuss practical strategies to mitigate this computational overhead. 

\noindent\textbf{Low-Cost Attack as an Add-on}. Among the four attacks we examined, the Difficulty Calibration Loss Attack requires much less time to prepare than the others, requiring only a single shadow model. This makes it an ideal add-on attack.

\noindent\textbf{Attacks Sharing the Same Process}. Many membership inference attacks share similar, if not identical, preparation processes. For example, LIRA and the Reference Attack both rely on the same shadow model training process (as detailed in Appendix Section~\ref{sec:appe_setup-for-mias}). In our experiments, LIRA and the Reference Attack utilized the same 20 shadow models, making their ensemble nearly as cost-effective as preparing just one of them. This ensemble identified approximately twice as many members as either individual attack. Similarly, the Difficulty Calibration Loss Attack can serve as a ``free'' add-on if another attack already involves training a shadow model, as it only requires one shadow model to calibrate the MIA score \cite{diff_calibration}.

\subsubsection{Cost Analysis}
\label{sec:cost-analysis}
\begin{figure*}
    \centering
    \begin{minipage}{0.8\textwidth}
        \centering
        \begin{subfigure}[t]{0.32\textwidth} 
            \includegraphics[width=\textwidth]{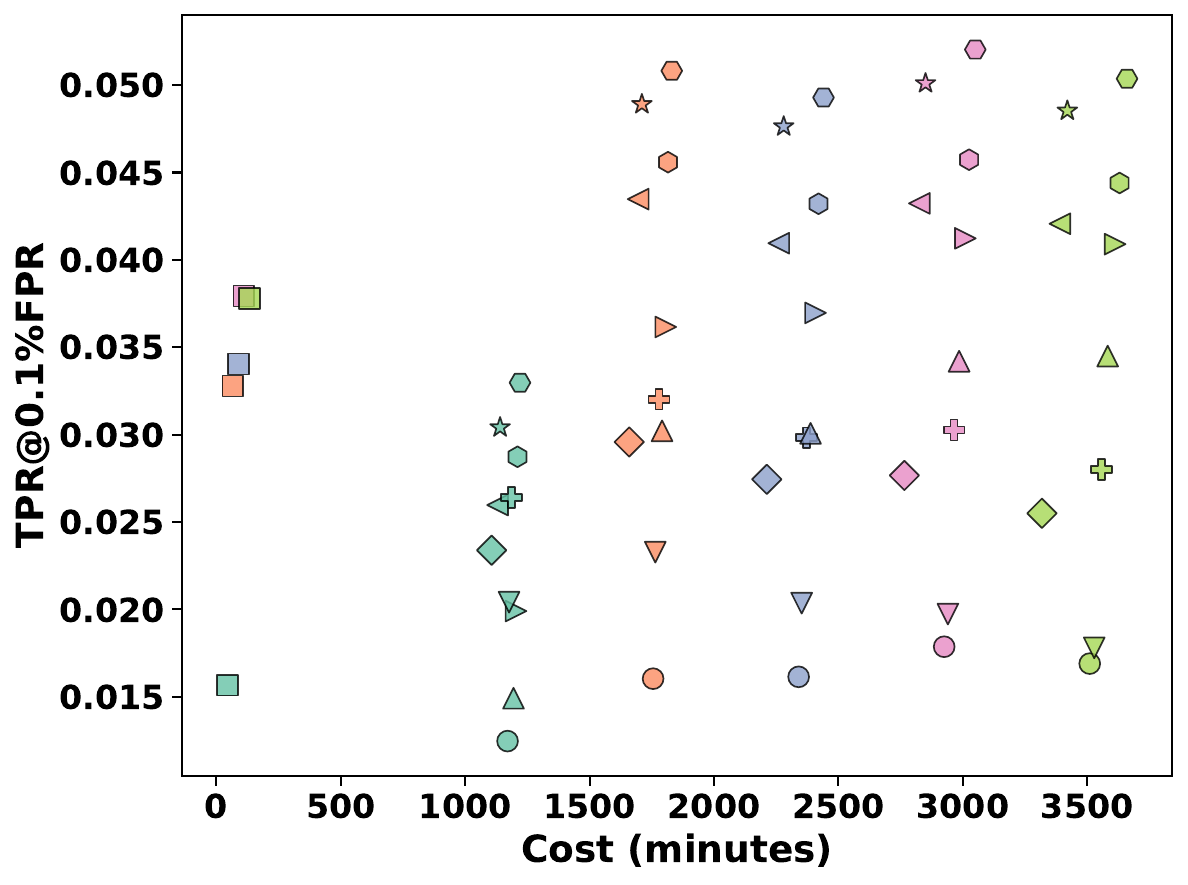}
            \caption{Stability Ensemble}
            \label{fig:cifar10_stability_cost}
        \end{subfigure}
        \begin{subfigure}[t]{0.32\textwidth} 
            \includegraphics[width=\textwidth]{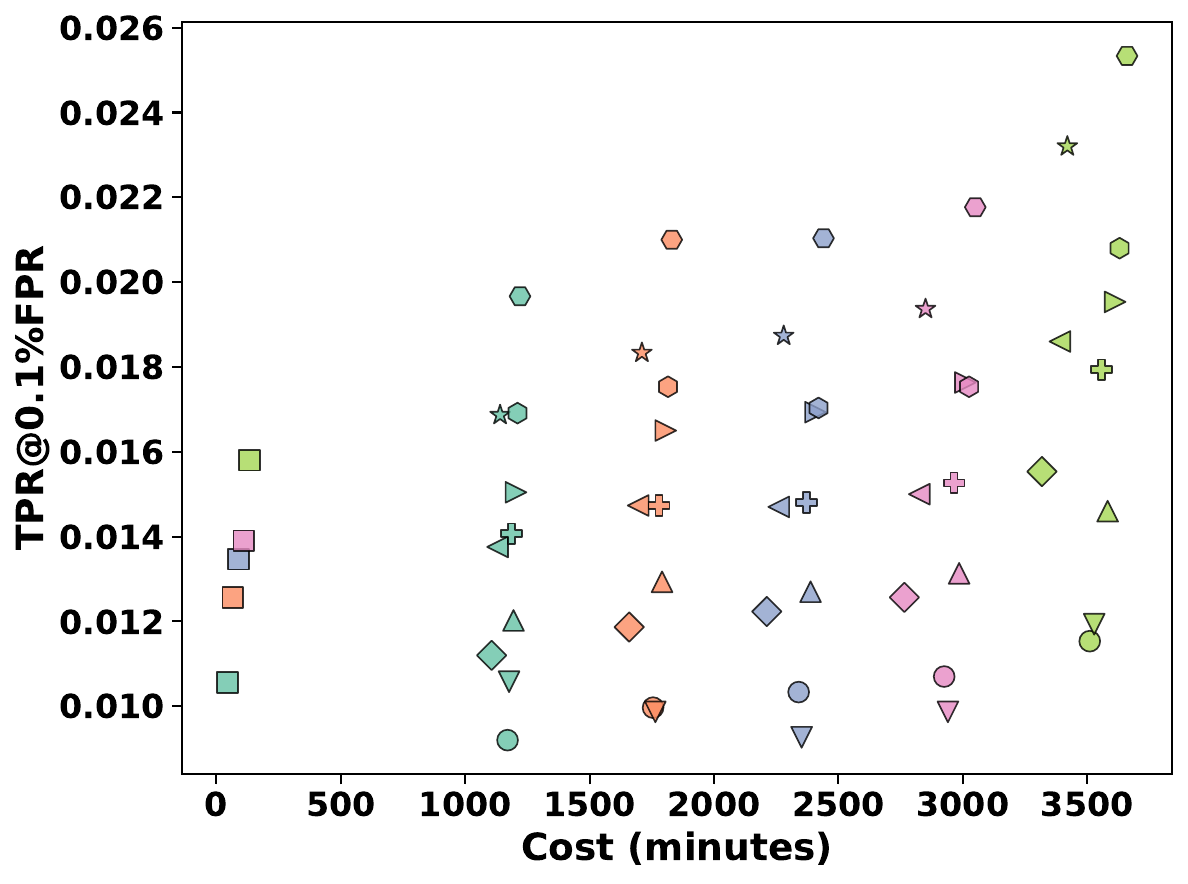}
            \caption{Coverage Ensemble}
            \label{fig:cifar10_coverage_cost}
        \end{subfigure}
            \begin{subfigure}[t]{0.32\textwidth} 
            \includegraphics[width=\textwidth]{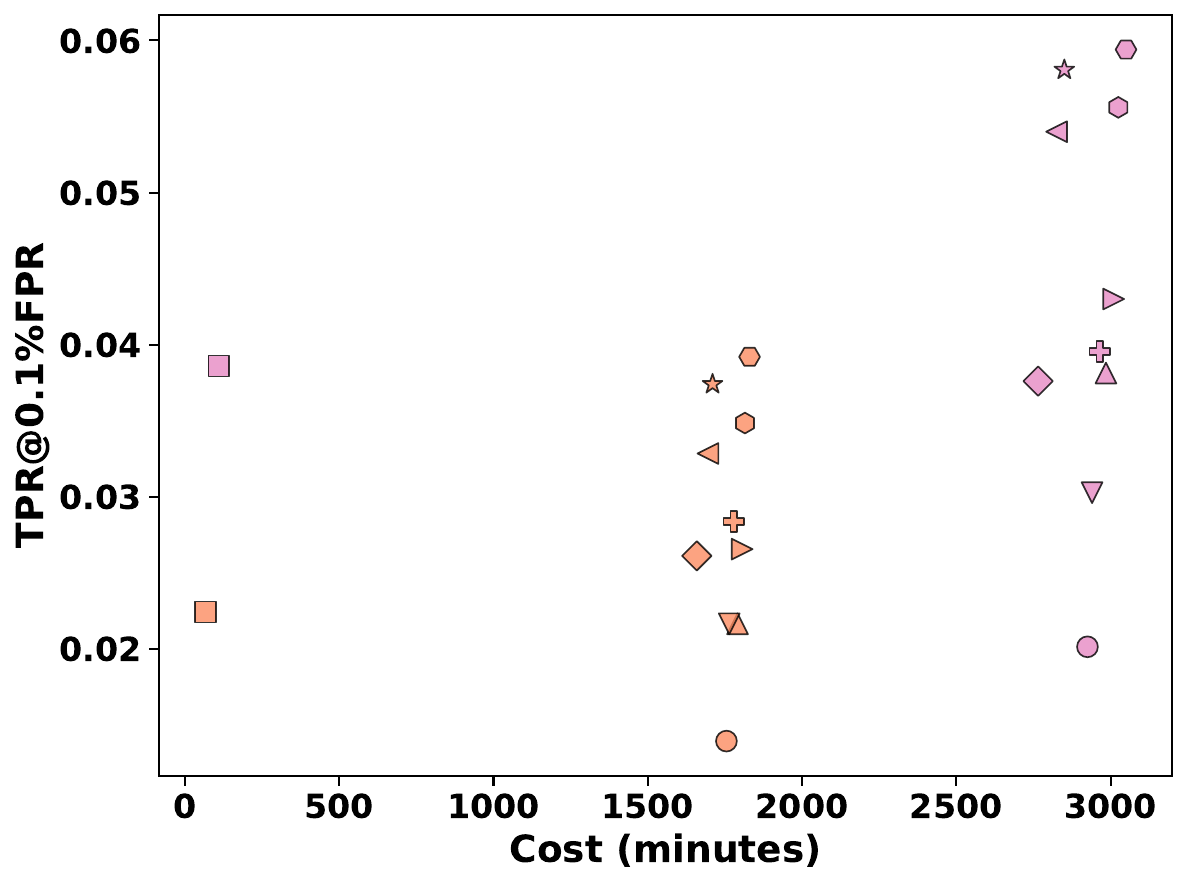}
            \caption{Majority Voting Ensemble}
            \label{fig:cifar10_majority_voting_cost}
        \end{subfigure}
    \end{minipage}
   \begin{minipage}{0.35\columnwidth} 
        \centering
        \includegraphics[width=\textwidth]{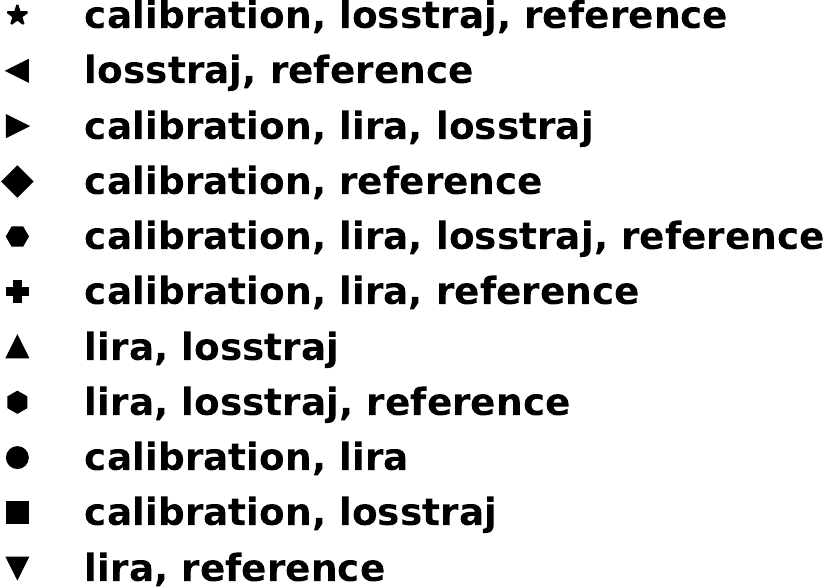}
        
        \vspace{0.2cm} 
        
        \flushleft
        \includegraphics[width=0.4\textwidth]{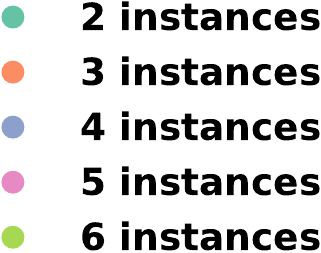}
    \end{minipage}

    \vspace{0.3cm}

    \caption{\textbf{Performance vs. Cost Analysis for CIFAR-10 using different ensembles.}}
    \label{fig:cost-analysis-body}
\end{figure*}
We measure the computation cost of ensembles in GPU hours for each MIA instance, considering different numbers of instances per ensemble. 
When both LiRA and the Reference Attack are included in an ensemble combination, we apply the above optimization strategy to combine and deduct their shadow model training time. 
The Majority Voting Ensemble is evaluated with odd numbers of instances to avoid ties in voting.

Figure~\ref{fig:cost-analysis-body} presents cost (in GPU time) v.s. performance (in TPR @0.1\%FPR) given different numbers of instances and different combinations of attacks. A more detailed description and study of the cost is provided in the Appendix Section~\ref{sec:app-cost-analysis}.
Overall, we observe a positive correlation between computation cost and performance. Notably, ensembles involving all four attacks consistently achieve the best performance, underscoring the importance of combining multiple attack methods in an ensemble. From additional experiments across different datasets, we conclude that this trend holds true for Stability and Majority Voting Ensembles but does not always apply to Coverage Ensembles.

Additionally, when comparing configurations with similar performance, we observe that cost-effective options often exist, achieving target TPRs with minimal GPU time (indicated by the leftmost points on a given TPR line). For example, in Figure~\ref{fig:cifar10_stability_cost}, with target TPR=0.05, an ensemble of four attacks with three instances achieves the same performance as using six instances, effectively reducing the training time by half, from around 3500 mins to 1700 mins. This significant reduction in computation cost demonstrates that a careful selection of attack combinations and instance counts can achieve similar levels of effectiveness without incurring unnecessary overhead. Practitioners may find their desired ensemble configuration to achieve robust privacy evaluations given a resource budget. We leave efficiently identifying optimal configurations for future work.

\section{Discussion}
The instance-level and method-level disparities among MIAs, along with the performance gains achieved through ensemble strategies, highlight the practical relevance of MIA disparities and the risk of underestimating privacy vulnerabilities in the tasks discussed in Section~\ref{ssec:implication}. In this section, we discuss actionable directions for addressing these issues in future MIA research.

\noindent\textbf{MIA performance evaluation and development.} We advocate for incorporating disparity analysis into the development and evaluation of MIAs, using our proposed coverage and stability measures to examine and quantify how an MIA differs from others in member detection. These measures offer a complementary perspective to traditional population-level metrics such as AUC and TPR@Low FPR by providing additional insight into the extent of privacy risks an attack can expose (via coverage) and the consistency with which it reveals those risks across different runs (via stability).

An MIA that achieves similar or even lower population-level metrics, such as AUC, may still hold significant value if it detects a substantially different subset of members—indicating high disparity—which can be revealed through our disparity analysis based on coverage and stability. This diversity enhances our understanding of privacy vulnerabilities by uncovering risks that other attacks may miss. Our ensemble results further support this insight, demonstrating that combining multiple attacks—including those traditionally considered “weaker”—often leads to improved overall performance. Therefore, MIAs with high disparity contribute to a more complete and robust assessment of privacy risks, especially when leveraged through our proposed ensemble strategies.

\noindent\textbf{Complete and reliable privacy evaluation.} As discussed in Section~\ref{ssec:implication}, the common practice of single-instance-based evaluation is insufficient and may lead to unreliable conclusions. Our ensemble framework addresses this by capturing the full spectrum of privacy risks posed by different MIA methods through multi-instance and multi-attack ensembles based on coverage, stability, and majority voting. Integrating these ensemble strategies into privacy evaluations—for instance, using ensemble attacks against defensive models— ensures a more accurate and robust assessment of privacy defenses. This approach can similarly enhance evaluations of unlearning mechanisms. Given this, it may also be necessary to revisit prior evaluations of MIA defenses and unlearning methods that relied solely on a single random MIA instance.

\noindent\textbf{Additional Caveats in Using MIA for Privacy Evaluation.} 
Recently, there has been significant interest in MIAs against large language models (LLMs). However, several works~\cite{das2025blindbaselinesbeatmembership, meeus2025sokmembershipinferenceattacks, zhang2025membershipinferenceattacksprove} have raised concerns about the construction of evaluation datasets, particularly regarding the distribution shift between member and non-member samples. In many LLM MIA evaluations, non-member data were collected from web content published after the model’s training cutoff date, such that member samples originate from the training data distribution, while non-member samples come from a different and later distribution (e.g., different time periods). Such distribution shifts can artificially inflate MIA performance, as attacks may exploit these distribution differences rather than truly detecting membership status. Consequently, evaluations based on such datasets may overstate privacy risks.

    
For privacy defense evaluation, \citet{aerni2024misleading} argues that prior evaluations using MIAs, which report attack performance averaged across all training samples, can be misleading because they may fail to reflect a defense’s effectiveness against the most vulnerable examples. In addition, some evaluations have relied on relatively weak, non-adaptive attacks, potentially overstating the robustness of the proposed defenses.

Our study is orthogonal to these concerns by addressing a different overlooked issue: the instance-level and method-level disparities among MIAs, which are often neglected in current evaluation practices. Addressing these disparities requires a more holistic evaluation protocol, such as our proposed ensemble framework, to enable more complete and reliable privacy assessments.

\section{Conclusion}
\label{sec:conclusion}
In this paper, we have provided critical insights into the disparities of Membership Inference Attacks (MIAs). Our findings challenge conventional evaluation methods of MIAs and highlight the impact of randomness and disparity in these MIAs.
Additionally, our proposed ensemble framework not only enables the construction of more powerful attacks but also offers a more comprehensive evaluation methodology.
Moving forward, our goal is to develop more sophisticated strategies for ensemble, further improving the efficiency and effectiveness of MIA.

\section*{Acknowledgements}
We thank all the anonymous reviewers and our shepherd for their insightful feedback and valuable suggestions. This work was supported in part by IBM-RPI AIRC A72193.

\bibliographystyle{ACM-Reference-Format}
\bibliography{bib}


\appendix




\section{Experiment Setup Details}
\label{sec:app_experiment_setup}

\subsection{Datasets}
\label{sec:app_dataset_details}
We use five datasets in our experiments: CIFAR-10, CIFAR-100, CINIC-10, Purchase100, and Texas100. 

\begin{itemize}[leftmargin=*,noitemsep,topsep=0pt,parsep=0pt,partopsep=0pt]
    \item \textbf{CIFAR-10}: Consists of 60,000 32x32 color images in 10 classes, with 50,000 training and 10,000 testing samples.

    \item \textbf{CIFAR-100}: Similar to CIFAR-10, but contains 100 classes with 600 images per class.

    \item \textbf{CINIC-10}: Extends CIFAR-10 to include 270,000 images from both CIFAR-10 and ImageNet. For experiments, we use a balanced subset of 30,000 CIFAR-10 images and 30,000 ImageNet images. Throughout the paper, we shuffle these two subsets.

    \item \textbf{Purchase100}: A structured dataset derived from Kaggle’s “Acquire Valued Shoppers” challenge, representing 197,324 shopping records. We use a subset of 60,000 samples, each with 600 binary features.

    \item \textbf{Texas100}: Contains hospital discharge records from the Texas Department of State Health Services. We use 60,000 samples from the dataset, predicting the 100 most frequent procedures.
\end{itemize}

\subsection{Models}
\label{sec:app_model_setup}
\begin{table*}[htbp]
\centering
\setlength{\tabcolsep}{3pt} 
\captionsetup{justification=centering} 

\begin{subtable}{\textwidth}
\centering
\scalebox{0.8}{
\begin{tabular}{l|ccc|ccc|ccc}
\toprule
Target Model & \multicolumn{3}{c}{CIFAR-10} & \multicolumn{3}{c}{CIFAR-100} & \multicolumn{3}{c}{CINIC-10} \\
& Train acc & Test acc & Gen Gap & Train acc & Test acc & Gen Gap & Train acc & Test acc & Gen Gap \\
\midrule
ResNet-56 & 89.3\%$\pm$6.1\% & 80.9\%$\pm$2.3\% & 8.4\% & 87.0\%$\pm$16.3\% & 49.6\%$\pm$4.8\% & 37.4\% & 80.4\%$\pm$0.3\% & 60.5\%$\pm$0.4\% & 19.9\% \\
VGG-16 & 96.7\%$\pm$5.3\% & 84.4\%$\pm$1.7\% & 12.4\% & 90.8\%$\pm$15.8\% & 53.9\%$\pm$5.0\% & 36.9\% & 98.8\%$\pm$0.2\% & 65.5\%$\pm$0.4\% & 33.3\% \\
MobileNetV2 & 90.9\%$\pm$7.6\% & 73.3\%$\pm$2.7\% & 17.6\% & 88.1\%$\pm$20.6\% & 39.7\%$\pm$7.4\% & 48.4\% & 88.4\%$\pm$0.5\% & 53.7\%$\pm$0.5\% & 34.7\% \\
WideResNet-32 & 83.3\%$\pm$2.7\% & 75.9\%$\pm$2.9\% & 7.4\% & 64.3\%$\pm$10.0\% & 41.6\%$\pm$3.1\% & 22.7\% & 66.9\%$\pm$2.0\% & 56.6\%$\pm$0.13\% & 10.3\% \\
\bottomrule
\end{tabular}
}
\vspace{0.5cm}

\caption{Training statistics for existing models on CIFAR-10, CIFAR-100, and CINIC-10 datasets.}
\label{table:existing_models}
\end{subtable}

\begin{subtable}{\textwidth}
\centering
\scalebox{0.8}{
\begin{tabular}{l|ccc|ccc}
\toprule
Target Model & \multicolumn{3}{c}{Texas100} & \multicolumn{3}{c}{Purchase100}\\
& Train acc & Test acc & Gen Gap & Train acc & Test acc & Gen Gap \\
\midrule
MLP & 99.9\%$\pm$0.0\% & 54.5\%$\pm$0.3\% & 45.4\% & 100.0\%$\pm$0.0\% & 78.6\%$\pm$0.4\% & 21.4\% \\
\bottomrule
\end{tabular}
}
\vspace{0.5cm}
\caption{Training statistics for the MLP model on Purchase-100 and Texas-100 datasets.}
\label{table:mlp_new_datasets}
\end{subtable}

\vspace{0.5cm}

\caption{\textbf{Target model's training statistics across various datasets and architectures.} 
All accuracies are averaged from 4 different sets of experiments, each with a distinct partition of the target and auxiliary datasets (available to the attacker). All models are trained and tested on 15,000 disjoint samples. Gen Gap represents the generalization gap between Top-1 training accuracy and testing accuracy. }
\label{table:train_test_acc}
\end{table*}
We utilize ResNet-56~\cite{resnet}, MobileNetV2~\cite{mobilenet}, VGG-16~\cite{vgg16}, and WideResNet-32~\cite{wrn} as model architectures. ResNet-56 is the primary architecture for reporting results due to its small generalization gap, making it a harder case for MIAs~\cite{Li_2021, chen2022relaxlossdefendingmembershipinference}. For the tabular datasets Purchase100 and Texas100, we use a 4-layer MLP with layer units=[512, 256, 128, 64]. The target models' performances on these datasets are described in \ref{table:train_test_acc}.

\mypara{Optimization and Training.} 
All models are trained using SGD with a momentum of 0.9 and an initial learning rate of 0.1, with a cosine learning rate scheduler~\cite{sgdr}. Target models and shadow models are trained for 60 epochs on CIFAR-10 and CINIC-10, 100 epochs on CIFAR-100, and 30 epochs on Purchase100 Texas100. Data augmentation techniques such as random cropping and horizontal flipping are applied to reduce over-fitting.
\begin{table}[]
\centering
\begin{tabular}{l|ccccc}
\toprule
Dataset& $\targetDataset^{train}$&  $\targetDataset^{test}$& $\auxDataset$\\
\midrule
CIFAR-10&  15,000& 15,000& 30,000\\
CIFAR-100& 15,000& 15,000& 30,000\\
CINIC-10&  15,000& 15,000& 30,000\\
Purchase-100&  15,000& 15,000& 30,000\\
Texas-100&  15,000& 15,000& 30,000\\
\bottomrule
\end{tabular}
\vspace{0.5cm}
\caption{\textbf{Dataset Partitioning Sizes}. Each subset is disjoint from the others after partitioning.}
\label{table:dataset_split}
\end{table}

\subsection{Setup for MIAs.} 
\label{sec:appe_setup-for-mias}

We implemented seven MIAs using the setup described below, which we adopt as the \textbf{standard setting}. Unless explicitly stated otherwise, this setup is used consistently across all experiments presented in the paper.

\begin{itemize}[leftmargin=*, noitemsep,topsep=0pt,parsep=0pt,partopsep=0pt]
    \item \textbf{LOSS \cite{yeom}.} This attack queries the target model $\targetModel$ with all samples from $\auxDataset$, and then we use the average loss as the global threshold to make membership predictions.

    \item \textbf{Augmentation Attack \cite{choquette2021label}.} We implement the Augmentation attack with 5 rotations as the augmentation technique. It uses a 2-layer MLP as the attack model.

    \item \textbf{Loss Trajectory Attack \cite{losstraj}.} We distillate both the target model and the shadow model for 100 epochs. We perform model distillation on a distillation set $\auxDataset^{d}$ that's half of the auxiliary dataset $\auxDataset$. For the rest of the dataset $\auxDataset^{s}$ that's disjoint from $\auxDataset^{d}$, we partition it again into halves as the train (members) and test (non-members)  set for the shadow model.

    \item \textbf{LiRA \cite{lira}.} Our implementation trains 20 shadow models. Each shadow model is trained using half of the auxiliary dataset. We follow the same procedure as \citet{lira} to split the auxiliary dataset $\auxDataset$ such that for each datapoint in $\auxDataset$, it's partitioned into a shadow training set 10 times and a shadow testing set 10 times. In LiRA's paper, they use 256 models for most experiments. In Appendix B.B, they introduce the estimation of global variance to improve the performance of LiRA with a smaller number of models, and demonstrate that 16 shadow models can perform on par with their best attack. We have also justified that the inconsistency of MIA is invariant to the number of shadow models used in Section~\ref{sec:inconsistency-of-membership-inference-attacks}. The version of LiRA considered in the paper is the "online" version. For a discussion of the "offline" version of LiRA, see Appendix Section~\ref{sec:lira-online-offline-comp-appendix}.

    \item \textbf{Class-NN \cite{shokri}.} This attack uses 10 shadow models and 3-layer MLP attack models. The number of attack models is equal to the number of labels, making it class-specific.

    \item \textbf{Difficulty Calibration Loss Attack \cite{diff_calibration}.} This attack trains a single shadow model \( f_s \) on \( \auxDataset \). The threshold \(\tau\) for making membership predictions is determined by sampling 1,000 thresholds and selecting the one that maximizes accuracy on \( \auxDataset \).

    \item \textbf{Reference Attack \cite{ye}.} To save computational resources, Reference Attack shares the 20 shadow models with LiRA. Reference Attack's source code\footnote{\url{https://github.com/privacytrustlab/ml_privacy_meter/tree/295e7e37e889e12df4083b812f71ed2e2ddd8b4a/research/2022_enhanced_mia}} re-used the source code\footnote{\url{https://github.com/tensorflow/privacy/tree/master/research/mi_lira_2021}} for LiRA. Therefore, we reuse the same shadow model set trained for LiRA for the reference attack.
\end{itemize}

\begin{figure}[h]
    \centering
    \begin{subfigure}[b]{0.2\textwidth}
        \includegraphics[width=\textwidth]{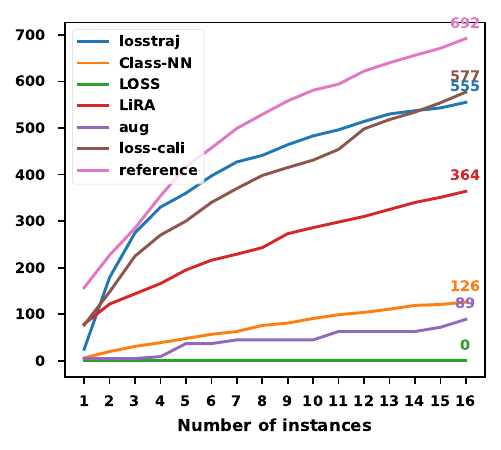}
        \caption{FPR = 0.001}
        \label{fig:coverage_over_random_diff_fpr_0.001}
    \end{subfigure}
    \hfill
    \begin{subfigure}[b]{0.2\textwidth}
        \includegraphics[width=\textwidth]{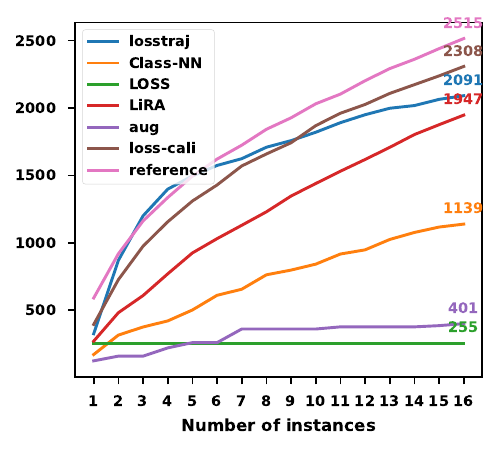}
        \caption{FPR = 0.01}
        \label{fig:coverage_over_random_diff_fpr_0.01}
    \end{subfigure}
    
    \vspace{1em}
    
    \begin{subfigure}[b]{0.2\textwidth}
        \includegraphics[width=\textwidth]{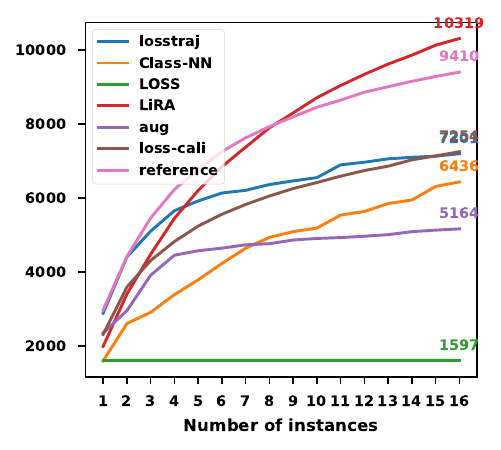}
        \caption{FPR = 0.1}
        \label{fig:coverage_over_random_diff_fpr_0.1}
    \end{subfigure}
    \hfill
    \begin{subfigure}[b]{0.2\textwidth}
        \includegraphics[width=\textwidth]{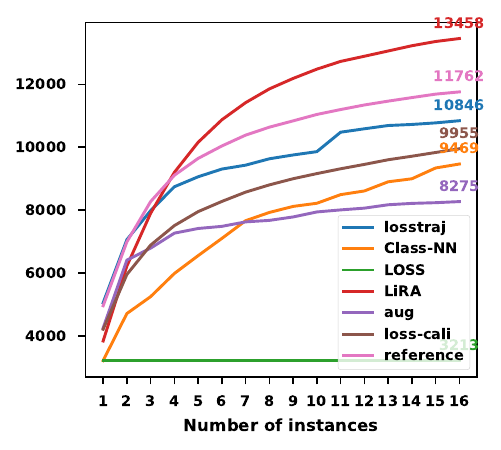}
        \caption{FPR = 0.2}
        \label{fig:coverage_over_random_diff_fpr_0.2}
    \end{subfigure}

    \vspace{1em}
    
    \caption{Trend of Coverage with Varying Numbers of Instances for Each MIA. Each subplot represents the coverage of instances with different FPRs.).}
    \label{fig:coverage-tp-diff-fpr}
\end{figure}

\begin{figure}[h]
    \centering
    \begin{subfigure}[b]{0.2\textwidth}
        \centering
        \includegraphics[width=\textwidth]{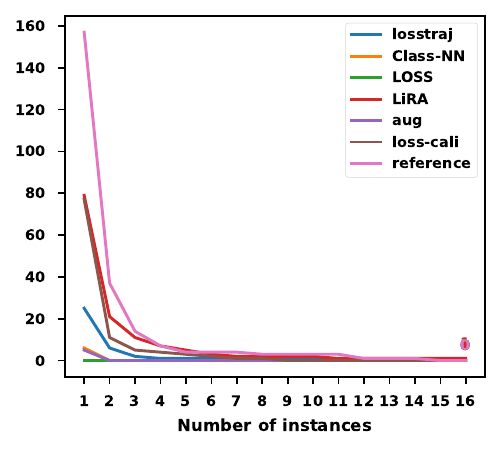}
        \caption{FPR = 0.001}
        \label{fig:stability_over_random_diff_fpr_0.001}
    \end{subfigure}
    \hfill
    \begin{subfigure}[b]{0.2\textwidth}
        \centering
        \includegraphics[width=\textwidth]{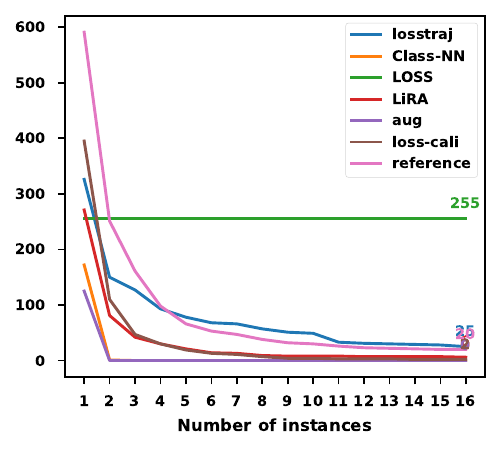}
        \caption{FPR = 0.01}
        \label{fig:stability_over_random_diff_fpr_0.01}
    \end{subfigure}
    
    \vspace{1em}
    
    \begin{subfigure}[b]{0.2\textwidth}
        \centering
        \includegraphics[width=\textwidth]{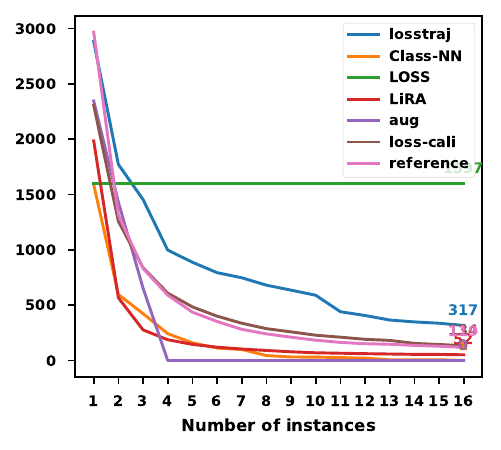}
        \caption{FPR = 0.1}
        \label{fig:stability_over_random_diff_fpr_0.1}
    \end{subfigure}
    \hfill
    \begin{subfigure}[b]{0.2\textwidth}
        \centering
        \includegraphics[width=\textwidth]{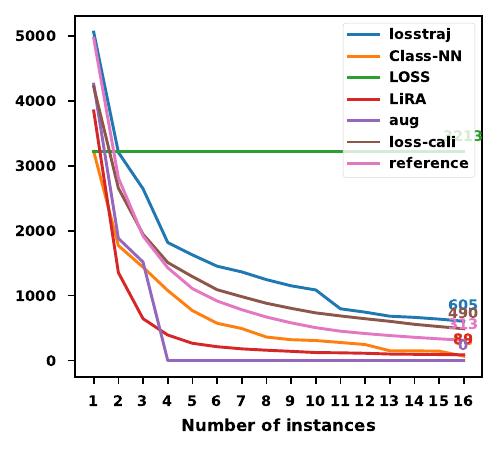}
        \caption{FPR = 0.2}
        \label{fig:stability_over_random_diff_fpr_0.2}
    \end{subfigure}
    
    \vspace{1em}
    
    \caption{\textbf{Trend of Stability.} Each line shows the stability of the respective attack across varying numbers of instances, following the setup in Figure \ref{fig:coverage-tp-diff-fpr}.}
    \label{fig:stability-tp-diff-fpr}
\end{figure}


\section{MIA Method Disparity at Different FPR}
\label{sec:mia-method-disparity-at-different-fpr}

In Appendix Figure~\ref{fig:trend-of-agreement-vs-fpr}, the trend of Jaccard Similarity shows a positive correlation with FPR, indicating that disparity between attacks is highest at low FPRs and decreases as FPR increases. To further analyze this trend, we present the similarity between pairs of attacks across various FPR levels in Appendix Figure~\ref{fig:jaccard-heat-map-appendix}.

The similarity between attacks, measured in terms of both coverage and stability, remains quite low overall. This is particularly evident for stability at FPR = 0.001 (Appendix Figure~\ref{fig:similarity-matrix_stability_fpr_0.001}), where half of the attack pairs exhibit similarity values below 0.004, and the highest observed similarity is just 0.08. As FPR increases, the patterns of high similarity between certain pairs of attacks persist. For example, LiRA and the Reference Attack, LiRA and the Calibration Loss Attack, and the Calibration Loss Attack and the Loss Trajectory Attack consistently show relatively high correlation across different FPR levels. Conversely, pairs with high disparity, such as the LOSS Attack and the Calibration Loss Attack or the LOSS Attack and the Loss Trajectory Attack, consistently demonstrate low similarity across all analyzed FPR levels (e.g., Appendix Figure~\ref{fig:similarity-matrix_coverage_fpr_0.2}).

We conclude that the inherent characteristics of each attack drive their disparity, with some attacks naturally aligning due to shared methodologies or signals, while others diverge significantly. This underscores the importance of leveraging a diverse set of attacks in privacy auditing to capture a comprehensive view of membership vulnerabilities.

\begin{figure*}[htbp]
    \centering
    \begin{subfigure}[b]{0.23\textwidth}
        \centering
        \includegraphics[width=\textwidth]{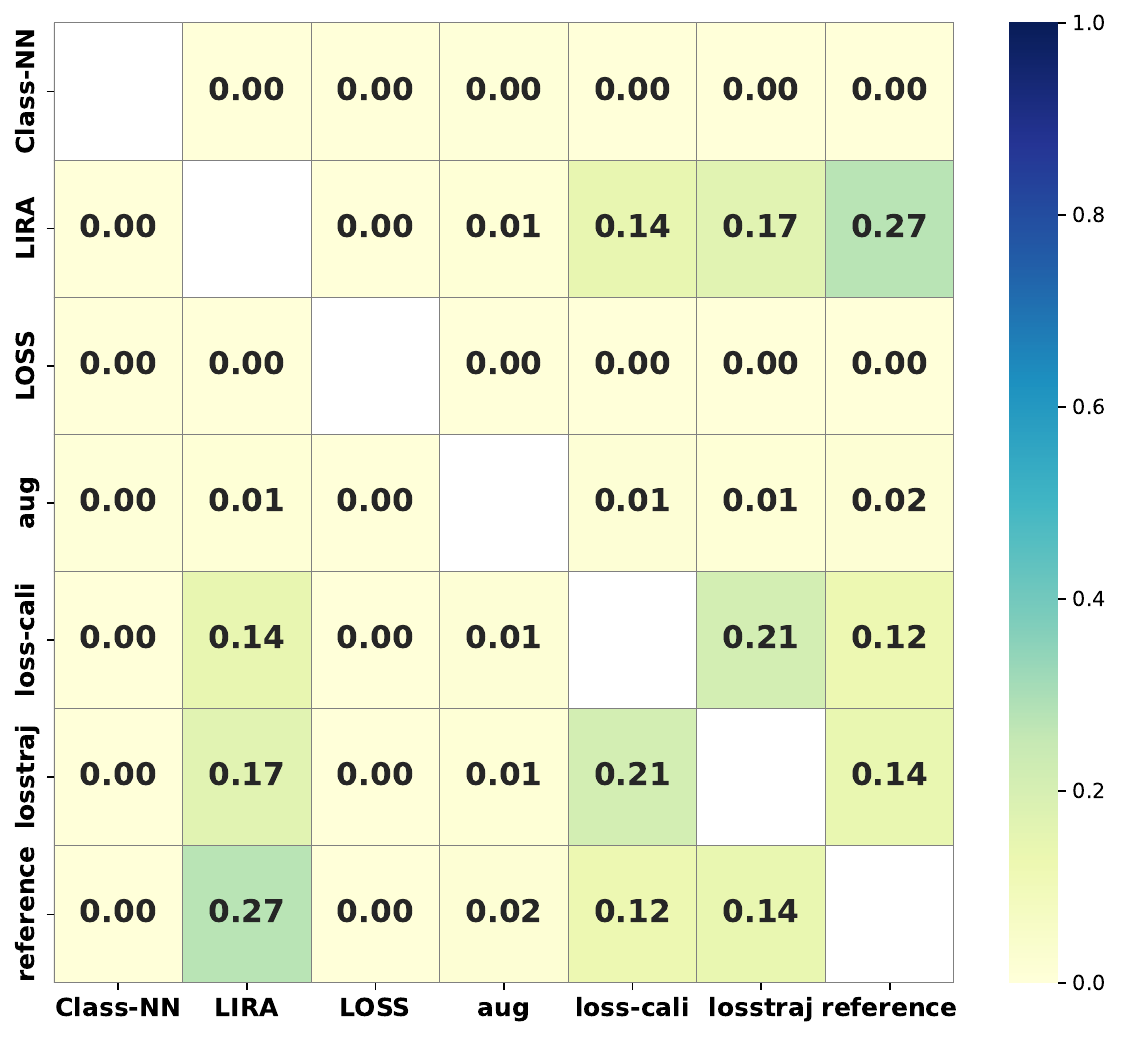}
        \caption{Coverage (FPR = 0.001)}
        \label{fig:similarity-matrix_coverage_fpr_0.001}
    \end{subfigure}
    \hfill
    \begin{subfigure}[b]{0.23\textwidth}
        \centering
        \includegraphics[width=\textwidth]{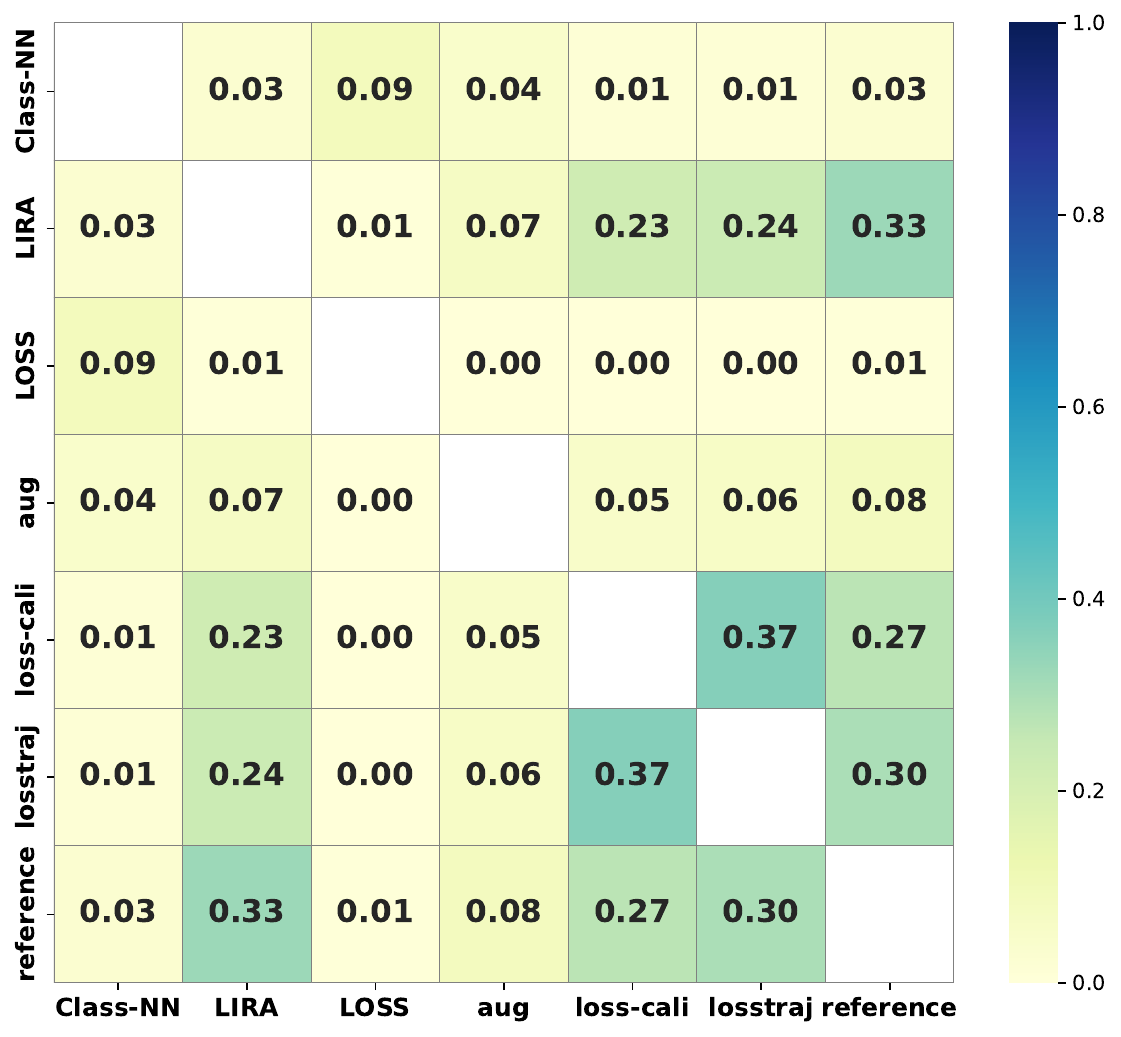}
        \caption{Coverage (FPR = 0.01)}
        \label{fig:similarity-matrix_coverage_fpr_0.01}
    \end{subfigure}
    \hfill
    \begin{subfigure}[b]{0.23\textwidth}
        \centering
        \includegraphics[width=\textwidth]{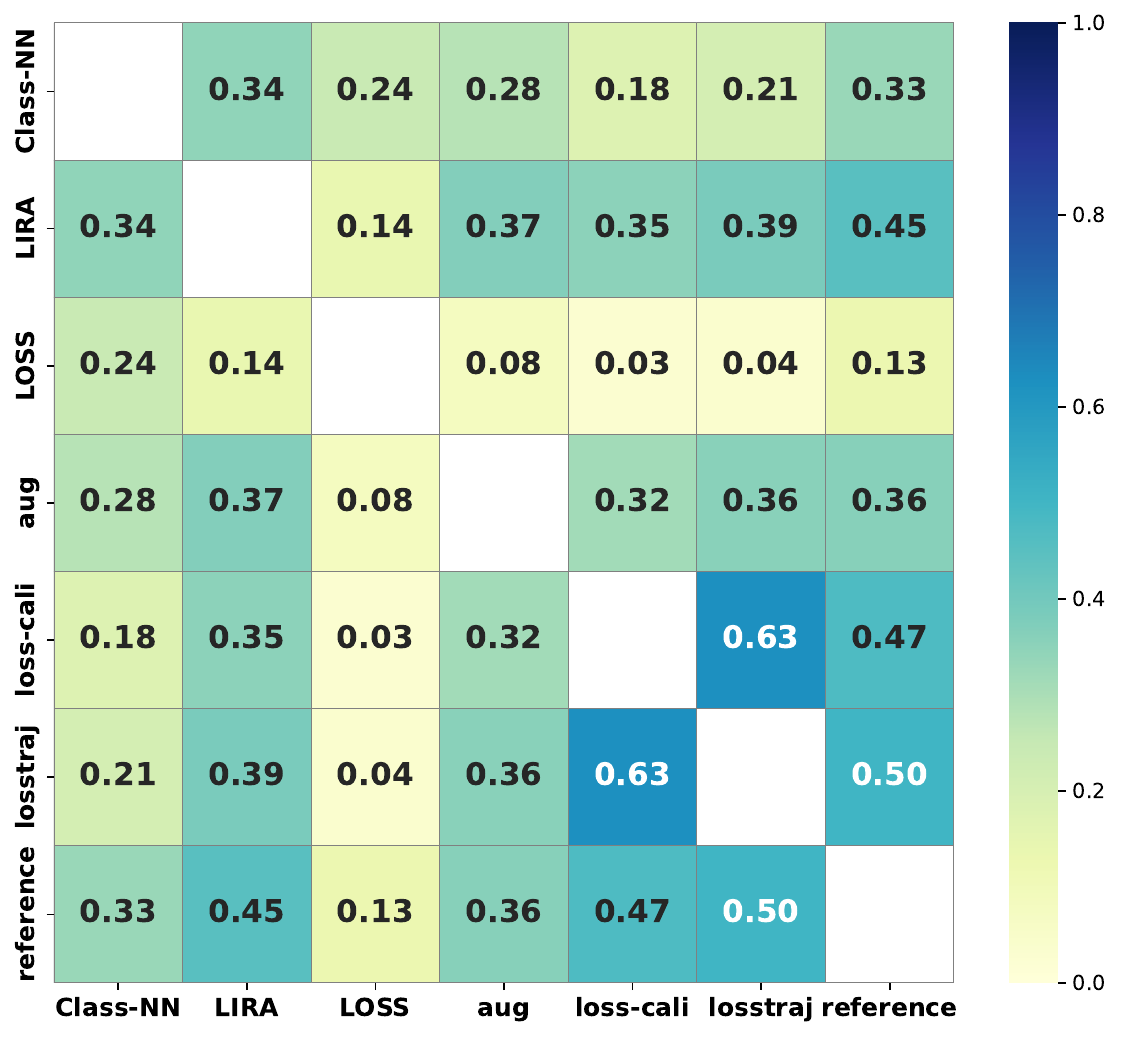}
        \caption{Coverage (FPR = 0.1)}
        \label{fig:similarity-matrix_coverage_fpr_0.1}
    \end{subfigure}
    \hfill
    \begin{subfigure}[b]{0.23\textwidth}
        \centering
        \includegraphics[width=\textwidth]{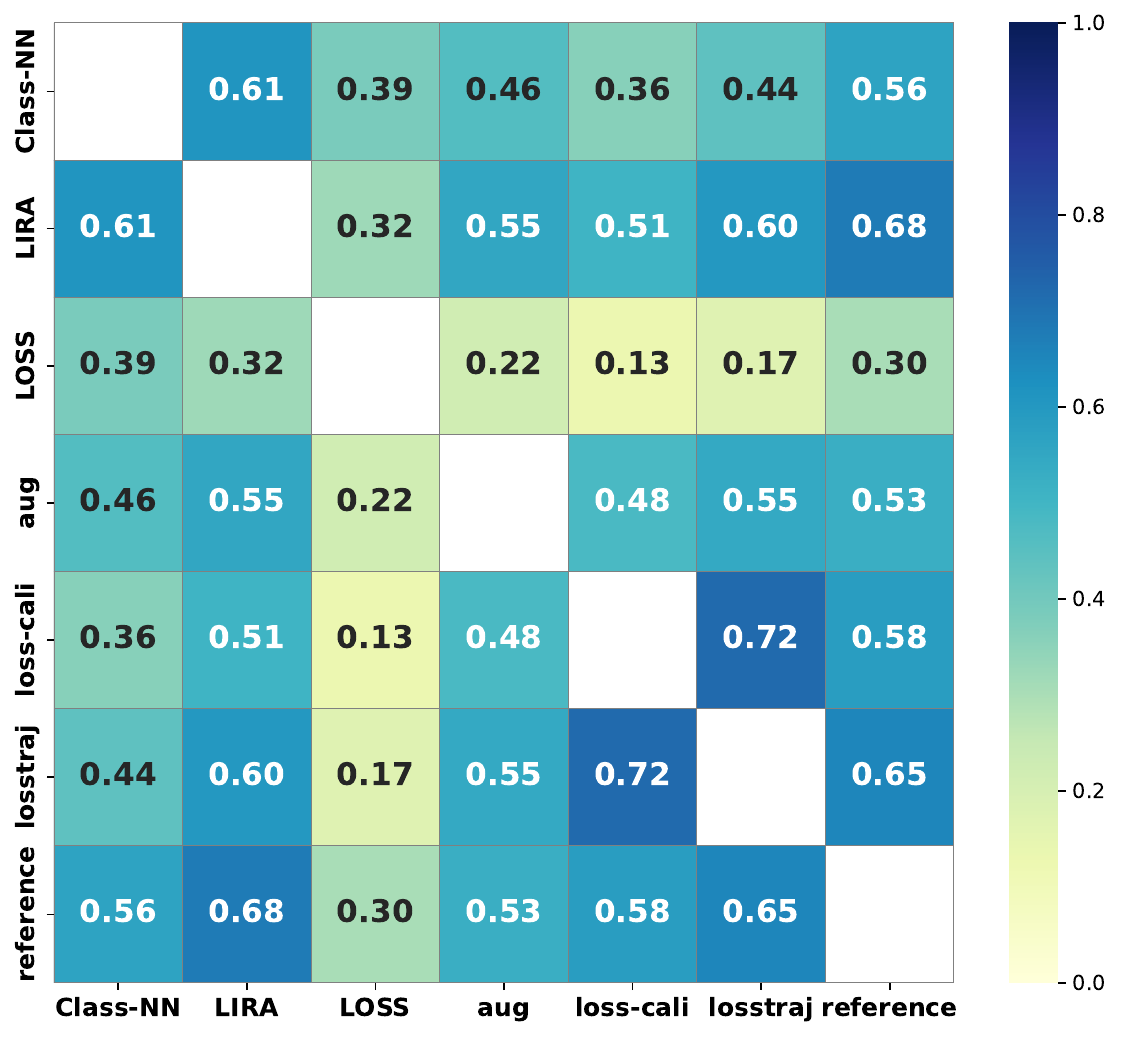}
        \caption{Coverage (FPR = 0.2)}
        \label{fig:similarity-matrix_coverage_fpr_0.2}
    \end{subfigure}
    
    \vskip\baselineskip  
    
    \begin{subfigure}[b]{0.23\textwidth}
        \centering
        \includegraphics[width=\textwidth]{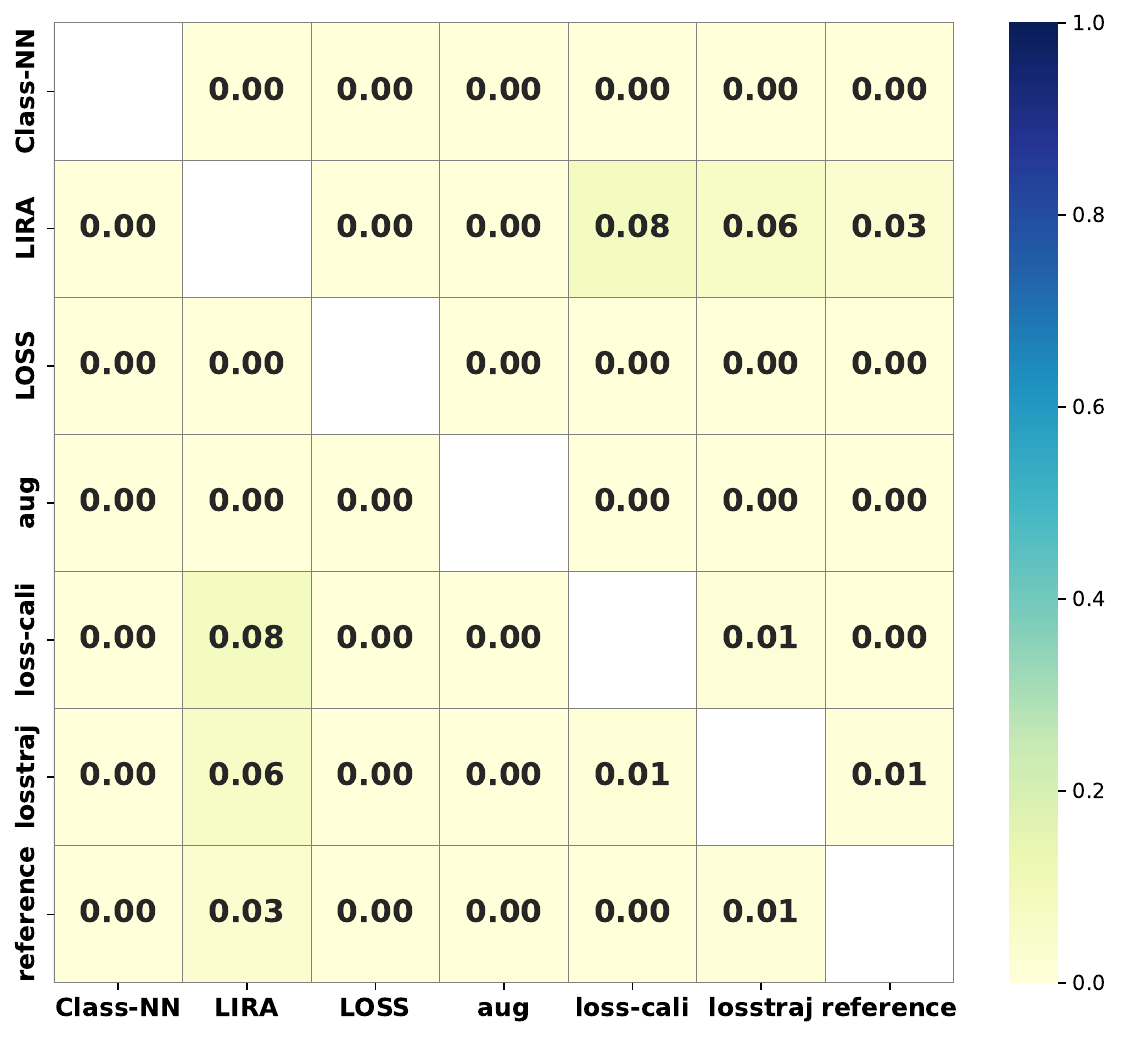}
        \caption{Stability (FPR = 0.001)}
        \label{fig:similarity-matrix_stability_fpr_0.001}
    \end{subfigure}
    \hfill
    \begin{subfigure}[b]{0.23\textwidth}
        \centering
        \includegraphics[width=\textwidth]{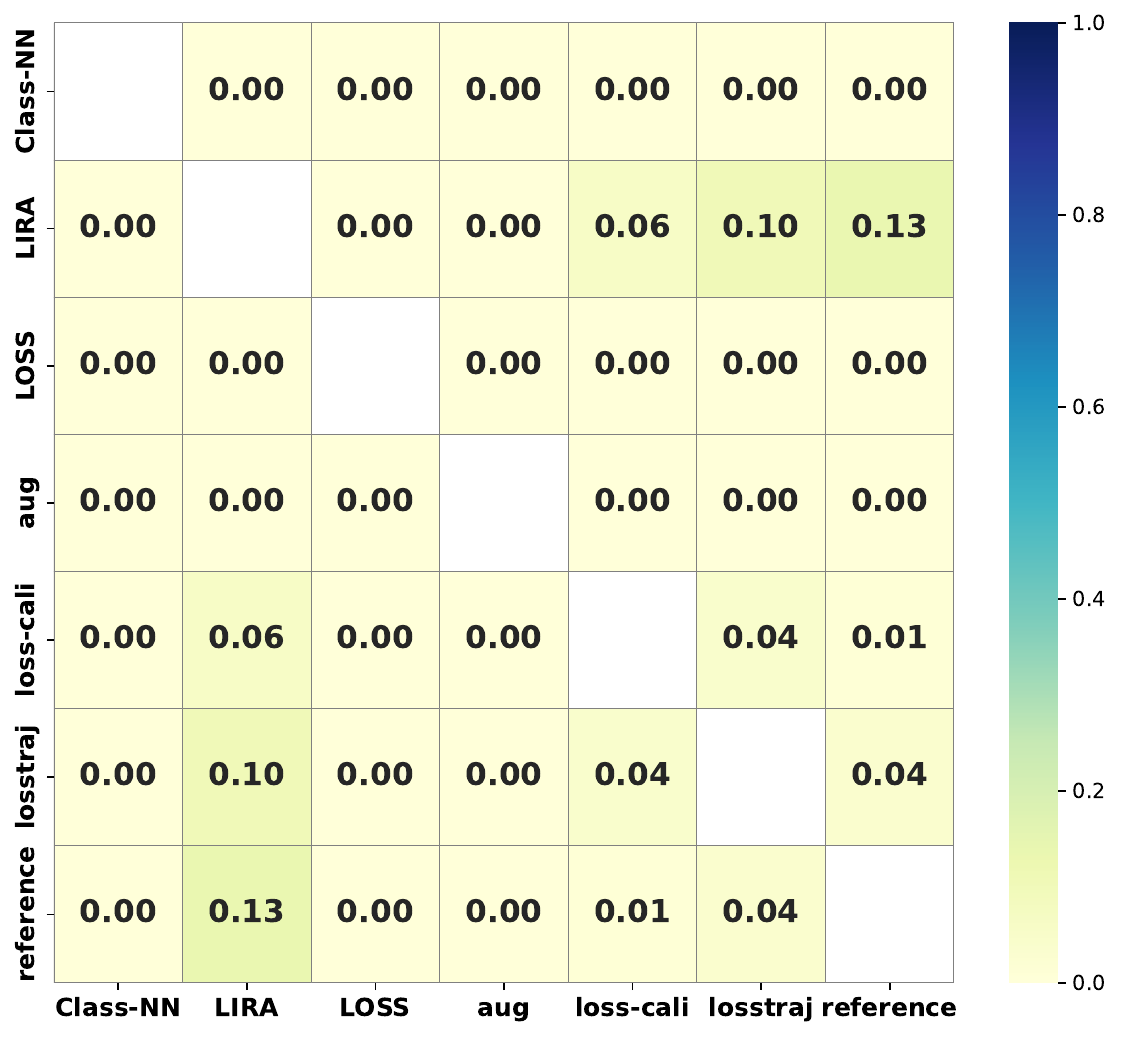}
        \caption{Stability (FPR = 0.01)}
        \label{fig:similarity-matrix_stability_fpr_0.01}
    \end{subfigure}
    \hfill
    \begin{subfigure}[b]{0.23\textwidth}
        \centering
        \includegraphics[width=\textwidth]{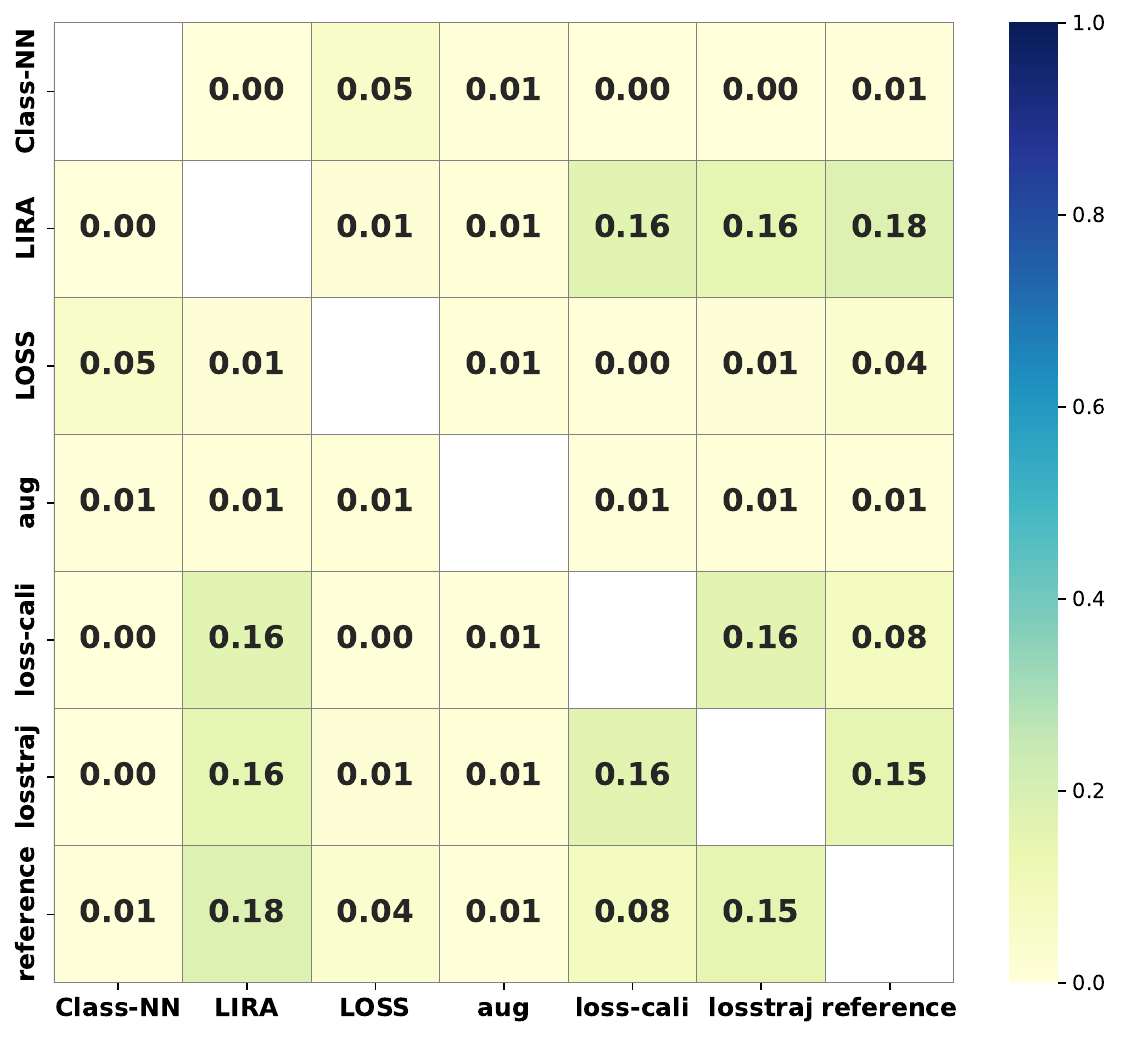}
        \caption{Stability (FPR = 0.1)}
        \label{fig:similarity-matrix_stability_fpr_0.1}
    \end{subfigure}
    \hfill
    \begin{subfigure}[b]{0.23\textwidth}
        \centering
        \includegraphics[width=\textwidth]{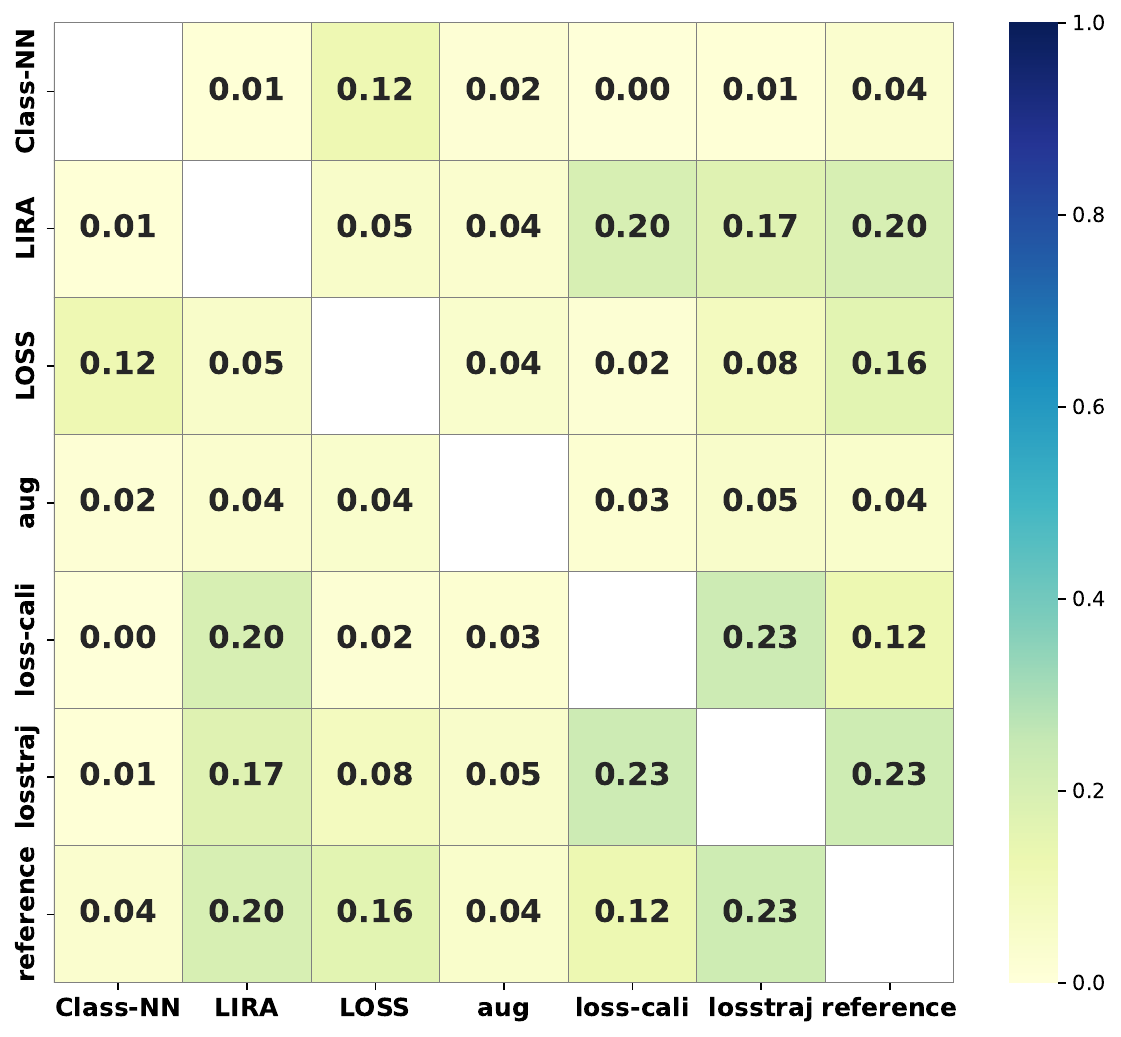}
        \caption{Stability (FPR = 0.2)}
        \label{fig:similarity-matrix_stability_fpr_0.2}
    \end{subfigure}

    \vspace{0.5cm}
    
    \caption{\textbf{Disparity of Membership Inference Attacks Across Different FPRs.}
    The values represent the average Jaccard similarity across 4 experimental runs.
    Coverage and stability for each attack are calculated with 6 instances at varying FPR values 
    (\(0.001\), \(0.01\), \(0.1\), \(0.2\)), illustrating how similarity changes across thresholds.}
    \label{fig:jaccard-heat-map-appendix}
    \vspace{0.5cm}
\end{figure*}

\section{Ensemble}
\label{sec:app-ensemble}

\subsection{Cost Analysis}
\label{sec:app-cost-analysis}

As detailed in the main body of the paper, we model the computational cost of each attack in terms of GPU minutes. Our experiments were conducted on Nvidia L40S GPUs, with both target and shadow models being ResNet-56. Below, we outline the parameters for each attack that influence compute time:

\begin{itemize}[leftmargin=*, noitemsep,topsep=0pt,parsep=0pt,partopsep=0pt]
    \item \textbf{LiRA (580 minutes)}: Shadow model training is performed with a batch size of 512, while queries to the target and shadow models use a batch size of 256. Each query includes 18 augmentations. A total of 20 shadow models are prepared for each instance.

    \item \textbf{Loss Trajectory Attack (17 minutes)}: Includes shadow model training and distillation for both the target and shadow models, with a batch size of 512.

    \item \textbf{Calibration Loss Attack (5 minutes)}: Shadow model training is conducted with a batch size of 512, and losses are queried sequentially with a batch size of 1.

    \item \textbf{Reference Attack (540 minutes)}: Shadow model training uses a batch size of 512, and queries to the target and shadow models use a batch size of 256. Similar to LiRA, 20 shadow models are prepared for each instance.
\end{itemize}

As discussed in Section~\ref{sec:optimization-strategies-for-ensemble}, the Reference Attack and LiRA share the same shadow models. To avoid double-counting in our cost analysis, the training cost for shadow models (540 minutes) is deducted when both LiRA and the Reference Attack are included in an ensemble combination.

We extend our analysis of the cost-performance relationship (Section~\ref{sec:cost-analysis}) in two key dimensions:

\subsubsection{Attack Combination} 
In Appendix Figure~\ref{fig:cost_perf_analysis_appendix}, each performance sample is represented by a distinct marker shape, corresponding to the attack combinations detailed in the legend. When comparing ensembles with the same number of instances (indicated by the same color), the ensemble combining all four attacks consistently dominates other combinations. This finding highlights that all three ensemble strategies can effectively utilize the computational cost associated with employing diverse attacks.

\subsubsection{Number of Instances} 
The general trend indicates that the best-performing ensemble does not always correspond to the one using the most instances. Notably, for the Coverage ensemble, the relationship between cost and performance is sometimes inverse. This occurs because the multi-instance coverage step (Equation~\ref{eq:multi-instances-coverage}) trades an increase in the false positive rate for a higher true positive rate. As a result, the ensemble’s TPR at low FPR is reduced, impacting performance. For the other two ensembles, more instances are usually associated with better performance.

From our analysis, we know our ensemble would always benefit from utilizing more attacks (in the multi-attack step), but not always from more instances (in the multi-instance step). We encourage researchers who wish to adapt our ensemble to prepare attacks with multi-instance to their compute budget, and then perform MIA on a held-out set to find their desired ensemble performance.

\begin{figure*}[]
    \centering
    \begin{minipage}{0.8\textwidth} 
        \centering
        \begin{subfigure}[t]{0.32\textwidth}
            \includegraphics[width=\textwidth]{images/appendix/appendix_cost_perf_analysis/cifar10_intersection_cost_vs_perf.pdf}
            \caption{Stability Ensemble (CIFAR-10)}
        \end{subfigure}
        \hfill
        \begin{subfigure}[t]{0.32\textwidth}
            \includegraphics[width=\textwidth]{images/appendix/appendix_cost_perf_analysis/cifar10_union_cost_vs_perf.pdf}
            \caption{Coverage Ensemble (CIFAR-10)}
        \end{subfigure}
        \hfill
        \begin{subfigure}[t]{0.32\textwidth}
            \includegraphics[width=\textwidth]{images/appendix/appendix_cost_perf_analysis/cifar10_majority_vote_cost_vs_perf.pdf}
            \caption{Majority Vote Ensemble (CIFAR-10)}
        \end{subfigure}

        \vspace{0.3cm}

        \begin{subfigure}[t]{0.32\textwidth}
            \includegraphics[width=\textwidth]{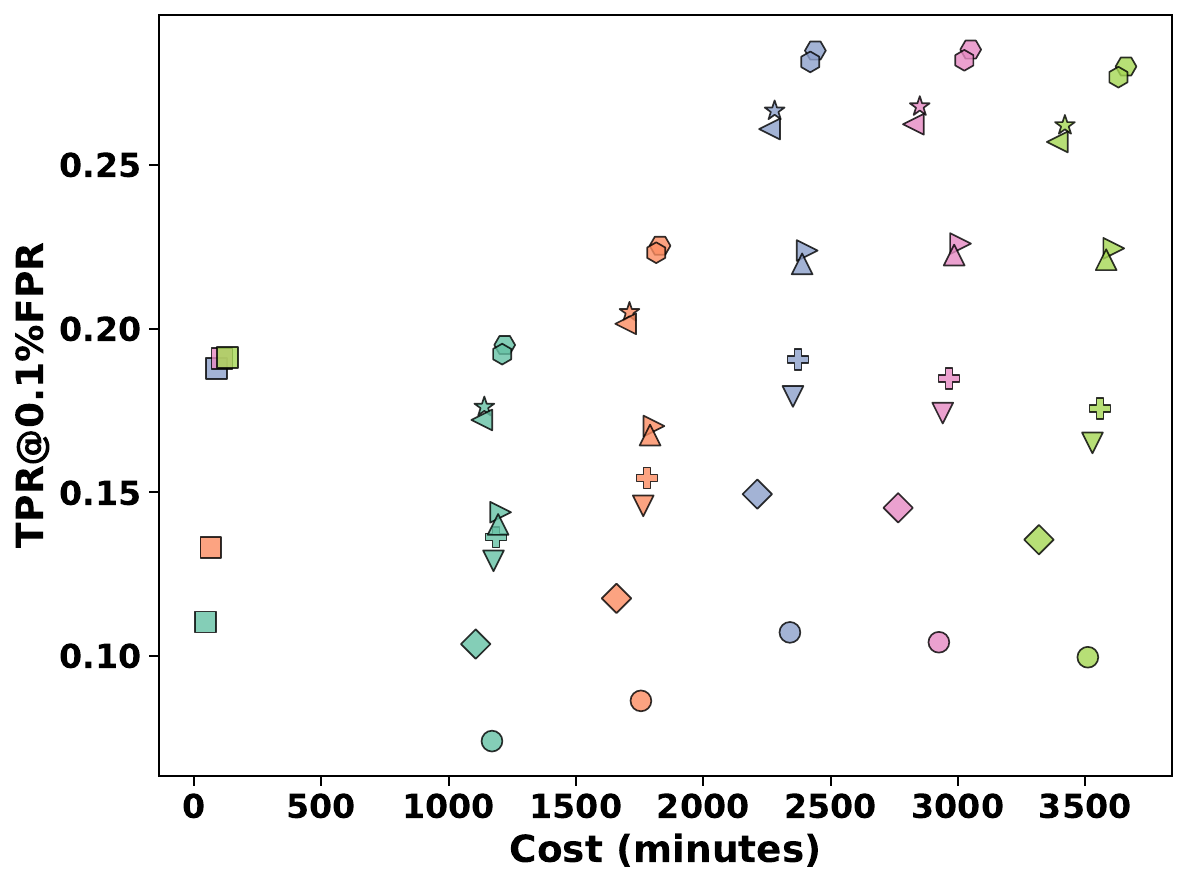}
            \caption{Stability Ensemble (CIFAR-100)}
        \end{subfigure}
        \hfill
        \begin{subfigure}[t]{0.32\textwidth}
            \includegraphics[width=\textwidth]{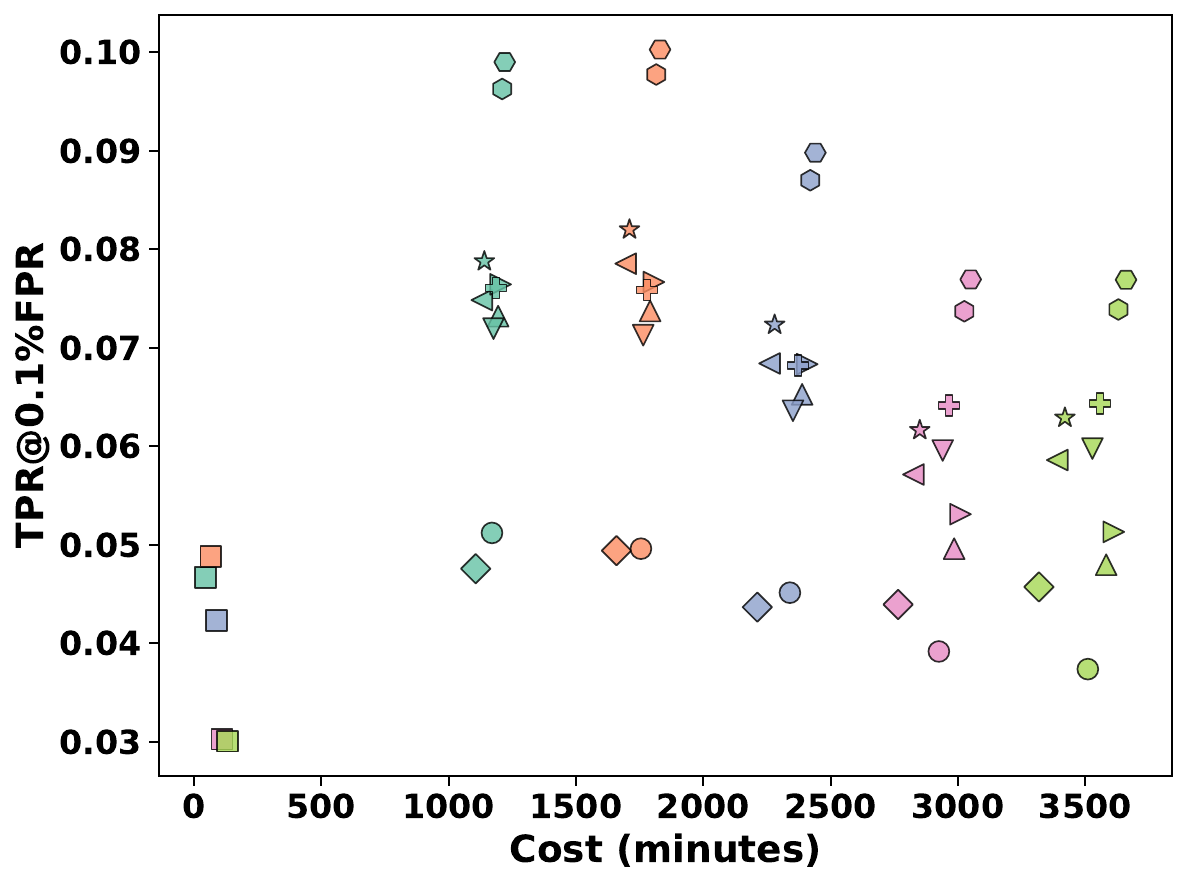}
            \caption{Coverage Ensemble (CIFAR-100)}
        \end{subfigure}
        \hfill
        \begin{subfigure}[t]{0.32\textwidth}
            \includegraphics[width=\textwidth]{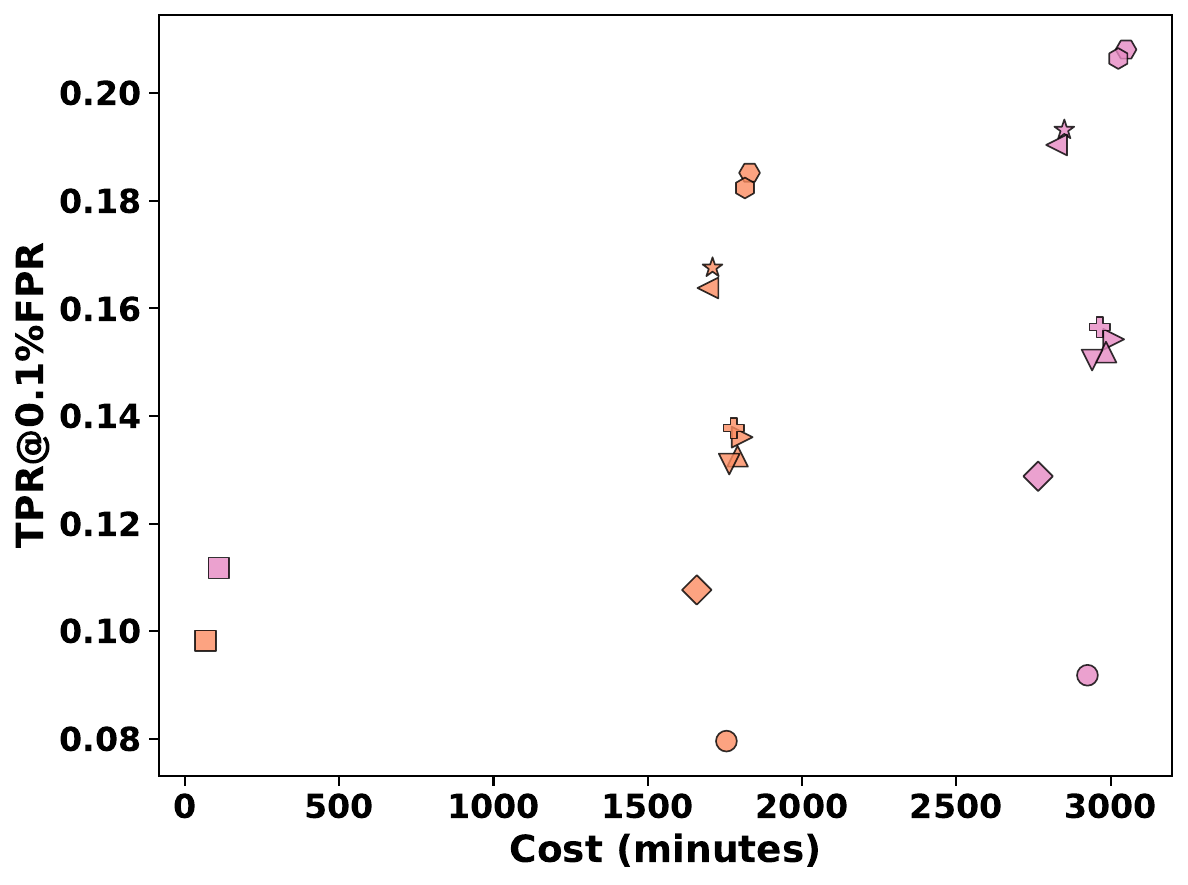}
            \caption{Majority Vote Ensemble (CIFAR-100)}
        \end{subfigure}

        \vspace{0.3cm}

        \begin{subfigure}[t]{0.32\textwidth}
            \includegraphics[width=\textwidth]{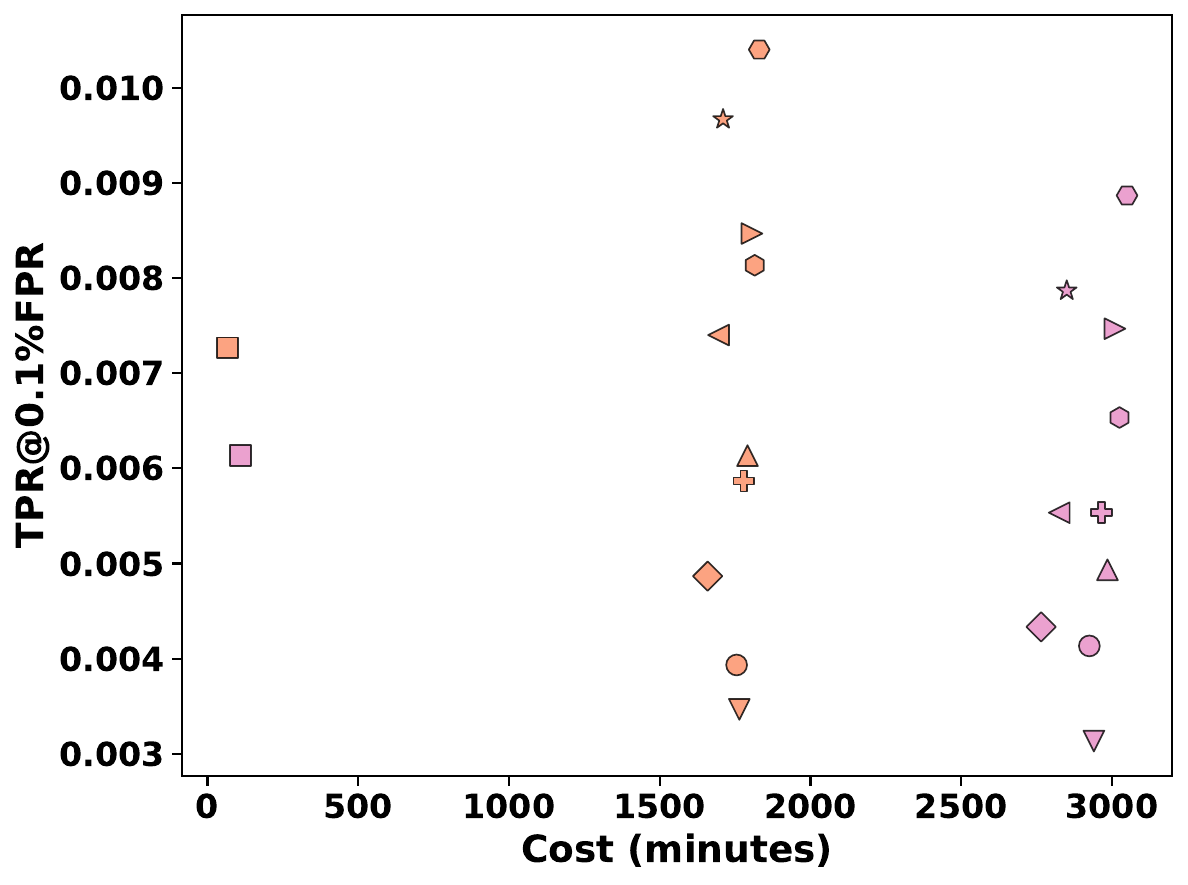}
            \caption{Stability Ensemble (CINIC-10)}
        \end{subfigure}
        \hfill
        \begin{subfigure}[t]{0.32\textwidth}
            \includegraphics[width=\textwidth]{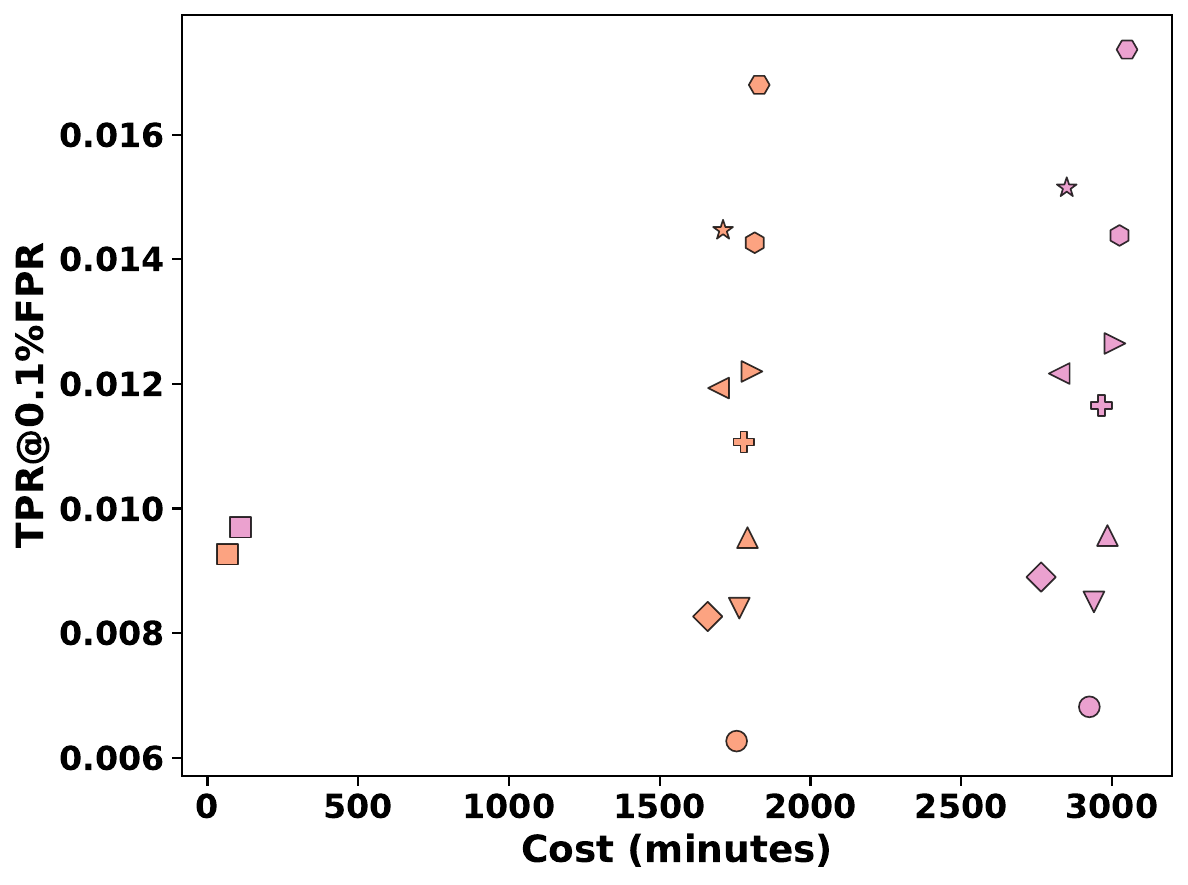}
            \caption{Coverage Ensemble (CINIC-10)}
        \end{subfigure}
        \hfill
        \begin{subfigure}[t]{0.32\textwidth}
            \includegraphics[width=\textwidth]{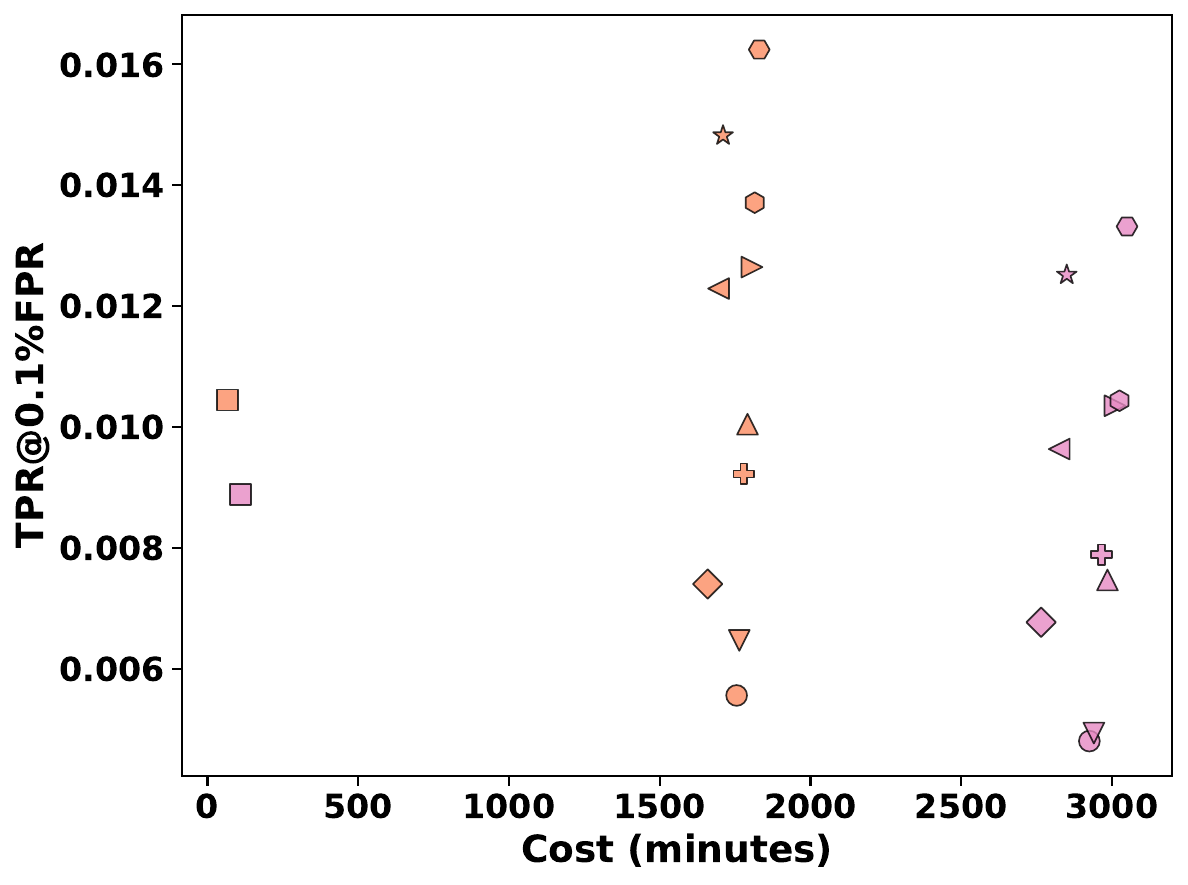}
            \caption{Majority Vote Ensemble (CINIC-10)}
        \end{subfigure}
    \end{minipage}
    \hfill
    \begin{minipage}{0.15\textwidth} 
        \centering
        \begin{figure}[H]
            \centering
            \includegraphics[width=\textwidth]{images/appendix/appendix_cost_perf_analysis/marker_legend.pdf}
        \end{figure}
        \begin{figure}[H]
            \flushleft
            \includegraphics[width=0.5\textwidth]{images/appendix/appendix_cost_perf_analysis/color_legend.pdf}
        \end{figure}
    \end{minipage}

    \vspace{0.3cm}

    \caption{Comparison of ensemble strategies (Stability Ensemble, Majority Vote Ensemble, Coverage Ensemble) across different datasets. Each instance makes membership inferences at FPR=0.01. The target model is ResNet-56. Each row corresponds to a dataset, and each column represents an ensemble strategy. The marker legend explains the markers used for different attack combinations, and the color legend represents the number of instances. For the Majority Vote, we have skipped even numbers of instances since the majority vote with an even number could lead to a tie.}
    \label{fig:cost_perf_analysis_appendix}
    \vspace{0.5cm}
\end{figure*}

\begin{figure*}[]
    \centering
    \begin{minipage}{0.75\textwidth} 
        \centering
        \begin{subfigure}[t]{0.32\textwidth}
            \includegraphics[width=\textwidth]{images/appendix/appendix_ensemble_roc/cifar10_intersection_roc_plot_4attacks.pdf}
            \caption{Stability Ensemble (4 Attacks)}
        \end{subfigure}
        \hfill
        \begin{subfigure}[t]{0.32\textwidth}
            \includegraphics[width=\textwidth]{images/appendix/appendix_ensemble_roc/cifar10_union_roc_plot_4attacks.pdf}
            \caption{Coverage Ensemble (4 Attacks)}
        \end{subfigure}
        \hfill
        \begin{subfigure}[t]{0.32\textwidth}
            \includegraphics[width=\textwidth]{images/appendix/appendix_ensemble_roc/cifar10_majority_vote_roc_plot_4attacks.pdf}
            \caption{Majority Vote Ensemble (4 Attacks)}
        \end{subfigure}

        \vspace{0.3cm}

        \begin{subfigure}[t]{0.32\textwidth}
            \includegraphics[width=\textwidth]{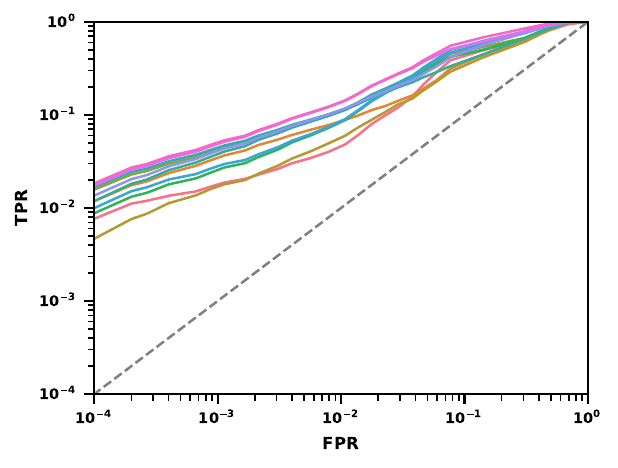}
            \caption{Stability Ensemble (All Combinations)}
        \end{subfigure}
        \hfill
        \begin{subfigure}[t]{0.32\textwidth}
            \includegraphics[width=\textwidth]{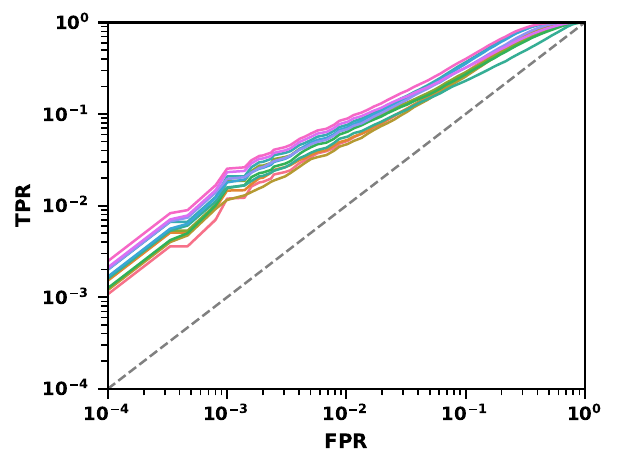}
            \caption{Coverage Ensemble (All Combinations)}
        \end{subfigure}
        \hfill
        \begin{subfigure}[t]{0.32\textwidth}
            \includegraphics[width=\textwidth]{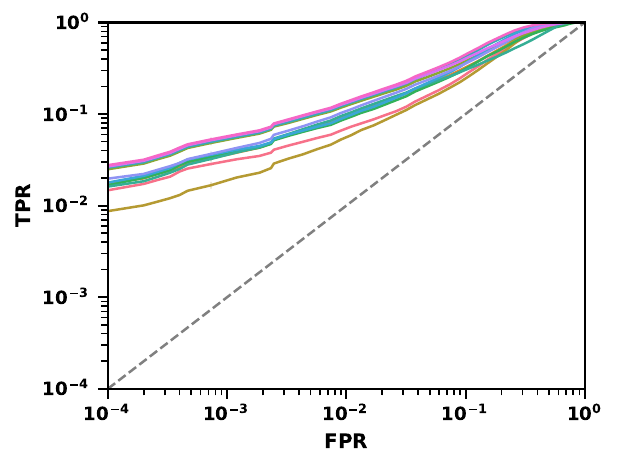}
            \caption{Majority Vote Ensemble (All Combinations)}
        \end{subfigure}
    \end{minipage}
    \hfill
    \begin{minipage}{0.2\textwidth} 
        \centering
        \begin{figure}[H]
            \flushleft
            \includegraphics[width=0.7\textwidth]{images/appendix/appendix_ensemble_roc/roc_4attacks_legend.pdf}
        \end{figure}
        \begin{figure}[H]
            \flushleft
            \includegraphics[width=1\textwidth]{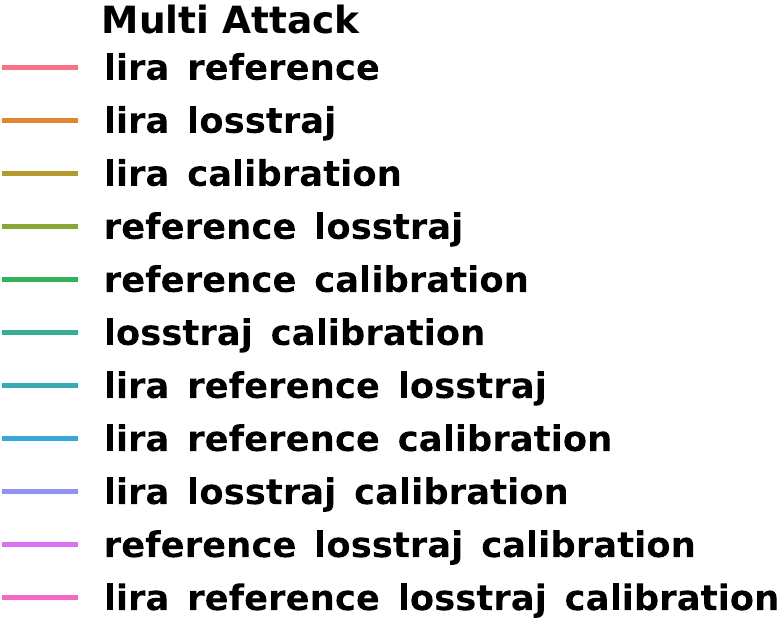}
        \end{figure}
    \end{minipage}

    \vspace{0.3cm}

    \caption{Comparison of ensemble ROC curves for different attack combinations on the CIFAR-10 dataset.  Each row compares different ensemble strategies (Stability Ensemble, Coverage Ensemble, Majority Vote Ensemble) for a given set of attacks. The row on the bottom shows different combinations of attacks used in the ensemble. All Ensemble are performed with 6 instances.}
    \label{fig:ensemble_roc_appendix}
    \vspace{0.5cm}
\end{figure*}

\begin{table*}[]
    \centering
    \small 
    \begin{tabular}{lcccc}
        \toprule
        \textbf{Attack} & \textbf{Shadow Models} & \textbf{Attack Models} & \makecell{\textbf{Other Factors} \\ \textbf{Involving Randomness}} & \textbf{Inference Classifier} \\
        \midrule
        Augmentation attack \cite{choquette2021label} & 1 & 1 & Data augmentation & Attack model \\
        Loss trajectory attack \cite{losstraj} & 1 & 1 & Model distillation & Attack model \\
        LiRA \cite{lira} & 20 & 0 & None & Likelihood-ratio test \\
        Reference attack \cite{ye} & 20 & 0 & None & Hypothesis test \\
        Class-NN \cite{shokri} & 10 & = \# of classes & None & Attack model \\
        LOSS \cite{yeom} & 0 & 0 & None & Global threshold \\
        Difficulty calibration loss attack \cite{diff_calibration} & 1 & 0 & None & Global threshold \\
        \bottomrule
    \end{tabular}
    \vspace{0.5cm}
   \caption{\textbf{A Summary of Membership Inference Attacks.} The set-up follows the \textbf{standard setting} in Appendix Section~\ref{sec:appe_setup-for-mias}.}
   \label{tab:attack_complexity}
\end{table*}

\subsection{Ensemble on Other Model Architectures}
\label{sec:ensemble-on-other-model-architectures-appendix}

Our ensemble methods demonstrate consistent improvements across three additional model architectures: WideResNet-32, MobileNetV2, and VGG-16, as shown in Table~\ref{tab:ensemble-perf-table-all-other-model-appendix}. This table extends Table~\ref{tab:ensemble-perf-table} by evaluating AUC, Accuracy, and TPR@0.1\% FPR metrics across CIFAR-10, CIFAR-100, and CINIC-10 datasets.

The Stability Ensemble and Majority Voting Ensemble consistently outperform single-instance attacks across all metrics and datasets. The Majority Voting Ensemble achieves performance comparable to the Stability Ensemble, particularly at low FPR, while maintaining competitive TPR at higher FPR levels. This highlights its balanced ability to draw strengths from both Stability and Coverage ensembles.

In contrast, the Coverage Ensemble demonstrates relatively lower performance on most metrics, particularly at 0.1\% FPR, where its TPR does not consistently surpass that of single-instance attacks. This behavior reflects its inherent trade-off, prioritizing broader member coverage at the expense of precision.

Overall, the Stability Ensemble delivers the highest precision and robustness across all architectures, making it the most reliable option when low FPR and high precision are critical. The Majority Voting Ensemble provides a strong alternative, particularly for scenarios requiring balanced performance across multiple metrics.

\subsection{Ensemble on Additional Datasets}

\begin{table}[]
\centering
\scriptsize
\scalebox{0.82}{%
  \renewcommand{\arraystretch}{1.2}%
  \begin{tabular}{ll|ccc|ccc}
    \toprule
    & & \multicolumn{3}{c|}{\textbf{TEXAS-100}} 
      & \multicolumn{3}{c}{\textbf{PURCHASE-100}} \\
    \cmidrule(lr){3-5}\cmidrule(lr){6-8}
    \textbf{Ensemble Level} & \textbf{Attack} 
        & \textbf{AUC} & \textbf{ACC} & \textbf{TPR} 
        & \textbf{AUC} & \textbf{ACC} & \textbf{TPR} \\
    \midrule
    \multirow{4}{*}{Single-instance}
      & losstraj    
        & 0.789 & 0.722 & 0.014
        & 0.667 & 0.629 & 0.005 \\
      & reference    
        & 0.841 & 0.785 & 0.066
        & 0.729 & 0.690 & 0.014 \\
      & lira         
        & 0.836 & 0.750 & 0.078
        & 0.726 & 0.664 & 0.010 \\
      & calibration  
        & 0.757 & 0.694 & 0.002
        & 0.661 & 0.639 & 0.002 \\
    \midrule
    \multirow{4}{*}{\shortstack{Multi-inst.\\Coverage}}
      & losstraj    
        & 0.732 & 0.667 & 0.018
        & 0.624 & 0.599 & 0.003 \\
      & reference
        & 0.873 & 0.789 & 0.068
        & 0.757 & 0.684 & 0.017 \\
      & lira         
        & 0.826 & 0.561 & 0.005
        & 0.711 & 0.651 & 0.010 \\
      & calibration  
        & 0.759 & 0.698 & 0.006
        & 0.602 & 0.592 & 0.003 \\
    \midrule
    \multirow{4}{*}{\shortstack{Multi-inst.\\Coverage}}
      & losstraj    
        & 0.789 & 0.715 & 0.027 
        & 0.678 & 0.635 & 0.004 \\
      & reference    
        & 0.854 & 0.789 & 0.083  
        & 0.744 & 0.691 & 0.020 \\
      & lira         
        & 0.848 & 0.759 & 0.072 
        & 0.743 & 0.674 & 0.014 \\
      & calibration  
        & 0.759 & 0.698 & 0.005 
        & 0.662 & 0.637 & 0.004 \\
    \midrule
    \multirow{4}{*}{\shortstack{Multi-inst.\\Stability}}
      & losstraj    
        & 0.849 & 0.766 & 0.046
        & 0.725 & 0.657 & 0.006 \\
      & reference    
        & 0.808 & 0.780 & 0.118
        & 0.731 & 0.688 & 0.018 \\
      & lira        
        & 0.853 & 0.761 & 0.079
        & 0.747 & 0.677 & 0.011 \\
      & calibration  
        & 0.758 & 0.696 & 0.008
        & 0.720 & 0.668 & 0.002 \\
    \midrule
    \multirow{3}{*}{Multi-attack}
      & Coverage     
        & 0.952 & 0.909 & 0.094
        & 0.850 & 0.789 & 0.026 \\
      & Stability    
        & 0.902 & 0.816 & 0.165
        & 0.880 & 0.812 & 0.046 \\
      & Maj-vote
        & 0.740 & 0.718 & 0.003
        & 0.694 & 0.647 & 0.008 \\
    \bottomrule
  \end{tabular}%
}
\vspace{0.5cm}
\caption{Performance Comparison for additional datasets in addition to Table~\ref{tab:ensemble-perf-table}. The model architecture is MLP.}
\label{tab:ensemble-perf-table-appendix}
\end{table}

In addition to Table~\ref{tab:ensemble-perf-table}, we have also evaluated the performance of our ensemble on two additional datasets mentioned in Section~\ref{sec:experiement-setup}. Appendix Table~\ref{tab:ensemble-perf-table-appendix}. It should be that our ensemble consistently outperforms single-instance attacks on these two additional datasets.

\begin{table}[h]
\centering
\scriptsize
\scalebox{0.82}{
\renewcommand{\arraystretch}{1.2}
\begin{tabular}{ll|ccc}
\toprule
& & \multicolumn{3}{c}{\textbf{CINIC-10}} \\
\cmidrule(lr){3-5}
\textbf{Ens. Lvl} & \textbf{Attack} & \textbf{AUC} & \textbf{ACC} & \textbf{TPR} \\
\midrule
\multirow{4}{*}{Single-inst.}
& losstraj    & 0.501 & 0.503 & 0.001 \\
& reference   & 0.506 & 0.507 & 0.001 \\
& lira        & 0.505 & 0.505 & 0.001 \\
& calibration & 0.499 & 0.503 & 0.001 \\
\midrule
\multirow{3}{*}{Multi-attack}
& Coverage    & 0.781 & 0.726 & 0.002 \\
& Stability   & 0.796 & 0.730 & 0.011 \\
& Maj-vote    & 0.775 & 0.708 & 0.008 \\
\bottomrule
\end{tabular}
}
\vspace{0.5cm}
\caption{\textbf{Performance on CINIC-10 dataset with distribution shift}. TPR is measured at 0.1\% FPR.}
\label{tab:ensemble_perf_distribution_shift_app}
\vspace{-5pt}
\end{table}

\begin{table}[!htbp]
\centering
\scalebox{0.82}{
\begin{minipage}{\textwidth}

\begin{subtable}{\textwidth}
\raggedright
\scriptsize
\renewcommand{\arraystretch}{1.2}
\begin{tabular}{ll|ccc|ccc|ccc}
\toprule
& & \multicolumn{3}{c|}{\textbf{CIFAR-10}} 
  & \multicolumn{3}{c|}{\textbf{CIFAR-100}} 
  & \multicolumn{3}{c}{\textbf{CINIC-10}} \\
\cmidrule(lr){3-5}\cmidrule(lr){6-8}\cmidrule(lr){9-11}
\textbf{Ens. Lvl} & \textbf{Attack} 
    & \textbf{AUC} & \textbf{ACC} & \textbf{TPR} 
    & \textbf{AUC} & \textbf{ACC} & \textbf{TPR} 
    & \textbf{AUC} & \textbf{ACC} & \textbf{TPR} \\
\midrule
\multirow{4}{*}{Single-inst.}
& losstraj    
  & 0.589 & 0.560 & 0.005
  & 0.716 & 0.650 & 0.007
  & 0.585 & 0.565 & 0.002 \\
& reference    
  & 0.536 & 0.542 & 0.003
  & 0.651 & 0.654 & 0.017
  & 0.542 & 0.543 & 0.003 \\
& lira         
  & 0.527 & 0.525 & 0.001
  & 0.661 & 0.624 & 0.008
  & 0.534 & 0.527 & 0.001 \\
& calibration  
  & 0.572 & 0.551 & 0.004
  & 0.671 & 0.631 & 0.006
  & 0.567 & 0.552 & 0.002 \\
\midrule
\multirow{4}{*}{\shortstack{Multi-inst.\\Coverage}}
& losstraj    
  & 0.556 & 0.535 & 0.003  
  & 0.678 & 0.615 & 0.009  
  & 0.572 & 0.546 & 0.002 \\
& reference    
  & 0.559 & 0.557 & 0.002  
  & 0.701 & 0.676 & 0.010 
  & 0.557 & 0.552 & 0.003 \\
& lira         
  & 0.516 & 0.524 & 0.001 
  & 0.654 & 0.623 & 0.009 
  & 0.525 & 0.525 & 0.001 \\
& calibration  
  & 0.545 & 0.528 & 0.003  
  & 0.637 & 0.595 & 0.006 
  & 0.563 & 0.548 & 0.002 \\
\midrule
\multirow{4}{*}{\shortstack{Multi-inst.\\Maj-vote}}
& losstraj    
  & 0.594 & 0.561 & 0.006  
  & 0.726 & 0.650 & 0.010
  & 0.597 & 0.570 & 0.003 \\
& reference    
  & 0.543 & 0.546 & 0.003  
  & 0.664 & 0.664 & 0.023
  & 0.546 & 0.548 & 0.003 \\
& lira         
  & 0.531 & 0.531 & 0.002  
  & 0.671 & 0.632 & 0.015 
  & 0.539 & 0.531 & 0.002 \\
& calibration  
  & 0.587 & 0.563 & 0.005  
  & 0.690 & 0.637 & 0.010
  & 0.565 & 0.550 & 0.002 \\
\midrule
\multirow{4}{*}{\shortstack{Multi-inst.\\Stability}}
& losstraj    
  & 0.605 & 0.571 & 0.005
  & 0.767 & 0.701 & 0.012
  & 0.593 & 0.567 & 0.003 \\
& reference    
  & 0.523 & 0.528 & 0.003
  & 0.509 & 0.619 & 0.016
  & 0.530 & 0.533 & 0.003 \\
& lira         
  & 0.542 & 0.537 & 0.003
  & 0.671 & 0.629 & 0.020
  & 0.544 & 0.533 & 0.003 \\
& calibration  
  & 0.587 & 0.565 & 0.005
  & 0.734 & 0.674 & 0.013
  & 0.568 & 0.551 & 0.002 \\
\midrule
\multirow{2}{*}{Multi-attack}
& Coverage     
  & 0.751 & 0.705 & 0.002
  & 0.764 & 0.695 & 0.016
  & 0.892 & 0.825 & 0.013 \\
& Stability    
  & 0.839 & 0.741 & 0.034
  & 0.935 & 0.868 & 0.125
  & 0.754 & 0.698 & 0.003 \\
& Maj-vote    
  & 0.831 & 0.758 & 0.028
  & 0.925 & 0.856 & 0.076
  & 0.832 & 0.770 & 0.006\\
\bottomrule
\end{tabular}
\vspace{0.5cm}
\caption{WideResNet-32}
\label{tab:ensemble-perf-table_wrn32_4_appendix}
\end{subtable}

\begin{subtable}{\textwidth}
\raggedright
\scriptsize
\renewcommand{\arraystretch}{1.2}
\begin{tabular}{ll|ccc|ccc|ccc}
\toprule
& & \multicolumn{3}{c|}{\textbf{CIFAR-10}} 
  & \multicolumn{3}{c|}{\textbf{CIFAR-100}} 
  & \multicolumn{3}{c}{\textbf{CINIC-10}} \\
\cmidrule(lr){3-5}\cmidrule(lr){6-8}\cmidrule(lr){9-11}
\textbf{Ens. Lvl} & \textbf{Attack} 
    & \textbf{AUC} & \textbf{ACC} & \textbf{TPR} 
    & \textbf{AUC} & \textbf{ACC} & \textbf{TPR} 
    & \textbf{AUC} & \textbf{ACC} & \textbf{TPR} \\
\midrule
\multirow{4}{*}{Single-inst.}
& losstraj    
  & 0.703 & 0.640 & 0.018
  & 0.497 & 0.506 & 0.001
  & 0.757 & 0.690 & 0.022 \\
& reference    
  & 0.652 & 0.632 & 0.006
  & 0.926 & 0.883 & 0.220
  & 0.710 & 0.681 & 0.006 \\
& lira         
  & 0.622 & 0.595 & 0.009
  & 0.935 & 0.857 & 0.260
  & 0.689 & 0.638 & 0.017 \\
& calibration  
  & 0.633 & 0.603 & 0.005
  & 0.772 & 0.730 & 0.012
  & 0.715 & 0.659 & 0.003 \\
\midrule
\multirow{4}{*}{\shortstack{Multi-inst.\\Coverage}}
& losstraj    
  & 0.640 & 0.584 & 0.014 
  & 0.897 & 0.830 & 0.072 
  & 0.721 & 0.646 & 0.020 \\
& reference    
  & 0.681 & 0.639 & 0.009  
  & 0.935 & 0.868 & 0.222
  & 0.761 & 0.692 & 0.010 \\
& lira         
  & 0.615 & 0.585 & 0.015  
  & 0.935 & 0.856 & 0.239 
  & 0.697 & 0.639 & 0.019 \\
& calibration  
  & 0.561 & 0.542 & 0.005 
  & 0.713 & 0.660 & 0.014
  & 0.672 & 0.625 & 0.002 \\
\midrule
\multirow{4}{*}{\shortstack{Multi-inst.\\Stability}}
& losstraj    
  & 0.706 & 0.662 & 0.031
  & 0.706 & 0.756 & 0.054
  & 0.713 & 0.685 & 0.057 \\
& reference    
  & 0.621 & 0.614 & 0.021
  & 0.901 & 0.891 & 0.371
  & 0.660 & 0.654 & 0.021 \\
& lira
  & 0.647 & 0.611 & 0.020
  & 0.952 & 0.882 & 0.331
  & 0.719 & 0.665 & 0.020 \\
& calibration  
  & 0.722 & 0.667 & 0.014
  & 0.894 & 0.863 & 0.050
  & 0.770 & 0.708 & 0.008 \\
\midrule
\multirow{4}{*}{\shortstack{Multi-inst.\\Maj-vote}}
& losstraj    
  & 0.746 & 0.664 & 0.029  
  & 0.961 & 0.894 & 0.100 
  & 0.823 & 0.735 & 0.042 \\
& reference    
  & 0.703 & 0.666 & 0.016  
  & 0.945 & 0.897 & 0.252 
  & 0.771 & 0.722 & 0.016 \\
& lira         
  & 0.652 & 0.614 & 0.017  
  & 0.956 & 0.887 & 0.310 
  & 0.738 & 0.676 & 0.022 \\
& calibration  
  & 0.647 & 0.610 & 0.012  
  & 0.778 & 0.690 & 0.045
  & 0.740 & 0.677 & 0.005 \\
\midrule
\multirow{2}{*}{Multi-attack}
& Coverage     
  & 0.848 & 0.781 & 0.028
  & 0.956 & 0.887 & 0.285
  & 0.916 & 0.849 & 0.053 \\
& Stability    
  & 0.909 & 0.834 & 0.091
  & 0.992 & 0.957 & 0.545
  & 0.892 & 0.829 & 0.032 \\
& Maj-vote
  & 0.887 & 0.817 & 0.085
  & 0.986 & 0.949 & 0.460
  & 0.914 & 0.849 & 0.048 \\

\bottomrule
\end{tabular}
\vspace{0.5cm}
\caption{MobileNetV2}
\label{tab:ensemble-perf-table_mobilenet_appendix}
\end{subtable}

\begin{subtable}{\textwidth}
\raggedright
\scriptsize
\renewcommand{\arraystretch}{1.2}
\begin{tabular}{ll|ccc|ccc|ccc}
\toprule
& & \multicolumn{3}{c|}{\textbf{CIFAR-10}} 
  & \multicolumn{3}{c|}{\textbf{CIFAR-100}} 
  & \multicolumn{3}{c}{\textbf{CINIC-10}} \\
\cmidrule(lr){3-5}\cmidrule(lr){6-8}\cmidrule(lr){9-11}
\textbf{Ens. Lvl} & \textbf{Attack} 
    & \textbf{AUC} & \textbf{ACC} & \textbf{TPR} 
    & \textbf{AUC} & \textbf{ACC} & \textbf{TPR} 
    & \textbf{AUC} & \textbf{ACC} & \textbf{TPR} \\
\midrule
\multirow{4}{*}{Single-inst.}
& losstraj    
  & 0.668 & 0.605 & 0.021
  & 0.826 & 0.738 & 0.031
  & 0.779 & 0.698 & 0.027 \\
& reference    
  & 0.663 & 0.632 & 0.016
  & 0.874 & 0.821 & 0.005
  & 0.775 & 0.717 & 0.028 \\
& lira         
  & 0.651 & 0.605 & 0.021
  & 0.888 & 0.793 & 0.140
  & 0.771 & 0.693 & 0.023 \\
& calibration  
  & 0.636 & 0.599 & 0.003
  & 0.728 & 0.702 & 0.008
  & 0.706 & 0.660 & 0.004 \\
\midrule
\multirow{4}{*}{\shortstack{Multi-inst.\\Coverage}}
& losstraj    
  & 0.609 & 0.563 & 0.013  
  & 0.743 & 0.664 & 0.025
  & 0.711 & 0.638 & 0.012 \\
& reference    
  & 0.657 & 0.619 & 0.016 
  & 0.853 & 0.777 & 0.005 
  & 0.763 & 0.687 & 0.045 \\
& lira         
  & 0.643 & 0.593 & 0.028  
  & 0.891 & 0.787 & 0.149 
  & 0.766 & 0.678 & 0.025 \\
& calibration  
  & 0.565 & 0.548 & 0.003  
  & 0.641 & 0.598 & 0.006 
  & 0.652 & 0.619 & 0.003 \\
\midrule
\multirow{4}{*}{\shortstack{Multi-inst.\\Stability}}
& losstraj    
  & 0.722 & 0.646 & 0.048
  & 0.838 & 0.792 & 0.079
  & 0.830 & 0.741 & 0.070 \\
& reference    
  & 0.673 & 0.649 & 0.037
  & 0.893 & 0.855 & 0.190
  & 0.794 & 0.752 & 0.075 \\
& lira         
  & 0.670 & 0.623 & 0.039
  & 0.903 & 0.810 & 0.194
  & 0.799 & 0.720 & 0.043 \\
& calibration  
  & 0.703 & 0.647 & 0.011
  & 0.875 & 0.821 & 0.026
  & 0.769 & 0.718 & 0.007 \\
\midrule
\multirow{4}{*}{\shortstack{Multi-inst.\\Maj-vote}}
& losstraj    
  & 0.672 & 0.604 & 0.031  
  & 0.857 & 0.757 & 0.047 
  & 0.796 & 0.703 & 0.037 \\
& reference    
  & 0.686 & 0.647 & 0.035  
  & 0.894 & 0.828 & 0.123 
  & 0.811 & 0.735 & 0.064 \\
& lira         
  & 0.680 & 0.626 & 0.032  
  & 0.919 & 0.826 & 0.180 
  & 0.810 & 0.722 & 0.037 \\
& calibration  
  & 0.646 & 0.606 & 0.006  
  & 0.729 & 0.687 & 0.018 
  & 0.704 & 0.653 & 0.004 \\
\midrule
\multirow{3}{*}{Multi-attack}
& Coverage     
  & 0.740 & 0.718 & 0.003
  & 0.926 & 0.843 & 0.132
  & 0.924 & 0.853 & 0.085 \\
& Stability    
  & 0.871 & 0.790 & 0.083
  & 0.976 & 0.934 & 0.322
  & 0.893 & 0.825 & 0.053 \\
& Maj-vote
  & 0.852 & 0.790 & 0.070
  & 0.958 & 0.896 & 0.273
  & 0.906 & 0.841 & 0.066 \\
\bottomrule
\end{tabular}
\vspace{0.5cm}
\caption{VGG-16}
\label{tab:ensemble-perf-table_vgg16_appendix}
\end{subtable}

\end{minipage}
}
\vspace{0.5cm}
\caption{Performance of Ensemble for Different Architectures. 
         TPR stands for True Positive Rate measured at 0.1\% FPR. This table is an extension of Table~\ref{tab:ensemble-perf-table}.}
\label{tab:ensemble-perf-table-all-other-model-appendix}
\end{table}

\subsection{Performance under Auxiliary-Target Distribution Mismatch}
\label{ssec:distributionshift}
For CINIC-10, we additionally consider a more challenging setting where the shadow model is trained on a different data distribution from the target model. Specifically, the auxiliary dataset $\auxDataset$ consists only of CIFAR-10 subsets from CINIC-10, while the target dataset $\targetDataset$ consists of ImageNet subsets, with each subset containing 30,000 samples. The results are presented in Appendix Table~\ref{tab:ensemble_perf_distribution_shift_app}. Under this distribution mismatch, all base attacks perform significantly worse compared to their performance on CINIC-10 without such a mismatch, while our ensemble still significantly outperforms the individual attacks (see Appendix Table~\ref{tab:ensemble-perf-table}).


\section{LiRA Online v.s. Offline}
\label{sec:lira-online-offline-comp-appendix}
\begin{figure}[]
    \centering
    \includegraphics[width=0.2\textwidth]{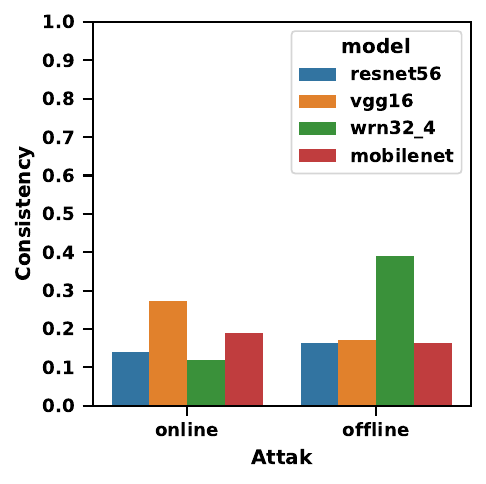}
    \caption{Consistency of LiRA Online vs LiRA Offline.}
    \label{fig:lira_online_vs_offline}
\end{figure}
    
In \citet{lira}, both online and offline versions of LiRA were proposed. For a given sample $(x, y)$, the online version trains half of the shadow models with $(x, y)$ included in the training set (IN models), and the other half with $(x, y)$ excluded (OUT models). The confidence scores from both sets are then used to perform a likelihood ratio test. This setup requires retraining IN models for each new inference sample, making the online version computationally expensive. To improve efficiency, \citet{lira} introduced an offline version that only trains OUT models and performs a one-sided hypothesis test. While this approach avoids the additional randomness introduced by IN models, it still exhibits similarly low consistency, as demonstrated in Appendix Figure~\ref{fig:lira_online_vs_offline}. The target model is CIFAR-10, and the offline LiRA shares the same 20 shadow models used for online LiRA, but it only obtains confidence scores in OUT models.

\begin{figure}[]
    \vspace{1em}
    \centering
    \includegraphics[width=0.35\textwidth]{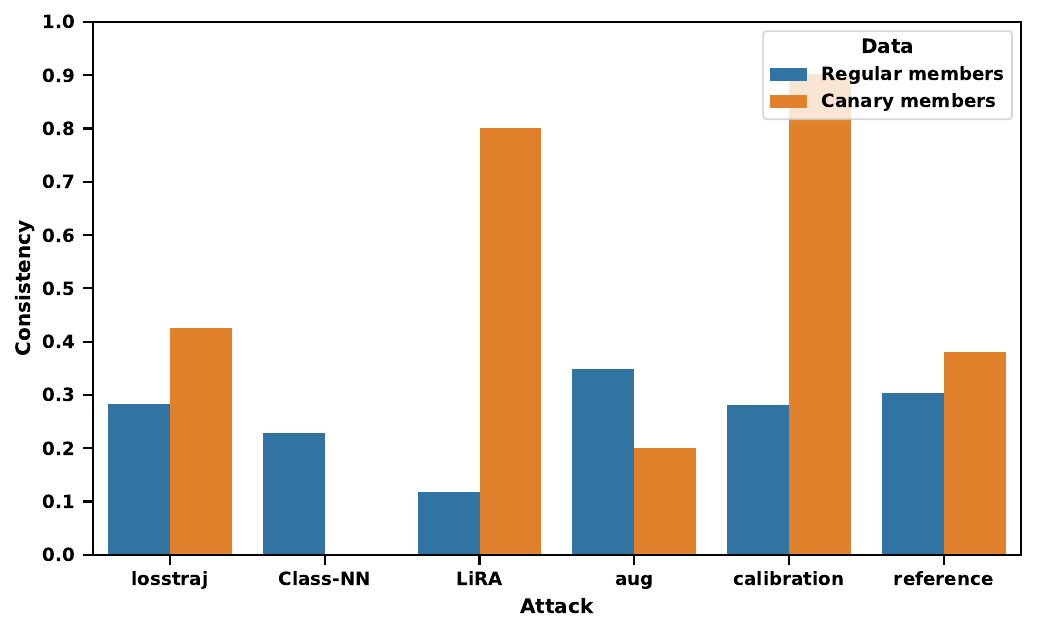}
    \caption{Consistency on Outliers VS Regular Samples.}
    \label{fig:consistency_canary}
\end{figure}
    
\begin{figure}[]
    \centering
    \vspace{1.5em}
    \begin{subfigure}[b]{0.23\textwidth}
        \centering
        \includegraphics[width=\textwidth]{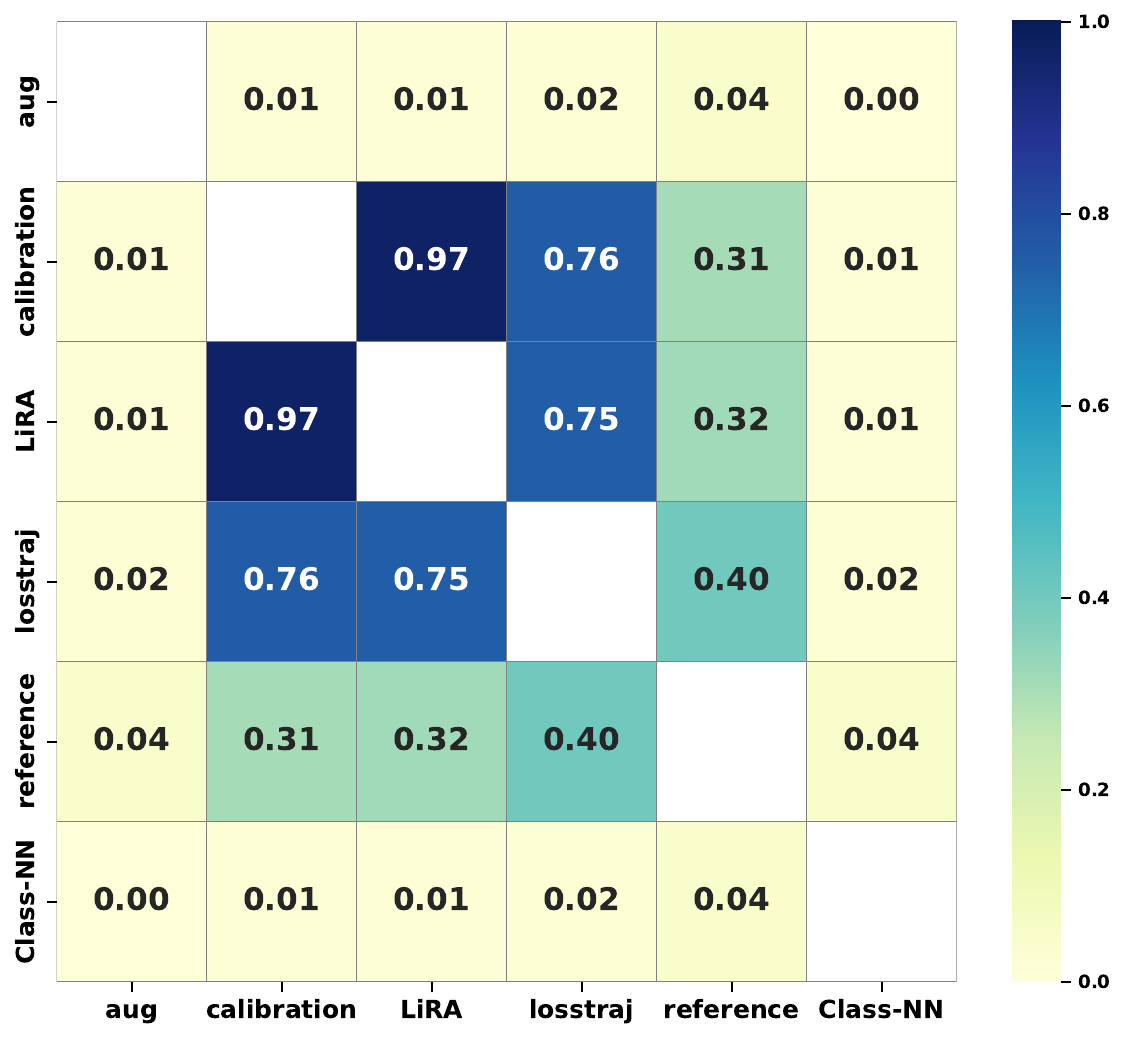}
        \caption{Coverage (Canary)}
        \label{fig:jaccard_coverage_can}
    \end{subfigure}
    \hfill
    \begin{subfigure}[b]{0.23\textwidth}
        \centering
        \includegraphics[width=\textwidth]{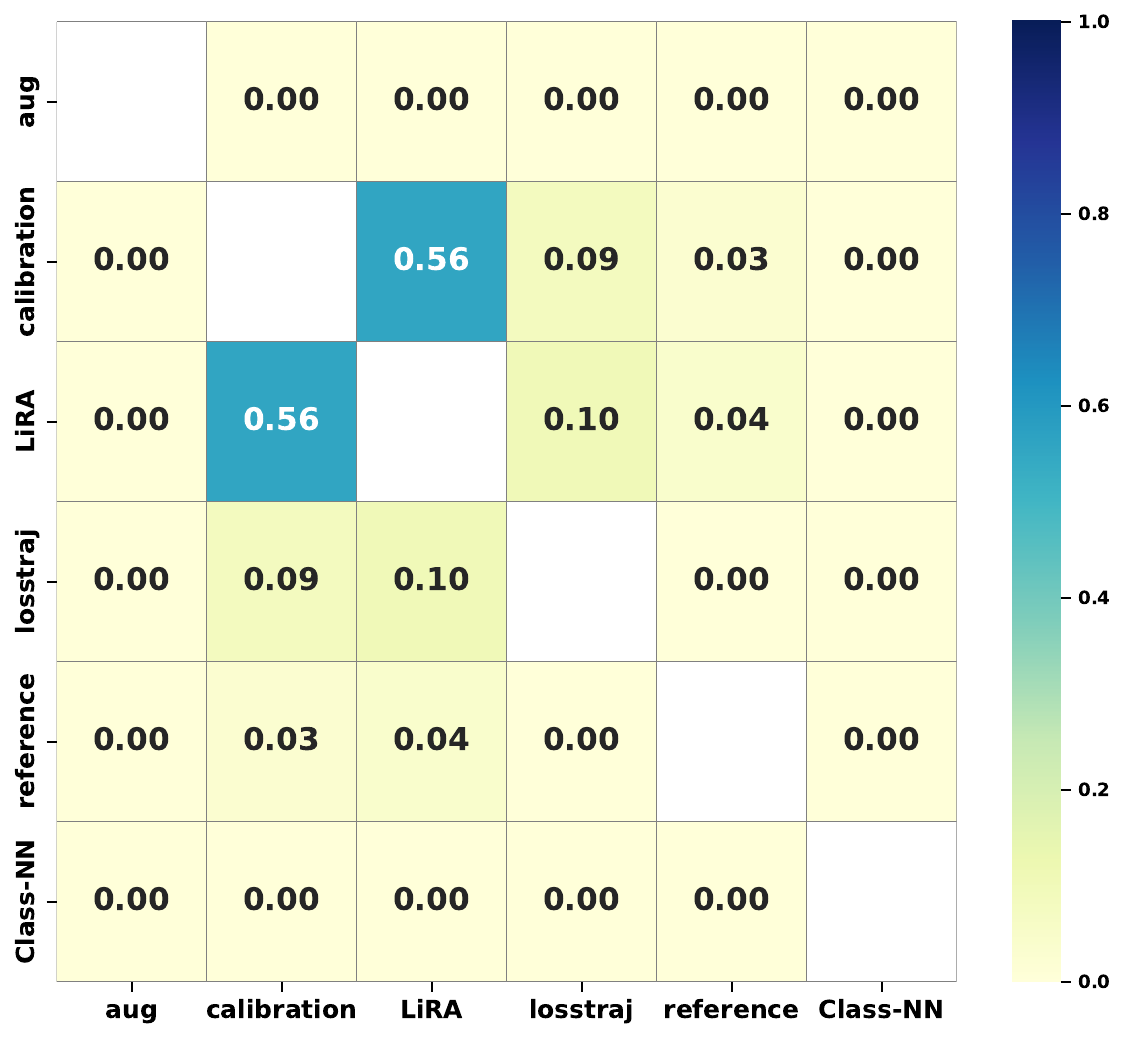}
        \caption{Stability (Canary)}
        \label{fig:jaccard_stability_can}
    \end{subfigure}
    
    \vspace{1.5em} 
    
    \begin{subfigure}[b]{0.23\textwidth}
        \centering
        \includegraphics[width=\textwidth]{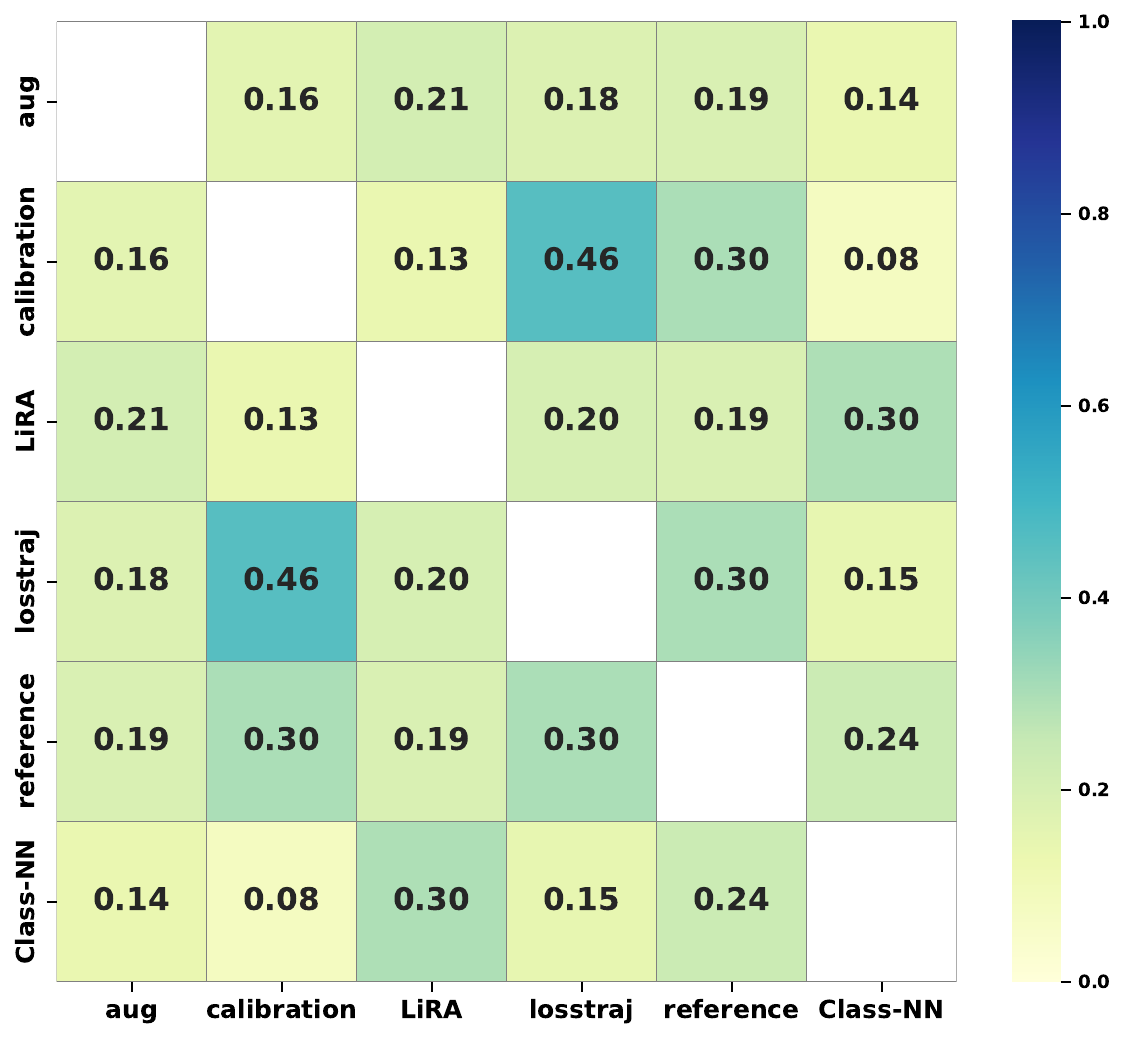}
        \caption{Coverage (Regular)}
        \label{fig:jaccard_coverage_non_can}
    \end{subfigure}
    \hfill
    \begin{subfigure}[b]{0.23\textwidth}
        \centering
        \includegraphics[width=\textwidth]{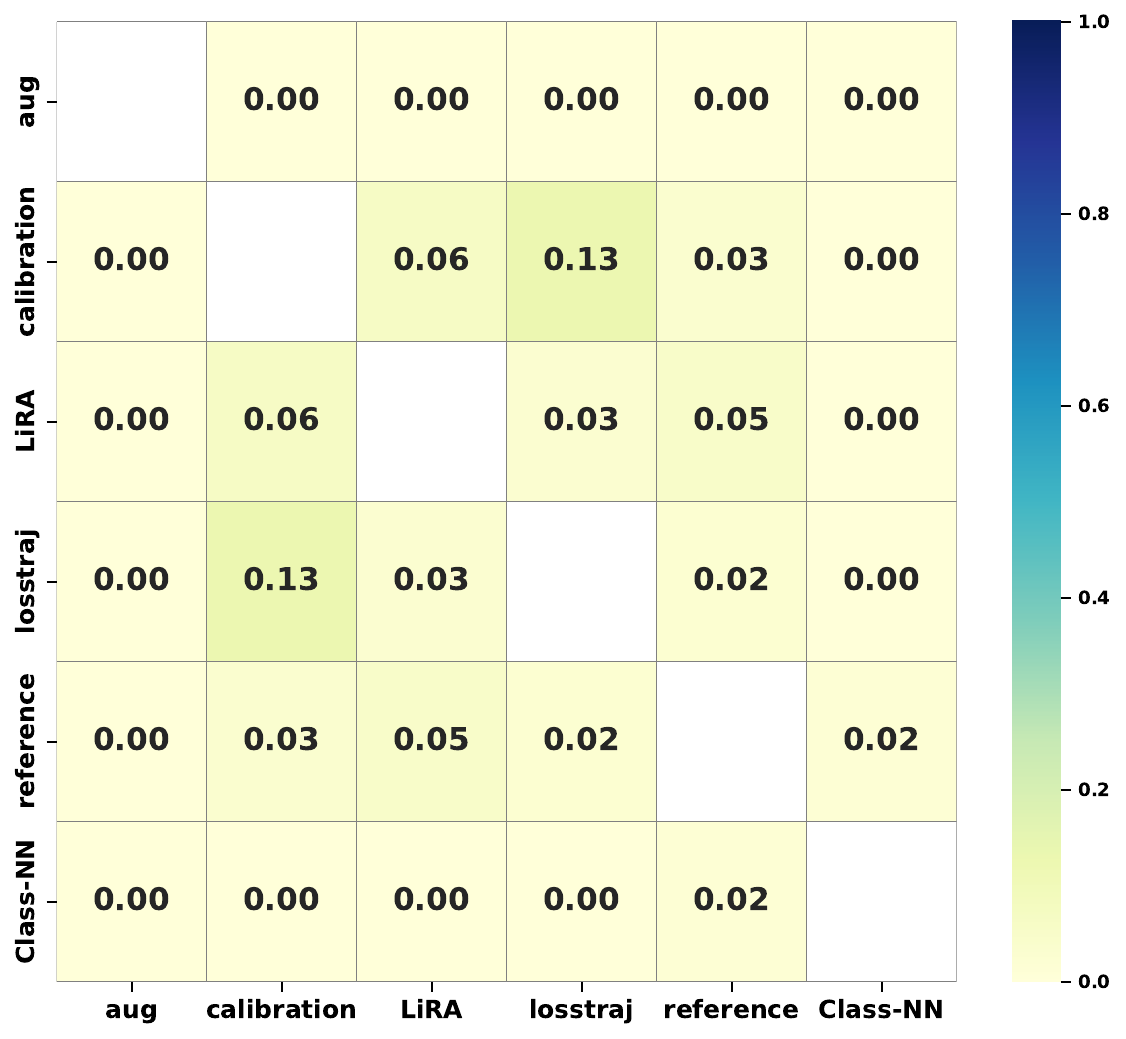}
        \caption{Stability (Regular)}
        \label{fig:jaccard_stability_non_can}
    \end{subfigure}

    \vspace{1em}
    
    \caption{Disparity of Membership Inference Attacks on Outliers (canary samples) VS regular samples. Instances are prediting at FPR=0.1.}
    \label{fig:disparity-of-MIAs-on-canaries-appendix}
\end{figure}

\section{MIA on Out of Distribution Sample}
\label{sec:mia-on-ood}

In this section, we extend our analysis to out-of-distribution (OOD) samples, focusing on the MIAs' disparities on extreme OOD members. Following prior work~\cite{aerni2024misleading, lira}, we create a set of mislabeled data points, referred to as \textit{canary samples}, to simulate such cases. Specifically, we sample a subset $\dataset_C$ from the target dataset $\targetDataset$ and relabel each point $(x, y) \in \dataset_C$ as $(x, y')$ such that $y' \neq y$. These mislabeled points are inserted back into $\targetDataset$ and used as members in the training set. Since the target model must overfit to these samples to classify them correctly, they represent highly vulnerable members.

We expect attacks to have higher consistency on canary members compared to regular members, as they should be easier to identify and most instances should agree on their membership. We also expect attacks to have higher similarity in predicting canary samples for the same reason.

In our experiment, we inject 300 canary samples into CIFAR-10. We exclude the LOSS attack, as its predictions are deterministic across different instances. In Appendix Figure~\ref{fig:consistency_canary}, the regular members refer to samples in the target training set that are not relabeled. Some attacks show higher consistency on canary members than on regular members. For example, the Calibration Loss Attack achieves consistency greater than 0.9. This is because the loss of mislabeled (canary) data is significantly higher on the shadow model, while the target model has memorized the new label ($y'$). Therefore, this task becomes trivial for the Calibration Loss Attack. In general, identifying an extreme outlier member is trivial for attacks that compare the loss on the target model to the loss on the shadow model—this is the case for both LiRA and the Calibration Loss Attack. It explains the large gap in consistency between regular and canary members for these two attacks. For the Loss Trajectory and Reference Attacks, this task is less trivial but still easier, as canary samples are more memorized. However, the Class-NN Attack and Augmentation Attack exhibit decreased performance, since these attacks don't have those extreme OOD samples presented in their attack training sets. In Appendix Figure~\ref{fig:set_size_with_canary_appendix}, we observe that the Class-NN and the Augmentation Attack perform worse on canary samples than on regular samples, further demonstrating their inability to handle extreme OOD samples.

In Appendix Figure~\ref{fig:disparity-of-MIAs-on-canaries-appendix}, we observe that for canary samples, the Jaccard similarity between the coverage and stability of some attack pairs is significantly higher compared to regular samples. For example, LiRA and the Calibration Loss Attack exhibit a coverage similarity of 0.97, indicating that they are almost predicting the same set of members. In Appendix Figure~\ref{fig:set_size_canary}, we confirm that the coverage of LiRA and the Calibration Loss Attack nearly includes all canary members, which explains this high similarity.

\begin{figure}[H]
    \begin{subfigure}[c]{0.25\textwidth}
    \centering
        \includegraphics[width=\textwidth]{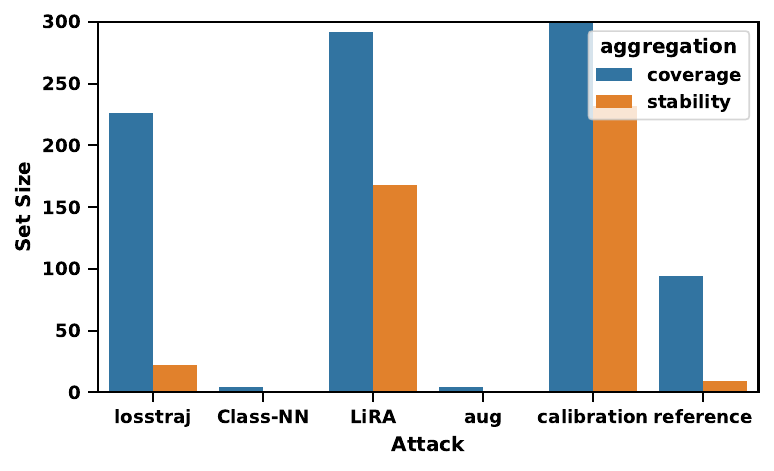}
        \caption{Set Sizes on Canary Members}
        \label{fig:set_size_canary}
    \end{subfigure}

    \vspace{1.5em}

    \begin{subfigure}[c]{0.25\textwidth}
    \centering
        \includegraphics[width=\textwidth]{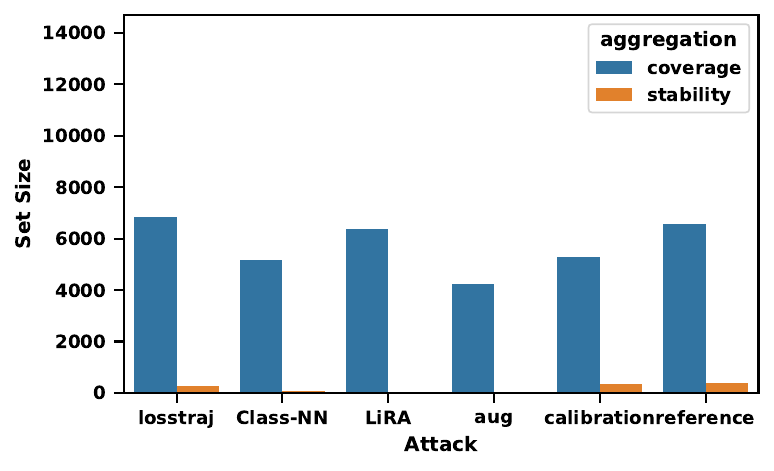}
        \caption{Set Sizes on Regular Members}
        \label{fig:set_size_regular}
    \end{subfigure}
    \vspace{1em}
    \caption{Set size of coverage and stability on canary and regular members. The target dataset is CIFAR-10 with 300 canary members and 14,700 regular members.}
    \label{fig:set_size_with_canary_appendix}
\end{figure}

\end{document}